\documentclass[11pt]{article}

\usepackage[margin=1in]{geometry}

\usepackage{tgtermes}
\usepackage{newtxtext}
\usepackage{newtxmath}

\usepackage{amsmath}
\usepackage{bm}

\usepackage{graphicx}
\usepackage{xcolor}
\usepackage{colortbl}
\usepackage{microtype}
\usepackage{tikz}
\usetikzlibrary{positioning}

\usepackage{booktabs}
\usepackage{tabularx}
\usepackage{multirow}
\usepackage{float}
\usepackage{placeins}

\usepackage{algorithm}
\usepackage{algpseudocode}

\usepackage{subcaption}   

\usepackage[numbers,sort&compress]{natbib}
\usepackage[hidelinks]{hyperref}

\DeclareRobustCommand{\revised}[1]{#1}
\DeclareRobustCommand{\minorrev}[1]{#1}

\newenvironment{revisedblock}{}{}

\title{How to Purchase Labels? A Cost-effective Approach Using Active Learning Markets}

\author{
  Xiwen Huang\thanks{Dyson School of Design Engineering, Imperial College London.
  Email: \texttt{xiwen.huang23@imperial.ac.uk}.}
  \and
  Pierre Pinson\thanks{Dyson School of Design Engineering, Imperial College London;
  Department of Technology, Management and Economics,
  Technical University of Denmark;
  CoRE, Aarhus University.
  Email: \texttt{p.pinson@imperial.ac.uk}.}
}

\date{}

\begin{document}
\maketitle

\begin{abstract}
    
We introduce and analyse active learning markets as a way to purchase labels, in situations where analysts aim to acquire additional data to improve model fitting, or to better train models for predictive analytics applications. This comes in contrast to the many proposals that already exist to purchase features and examples. By originally formalizing the market clearing as an optimisation problem, we integrate budget constraints and improvement thresholds into the label acquisition process. We focus on a single-buyer-multiple-seller setup and propose the use of two active learning strategies (variance based and query-by-committee based), paired with distinct pricing mechanisms. \minorrev{They are compared to benchmark baselines including random sampling and a greedy knapsack heuristic.} The proposed strategies are validated on real-world datasets from two critical application domains: real estate pricing and energy forecasting. Results demonstrate the robustness of our approach, consistently achieving superior performance with fewer labels acquired compared to conventional methods. Our proposal comprises an easy-to-implement practical solution for optimizing data acquisition in resource-constrained environments.
\end{abstract}







 \section{Introduction}\label{Intro}

In today’s data-driven economy, data is often referred to as the ``new oil", reflecting its critical role in driving technological innovation and economic growth. The global market for data analytics is projected to expand significantly in the coming years, growing from USD 348.21 billion in 2024 to USD 924.39 billion in 2032, with a compound annual growth rate of 13.9\% \citep{fortunebusinessinsights2023}. The rapid proliferation of data, generated on massive scales daily, has paved the way for advances in big data analytics, offering innovative solutions in areas such as transportation, energy, real estate, and healthcare, among others. As an example, in the UK, Citymapper leverages freely available real-time transportation data to enhance urban navigation and improve public transport efficiency \citep{tavmen2020data}. However, in many cases, the ownership of data is distributed among heterogeneous agents, while data is deemed a commodity that must be strategically acquired. For instance, as businesses increasingly recognize the importance of predictive capabilities, many face the challenge of acquiring the necessary volume and quality of data to train machine learning models. It is of utmost importance to develop appropriate platforms and market ecosystems supporting the sharing, valuation, and acquisition of data. 

To set the scene, consider the following two scenarios. First, an analyst in a real estate company wants to train a model to predict house prices in a certain area. The analyst has access to all transactions history of houses from their company. Also, in order to have more training data, they may obtain information about other houses with different characteristics (e.g., numbers of rooms, distance to local transportation and schools, etc.). However, the corresponding labels (i.e., the actual prices at which these houses were sold) may not be public, while being necessary for training. How should the analyst decide which labels to acquire and at what price? Second, an analyst in the energy sector aims to build a model to forecast energy consumption for an educational building using time-series data from various sources. Data from other buildings (labels) could be very useful to additionally train the model. Given a limited budget, what strategy should be used to acquire these additional labels?

In a general setup, we place ourselves in a supervised learning framework, where data for input features (equivalently referred to as explanatory variables) may be abundant, while the availability of labels (i.e., values for the target variable) is limited. To develop the idea and a proof of concept, we specifically use linear regression. Linear regression has many practical advantages, including simplicity, interpretability, and the availability of a closed-form solution for parameter estimates. Furthermore, linear regression has also been used by others as a basis to introduce various types of data markets \citep{dekel2010incentive, cummings_truthful_2015}, \revised{which produces} important market properties such as incentive compatibility and truthfulness. However, one should be able to broaden the concepts presented to a class of models that is more general than linear regression only. For instance, as discussed by \citet{pinson2022regression}, data markets developed for the linear regression case can be readily generalized to the case of regression models that are linear in their parameters; this includes, for instance, polynomial- and spline-based regression. In our framework, an analyst seeks to acquire additional labels to improve model fitting, or to improve the predictive performance of the model at hand. These may be highly dependent on the availability and quality of the labels that are used. Labels may be difficult or costly to obtain. 

Existing approaches to data and information markets fall into various categories. Given an analytics task at hand, one may purchase solutions for that task (as in prediction markets), or one may purchase actual data to be used as input to solving that analytics task. For descriptive and predictive analytics, this may translate to having (i) observation markets and (ii) feature markets (also referred to as regression markets). In the first case, the buyer aims to acquire complete rows (with values for both input features and the target variable) to have more data to train their model. In the second case, emphasis is placed on purchasing additional features (i.e., complete columns) to augment the model at hand and to improve their ability to explain variations in the target variable. These markets have been extensively studied \citep{agarwal_marketplace_2019, pinson2022regression} and \revised{are mainly based on} Shapley values to define revenues and payments. Alternatively, \citet{han2022trading} proposed the use of Lasso-based estimation to elicit the willingness to pay of data buyers. Moreover, existing research often emphasizes bulk data acquisition or continuous data delivery, such as in Data as a Commodity (e.g., Amazon Mechanical Turk) or Data as a Service (e.g., Google BigQuery). These approaches fall short of addressing the need for targeted and cost-effective data acquisition, especially for machine learning applications where high-quality, informative data points are more valuable than large quantities of generic data. \revised{At the same time, the growing presence of platforms such as AWS Data Exchange, Ocean Protocol, and crowdsourcing markets like Amazon Mechanical Turk demonstrates that market-style procurement of data is already a practical reality. What remains unexplored is how to integrate active learning principles into these emerging markets to guide selective, budget-constrained acquisition of the most informative labels---which is the gap this paper addresses.}

In contrast, here, we introduce a new type of data market to specifically purchase labels. We refer to such markets as \emph{active learning markets}, since active learning serves as a basis for acquiring labels in a value-oriented manner. In active learning markets, buyers possess full feature sets, but only a subset of values for the target variable of interest. This calls for strategic acquisition of additional labels to improve model performance. This setup closely mirrors real-world scenarios faced by companies where the resources for data acquisition (for the target variable) may be limited. Despite advances in data monetization and pricing models \citep{agarwal_marketplace_2019, mehta2021sell}, frameworks often overlook budget constraints and fail to propose efficient strategies for buyers to prioritize high-value data points. Furthermore, although studies such as \citet{cummings_accuracy_2015} propose variance-constrained mechanisms for data selection, they do not accommodate budget constraints and strategic selection of the data to be purchased. This gap is particularly pronounced in industries \revised{such as} energy forecasting, where data acquisition \revised{is} expensive, but essential to achieve high prediction accuracy \citep{wang2023data, settles2011theories}. 

We bridge this research gap by introducing and analyzing active learning markets. Our core contribution is then to incorporate both overall budget constraints and minimum improvement requirements in data acquisition problems. This dual consideration is a departure from traditional data purchase problems. Second, we extend the use of \revised{the} active learning principle within a data market context. Eventually, we showcased the interest of our proposal based on case studies inspired by real-world applications. To realize these contributions, we first define the market setup, introducing the market participants and the key components of the underlying resource allocation problem. We then formalize that resource allocation problem within an optimization framework. Rather than solving this optimization problem directly, we use active learning as a greedy approach to data acquisition. This is based on two alternative active learning strategies (variance-based active learning, VBAL, and query-by-committee-based active learning, QBCAL), each paired with distinct data pricing mechanisms. For comparison, we \revised{use} a random sampling \revised{corrected} strategy as a benchmark, and \minorrev{also include a greedy knapsack heuristic baseline (GK) when seller costs are heterogenous }. To validate our proposed strategies, we apply them to real-world datasets in two critical domains: real estate pricing and energy forecasting. Through these applications, we analyse and discuss the salient features of active learning markets, with the aim \revised{of underlining} their practical value in optimizing data acquisition under various constraints.

The remainder of this work is organized as follows. Section \ref{Introducing active learning markets} introduces the preliminaries of data markets in general and of active learning markets in particular. Section \ref{methodology} provides an overview of our active learning marketplace and details our VBAL-based and QBCAL-based approaches while using random sampling corrected strategy as a benchmark. Section \ref{real-world applcations} \revised{presents our proposal for the active learning market} for 2 different real-world applications and discusses the results from the perspective of both \revised{the analyst and the sellers}. Section \ref{conclusions} concludes the work while offering perspectives for future work. 

\section{Introducing active learning markets}\label{Introducing active learning markets}

After briefly discussing current challenges within data valuation, we introduce how active learning markets fit within the current landscape of data markets for analytics applications. We subsequently describe market participants, the core optimization problem for purchasing labels, and related pricing considerations.

\subsection{Data valuation}
Data, as an asset, \revised{has} three unique characteristics that distinguish it from traditional goods, each presenting its own set of challenges. First, data can be easily replicated at zero cost, 
\revised{introducing} new issues in \revised{data} pricing \citep{acemoglu_2022_too, mehta2021sell, liang2018survey, pei2020survey,yu2017data, li2014theory}. Second, data is generated in vast volumes and at high velocity. For instance, according to \citet{statista2024}, the data generated in 2025 is estimated at \revised{reaching} 181 zettabytes, increased by \revised{more than} 150\% compared to \revised{the} data generated in 2023---\revised{120 zettabytes}. This massive scale creates significant challenges for big data analytics \citep{maheshwari2021role, mariani2020exploring, wang2020analytics}. Third, data ownership is distributed \revised{by} companies, each with heterogeneous preferences due to differing privacy concerns and competitive dynamics \citep{abbas2021business, busch2022personal, oecd2018}.

Addressing these challenges requires new paradigms for data marketplaces, particularly given that the value of data is not intrinsic, but highly dependent on its context---its quality, timeliness, and relevance to specific analytics tasks.

\subsection{Alternative data market setups}
\label{subsection:market scenario}

One can think of data markets for analytics in different ways, depending on the type of data to be acquired and the way they may be used as input to analytics tasks. \emph{Data purchase} generally refers to the acquisition of additional data points from data sellers against some form of monetary transaction. Analysts may engage in data purchase to improve the fit of their model and to improve the predictive performance of their model.

\revised{Within a supervised learning framework, let \(\Omega = \{1, \dots, P\}\) index the set of input features. 
For each \(p \in \Omega\), the \(p\)-th input feature is denoted by \(x_p\) (also referred to as an explanatory variable).
} Let \(Y\) denote the target variable (also known as response variable). As a basis \revised{for} introducing data markets, we restrict ourselves to a simpler linear regression framework. The notation \(x\) is used for the observations of the features in \(\Omega\) and \(y\) for the observed values of the target variable \(Y\). Let \(N\) denote the total number of observations, for both input features and response variable. We start by introducing two cases which have already been explored within the scientific literature, before introducing our novel framework.

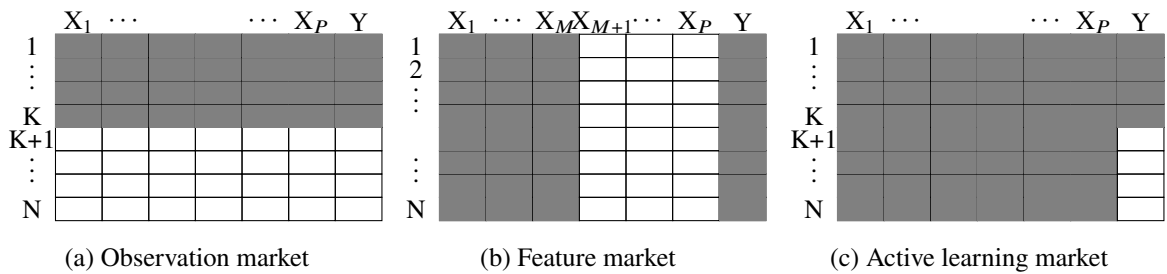
\begin{figure}[!htbp]
\subfloat[Observation market]
{
\begin{tikzpicture}[scale=0.31]
        \begin{scope}[shift={(0, -7)}]
            \node at (1, 5.5) {\small X$_1$};
            \node at (3, 5.5) {\small $\cdots$};
            \node at (5, 5.5) {\small };
            \node at (7, 5.5) {\small };
            \node at (9, 5.5) {\small $\cdots$};
            \node at (11, 5.5) {\small X$_P$};
            \node at (13, 5.5) {\small Y};

            \def\k{2}  

            \foreach \x in {0, 2, 4, 6, 8, 10, 12} {
                \foreach \y in {-3, -2, -1, 0, 1, 2, 3, 4 } {
                    \draw (\x, \y) rectangle (\x+2, \y+1);
                    \ifnum \y=4 
                        \fill[gray] (\x, \y) rectangle (\x+2, \y+1);
                    \fi                    \ifnum \y=3 
                        \fill[gray] (\x, \y) rectangle (\x+2, \y+1);
                    \fi
                    \ifnum \y=2 
                        \fill[gray] (\x, \y) rectangle (\x+2, \y+1);
                    \fi
                    \ifnum \y=1 
                        \fill[gray] (\x, \y) rectangle (\x+2, \y+1);
                    \fi
                          }
            }
            
            \node at (-1, 4.5) {\small 1};
            \node at (-1, 3.5) {\small $\vdots$};
            \node at (-1, 2.5) {\small };
            \node at (-1, 1.5) {\small K};
            \node at (-1, 0.5) {\small K+1};
            \node at (-1, -0.5) { \small $\vdots$};
            \node at (-1, -1.5) {};
            \node at (-1, -2.5) {\small N};

        \end{scope}
    \end{tikzpicture}
}
\subfloat[Feature market]
{
         \begin{tikzpicture}[scale = 0.31]
        \begin{scope}[shift={(0, -7)}]
            \node at (1, 5.5) {\small X$_1$};
            \node at (3, 5.5) {\small $\cdots$};
            \node at (5, 5.5) {\small X$_M$};
            \node at (7, 5.5) {\small X$_{M+1}$};
            \node at (9, 5.5) {\small $\cdots$};
            \node at (11, 5.5) {\small X$_P$};
            \node at (13, 5.5) {\small Y};

          \foreach \x in {0, 2, 4, 6, 8, 10, 12} {
                \foreach \y in {-3, -2, -1, 0, 1, 2, 3, 4 } {
                    \draw (\x, \y) rectangle (\x+2, \y+1);
                    \ifnum \x=0
                        \fill[gray] (\x, \y) rectangle (\x+2, \y+1);
                    \fi
                    \ifnum \x=2
                        \fill[gray] (\x, \y) rectangle (\x+2, \y+1);
                    \fi
                    \ifnum \x=4
                        \fill[gray] (\x, \y) rectangle (\x+2, \y+1);
                    \fi
                    \ifnum \x=12
                        \fill[gray] (\x, \y) rectangle (\x+2, \y+1);
                    \fi
                }
            }
            \node at (-1, 4.5) {\small 1};
            \node at (-1, 3.5) {\small 2};
            \node at (-1, 2.5) {\small $\vdots$};
            \node at (-1, 1.5) { };
            \node at (-1, 0.5) { };
            \node at (-1, -0.5) {\small $\vdots$ };
            \node at (-1, -1.5) {};
            \node at (-1, -2.5) {\small N};

        \end{scope}
    \end{tikzpicture}
}
\subfloat[Active learning market]
{
     \begin{tikzpicture}[scale = 0.31]
        \begin{scope}[shift={(0, -7)}]
            \node at (1, 5.5) {\small X$_1$};
            \node at (3, 5.5) {\small $\cdots$};
            \node at (5, 5.5) {\small };
            \node at (7, 5.5) {\small };
            \node at (9, 5.5) {\small $\cdots$};
            \node at (11, 5.5) {\small X$_P$};
            \node at (13, 5.5) {\small Y};

            \def\k{2}  

            \foreach \x in {0, 2, 4, 6, 8, 10, 12} {
                \foreach \y in {-3, -2, -1, 0, 1, 2, 3, 4 } {
                    \draw (\x, \y) rectangle (\x+2, \y+1);
                    \ifnum \y=4 
                        \fill[gray] (\x, \y) rectangle (\x+2, \y+1);
                    \fi                    \ifnum \y=3 
                        \fill[gray] (\x, \y) rectangle (\x+2, \y+1);
                    \fi
                    \ifnum \y=2 
                        \fill[gray] (\x, \y) rectangle (\x+2, \y+1);
                    \fi
                    \ifnum \y=1 
                        \fill[gray] (\x, \y) rectangle (\x+2, \y+1);
                    \fi
                    \ifnum \x<12
                        \ifnum \y=-3
                            \fill[gray] (\x, \y) rectangle (\x+2, \y+1);
                        \fi
                        \ifnum \y=-2
                            \fill[gray] (\x, \y) rectangle (\x+2, \y+1);
                        \fi
                        \ifnum \y=-1
                            \fill[gray] (\x, \y) rectangle (\x+2, \y+1);
                        \fi
                        \ifnum \y=0
                            \fill[gray] (\x, \y) rectangle (\x+2, \y+1);
                        \fi
                    \fi                }
            }
            
            \node at (-1, 4.5) {\small 1};
            \node at (-1, 3.5) {\small $\vdots$};
            \node at (-1, 2.5) {\small };
            \node at (-1, 1.5) {\small K};
            \node at (-1, 0.5) {\small K+1};
            \node at (-1, -0.5) { \small $\vdots$};
            \node at (-1, -1.5) {};
            \node at (-1, -2.5) {\small N};

        \end{scope}
    \end{tikzpicture}
}

    \vspace{2mm}
    \caption{Graphical representations of observation, feature and active learning markets, based on both design matrix and response vector (grey: data/features owned by the buyer; white: data/features the sellers may offer). \label{scenario}}
\vspace{-3mm}
\end{figure}

In the first case, which we refer to as \emph{observation market}, the analyst may already own, or have access to, a number of observations for both explanatory variables and the response variable they are interested in. This situation is illustrated in Figure~\ref{scenario} (a). There, the analyst already has the first $K$ rows and may be interested in purchasing additional rows (from the $N-K$ they do not have already). In the second case, which we refer to as \emph{feature market} (the term \emph{regression market} has also been used in the literature, see \citet{pinson2022regression}, for instance), the analyst has access to all $N$ observations for $M$ input features, as well as for the response variable of interest. The analyst may then be interested in purchasing additional columns, i.e., data for additional $P-M$ potential features that they do not already have access to.

An example \revised{of} the observation market is that of an e-commerce site (e.g. Amazon) predicting customer lifetime value, where available observations consist of past purchases, frequency purchases, and customer demographics. The analyst may be interested in purchasing additional data for these same features from another e-commerce site (e.g., Shein). With more comprehensive data available, the analyst may improve the fit of their model(s) and/or improve their generalizability. In parallel, for the feature market, an example is that of two wind farms buying data from each other. Each wind farm will have their own features, e.g., local measurements and weather forecasts, and may be interested in purchasing measurements and weather forecasts from the other wind farm.



In contrast, here we introduce active learning markets (which may be equivalently seen as \emph{label markets}). We define \(D_\text{L}\) the labelled data that the analyst already has access to (including observations for both explanatory and target variables), \(D_\text{U}\) the unlabelled data that is either owned by data sellers or can be obtained (e.g., \revised{through} public sources), and \(D_\text{ML}\) the missing labels that are owned exclusively by data sellers. The analyst is interested in purchasing \emph{some of} these labels. An example active learning market is illustrated in Figure~\ref{scenario}(c), based on the following sets of data:
\vspace{-5mm}
\begin{subequations}{\label{eq:active_learning_dataset}}
\begin{align}
    D_\text{L} &= \left\{ \left( x_{1i}, x_{2i}, \dots, x_{Pi}, y_{i} \right) \, , \, \, i = 1, 2, \dots, K \right\}, \\
    D_\text{U} &= \left\{ \left( x_{1j}, x_{2j}, \dots, x_{Pj} \right) \, , \, \,  j = K+1, \dots, N \right\}, \\
    D_\text{ML} &= \left\{ y_{j} \, , \, \,  j = K+1, \dots, N \right\}.
\end{align}
\end{subequations}
where \(K < N\). Here, the data analyst has \(D_\text{L}\) and \revised{\(D_\text{U}\)} with all the features for all \(N\) observations, while the data sellers have \(D_\text{ML}\) with the target variables for \(N-K\) observations.

\subsection{Active learning market: Participants}

In the active learning market, let \(A\) represent the data analyst and \(S_j\) (\(j = 1, 2, \dots, m\)) the data sellers, where \(m = N-K\) denotes the total number of data sellers and each seller has a unique label \(y_j \in D_\text{ML}\). In the more general case, there may be fewer than $m$ sellers and they may own more than 1 label each. However, this does not change the market setup in any way, and we consider this simpler setup in which each seller has a single label only for simplicity.

The data analyst \(A\) works with several defined components. Note that we use the term data analyst for the data buyer, to be consistent with the terminology used by, e.g., \citet{cummings_accuracy_2015}. Identified data \(D_\text{L}\) comprises the data set owned by the analyst, including both features and their corresponding target values (that is, the corresponding labels). The unlabelled data \(D_\text{U}\) consists of data points that the analyst can access, covering values for all relevant features, but without associated labels for the target variable of interest. The analyst is willing to pay (WTP) $\phi$, which represents what the analyst is ready to spend (per unit \revised{cost}, e.g., £, \$, ¥) for a unit improvement in some model performance metric $L$. For instance, both the variance of model estimates and the mean square error of model-based forecasts will be used in the following. \revised{Here, we treat $\phi$ as a constant, reflecting ex-ante scenarios in which the analyst has no prior information to differentiate among candidate labels and therefore sets a uniform value for performance improvement. This simplification follows the tradition of single-parameter environments in mechanism design, where each participant’s valuation or cost is summarized by a single scalar parameter (e.g., \citet{ghosh2011selling,cummings_truthful_2015,agarwal_marketplace_2019}). Such models provide tractability while still capturing key buyer–seller interactions. In practice, this corresponds to organizations that translate forecast errors into well-defined financial costs (e.g., inventory misallocation or excess capacity), which naturally yield a constant willingness to pay per unit of accuracy. Additionally, our analysis is conducted in the batch active learning setting; therefore, the buyer’s willingness to pay (WTP) is assumed constant over time rather than dynamically updated. The potential variation of WTP in online or adaptive settings is discussed in Section~\ref{sec:future works}.} The analyst also establishes a model performance threshold $\alpha$, which defines the maximum expected decrease in the model performance metric (assuming that the metric is negatively oriented) \revised{for} which they are ready to pay. In parallel, the data purchase budget $B$ specifies the maximum amount the analyst may be able to spend to purchase data.

Data sellers, on the other hand, are characterized by the labels they own, gathered in the set \(D_\text{ML}\), which are made available to the analyst for purchase. These labels are not shared with the analyst prior to the purchase, since active learning will allow one to decide which label the analyst wants to acquire, without having used that label yet. Each data seller has a willingness to sell (WTS), denoted \(\eta_j\), which represents the minimum price each seller is willing to accept for their label $y_j$. 

\subsection{How the market operates}\label{subsec:market_operate}

\revised{Having defined the participants and their attributes, we now formalize the operation of the market. The mechanism must determine both the allocation of labels to be purchased and the corresponding prices. Formally,} the market \revised{operates} as a mapping \( \mathcal{F}^{a}_{L} \), which, based on input data $D_\text{L}, D_\text{U}, D_\text{ML}$ and given the (linear) regression model $\mathcal{M}$ chosen by the analyst and the loss function $l$ of interest, performs both resource allocation (i.e., choosing the labels to be purchased and from which seller) and pricing of these labels, i.e., 
\begin{equation}\label{eq:map_market}
\begin{aligned}
    \mathcal{F}^{a}_{L}: (D_\text{L}, D_\text{U}, D_\text{ML}; \mathcal{M}, l) 
    \rightarrow 
    \big(D^\text{ac}_\text{ML}, \mathbf{p}
    \big)
\end{aligned}
\end{equation}
where $D^\text{ac}_\text{ML} \subseteq D_\text{ML}$ is the set of data eventually purchased and transferred to the analyst, while $\mathbf{p} = \{p_j \, | \, y_j \in D^\text{ac}_\text{ML}\}$ is the set of prices for these labels, used as a basis for defining revenues and payments for both analysts and sellers.

At the core of the resource allocation problem is the label purchase decision. It involves a binary variable \(z_j\), which is used to indicate whether the label is purchased or not. The rule is such that the improvement $l_j$ in the loss function $l$ has to be greater than a threshold defined by the label price and the analyst's willingness to pay, for the label to be purchased. This writes
\begin{equation}\label{eq:x_j}
    z_j = \begin{cases} 
    1, & l_j \geq \dfrac{\eta_j}{\phi} \\
    0, & \text{otherwise}
    \end{cases}
\end{equation}

Regarding the pricing of the labels actually purchased, we consider two alternatives which are to be seen as the most extreme cases. \revised{That is}, the pricing can be based either on the buyer or on the offer from the seller. In a single-buyer-multiple seller setup, one cannot rely on supply-demand equilibrium concepts, as would be the case for more general resource allocation problems with multiple buyers and multiple sellers. Here, in a seller-centric pricing approach (SC), the price is defined as \(p_j = \eta_j\), which corresponds to the willingness to sell for the label $y_j$. Following a buyer-centric (BC) pricing approach, instead, each purchased label $y_j$ is priced as \(p_j = \phi l_j\), i.e., reflecting the willingness to pay $\phi$ and the \revised{perceived} value of the label $l_j$. These two approaches can be seen as extreme cases since labels will only be purchased \revised{if} $\phi l_j \geq \eta_j$, and then any price $p_j \in [\eta_j, \, \phi l_j]$ could be acceptable to both both buyer and the seller. 
As a by-product of market clearing, the analyst can estimate a set of parameters \( \hat{\theta} (D_\text{L}, D_\text{U}, D^\text{ac}_\text{ML}) \) based on their own data, as well as acquired labels, for their regression model. The analyst also deduces the estimate of the overall loss function by acquiring \(D^\text{ac}_\text{ML} \subseteq D_\text{ML}\). We write \(l_j\) the loss reduction induced by acquiring the label $y_j$.

In principle, acquiring the labels \(y_j
\in D^\text{ac}_\text{ML}\) leads to an overall reduction \(\Delta L\) in the loss function of interest. It can be expressed as
\begin{equation}\label{eq:loss_f}
\Delta L = L (D_\text{L}, D_\text{U}; \mathcal{M}, l) - L (D_\text{L}, D_\text{U}, D^\text{ac}_\text{ML}; \mathcal{M}, l)    
\end{equation}
where the first and second terms are the overall loss values before and after acquiring the labels in $D^\text{ac}_\text{ML}$, respectively. In the following, the type of loss functions we will consider will include the variance of parameter estimates and the generalized prediction performance (i.e., Mean Squared error, MSE, on a validation set).

\subsection{Data purchase as an optimisation problem}\label{optimisation}



Based on the data market framework introduced in Section~\ref{subsection:market scenario}, we then formulate our \revised{simplified} label acquisition problem. Consider a data analyst who wishes to purchase labels from a set of sellers to improve some negatively-oriented metric $l$ (the lower, the better) related to their regression model (e.g., variance of parameter estimates or some measure of prediction error). The key challenge is to decide which labels \(y_j \in D_\text{ML}\) should be acquired. Conceptually, assuming that each data point contributes with an additive improvement \(l_j\), we have a combinatorial optimization problem with budget constraints. Indeed, the data analyst has a budget limit \(B\) and an expected threshold \(\alpha\) for the reduction of the loss function. In practical applications, those constraints can be excessively strict, potentially leading to an insufficient number of purchased data points. To mitigate this issue, we relax the constraints as either the budget or loss function threshold is met. \revised{Therefore the simplified setup} can be defined as
\vspace{-3mm}
\begin{subequations}
\label{analyst_problem_relax}
\begin{align}
\min_{\{z_j\}} \quad &\ \sum_{j=1}^{m} \, z_j \, p_j \label{objective function}\\
\text{subject to} \quad 
& \ \tilde{l} = l - \sum_{j=1}^{m} \, z_j \, l_j \label{constraint1} \\
& \ \frac{\eta_j}{l_j} \leq \phi \, , \quad \forall j \label{constraint2} \\
& \ p_j = \begin{cases} 
 \phi \,  l_j & \text{(BC)} \\ 
 \eta_j & \text{(SC)} \end{cases} , \quad \forall j \label{constraint3} \\
& \ \sum_{j=1}^{m} z_j p_j \leq B \perp \ \tilde{l} \leq \alpha \, , \quad \forall j \label{constraint4} \\
& \  z_j \in \{0, 1\} \label{constraint5}
\end{align}
\end{subequations}
where the price $p_j$ \revised{is} defined as $p_j = \phi \,  l_j$ or $p_j = \eta_j$, following the BC and SC pricing approaches, respectively (as expressed in \eqref{constraint3}). 
\revised{Problem~\eqref{analyst_problem_relax} is intentionally presented as a simplified, stylized abstraction of the label-purchase process. Its additive and separable structure assumes that the benefits from purchasing labels can be written as $\sum_j z_j l_j$. This is, of course, a simplification: in real-world  data-purchase settings, the value of data is typically \textit{state-dependent}—the contribution of label $j$ depends on which other labels have already been acquired—and information may be shared across labels. Hence, the effective contribution of label~$j$ is generally a function of $(z_1,\dots,z_j)$, and market participants may behave strategically. The formulation in \eqref{analyst_problem_relax} therefore does not aim to capture the full richness of market dynamics; rather, it serves as an analytically tractable benchmark that highlights the allocation trade-offs that motivate our active-learning framework.}


\revised{Although the simplified structure in \eqref{analyst_problem_relax} superficially resembles a knapsack formulation, applying knapsack algorithms directly is infeasible in our setting. Knapsack methods—exact, approximate, or heuristic \citep{martello1990knapsack, bourdache2019active}—all require that the utilities of all items be known in advance. In our context, these utilities correspond to the realised improvement from labeling each point, which can only be observed after the label has been purchased. This makes classical knapsack approaches unsuitable for strategic and limited label acquisition. Moreover, even if approximate utilities were available, the combinatorial nature of the problem combined with the scale of modern datasets renders exact solvers and approximation schemes computationally burdensome. \minorrev{Nevertheless, to provide a reference baseline, we later report a greedy knapsack-style heuristic (GK) that ranks candidates using ex-ante proxies rather than realised utilities; it is therefore not a direct knapsack solution to \eqref{analyst_problem_relax} and offers no optimality guarantee.
}In contrast, active learning provides a scalable and adaptive alternative: it uses estimators of label utility that can be updated as labels are acquired, without requiring all labels to be observed upfront \citep{cacciarelli_active_2024}. Active learning trades global optimality guarantees for adaptability and computational feasibility, making it more practical for real-world large-scale label acquisition tasks.}

\paragraph{Theoretical intuition under simplifying assumptions.}
Deriving full optimality or regret guarantees for \eqref{analyst_problem_relax} is challenging because the marginal utilities $\{l_j\}$ are \emph{revealed only after purchase} and are \emph{state-dependent} (they vary with the acquired set), so standard analyses that assume a static set function do not directly apply. Nevertheless, partial theoretical insights may be obtainable. For instance, assume that (i) costs are bounded or drawn i.i.d. from a distribution independent of informativeness, i.e. $\eta_j \in [\eta_{\min},\eta_{\max}]$, and (ii) a pre-purchase proxy $\widehat l_j$ is bounded and (approximately) unbiased for the true marginal improvement $l_j$. Under such assumptions, a \emph{proxy-based} cost-weighted greedy AL rule that ranks candidates by $\widehat l_j/\eta_j$ aligns with cost-sensitive greedy query-tree analyses that select by a benefit--cost ratio (e.g., shrinkage-cost ratio) \citep{guillory2009average}, and can be studied via (a) approximation-style guarantees against optimal (or baseline) query trees under simplified models \citep{dasgupta2004analysis}, or (b) myopic (one-step) greedy optimality properties implied by selecting the maximising ratio at each step \citep{guillory2009average}.
Our market setting does not satisfy these static-utility assumptions: both proxies and realised improvements evolve as labels are acquired, and acceptance decisions couple $\eta_j$ and $l_j$ through \eqref{constraint2}--\eqref{constraint4}. Establishing formal bounds would therefore require additional structural conditions (e.g., diminishing returns / adaptive submodularity) under which greedy policies are provably near-optimal \citep{golovin2011adaptive}, which we leave for future work.

\subsection{Market properties}
The proposed active learning market adheres to the following desirable market properties. \\
(i) \textit{Budget Balance} \\
A market is \emph{budget-balanced} if the sum of revenues on the seller side is equal to the sum of payments on the buyer side. In a single-buyer-multiple-seller setup like ours, for both pricing approaches considered, it is straightforward to see that the buyer readily pays the sellers on an individual label basis, \revised{ensuring budget balance by construction in~\eqref{objective function}}. The market is then budget-balanced by design.\\
(ii) \textit{Symmetry} \\
\emph{Symmetry} holds if sellers with identical contributions \( l_j \) and identical willingness-to-sell \( \eta_j \) are treated equivalently. From our definition of prices \revised{in \eqref{constraint3}}, i.e.,
\begin{equation} \label{eq:pricing}
    p_j = \begin{cases}
\phi l_j, & \quad \text{buyer-centric pricing}\\ 
\eta_j, & \quad \text{seller-centric pricing}
\end{cases}
\end{equation}
we can see that if \( l_j = l_j' \) and \( \eta_j = \eta_j' \), then \( p_j = p_j' \). Moreover, the selection decision \revised{in \eqref{constraint5}} depends solely on \( l_j\), \( \eta_j \) and \( \phi \), ensuring that there is no preference in the case of having sellers with identical information and willingness to sell. \revised{ Note that symmetry holds only in the simplified formulation of \eqref{analyst_problem_relax}. In our active learning market, utilities are updated adaptively, so if two labels contain similar information, acquiring one naturally reduces the marginal value of the other.  As a result, exact symmetry does not persist in practice—an expected and even beneficial property, as it discourages spending on labels that add little new information.
We emphasise that the symmetry discussed here is a property of the 
stylised optimisation formulation, not an assumed behavioural property of the active 
learning process itself.}\\

    (iii) \textit{Truthfulness.} \\
    \emph{Truthfulness}, also referred to as \emph{incentive compatibility}, is the property such that, for all agents involved, it is a weakly dominant strategy to truthfully reveal their preferences through the market mechanism. \revised{Our setting does not aim to fully model strategic behaviour; rather, we use
“truthfulness’’ in an operational sense: under the simple market rules we adopt,
deviating from truthful reporting does not increase a participant's expected payoff.} Here, this would translate into sellers providing their true labels while revealing their true willingness to sell. On the buyer's side, this would translate to truthfully revealing their willingness to pay. This property should hold regardless of the pricing approach chosen \revised{in \eqref{constraint3}} (buyer-centric or seller-centric).
\begin{revisedblock}

On the buyer side, reporting the truthful willingness to pay (WTP) is operationally optimal. If the
analyst understates their WTP, they may fail to acquire labels
that are quite informative, thereby reducing overall model
performance. Conversely, overstating the WTP may result in purchasing labels
whose value does not justify the payment, making the purchase strictly
suboptimal. 

On the seller side, changing the WTS cannot increase the seller’s expected payoff. This is because, under both pricing mechanisms, a label must
first pass the market’s selection criterion in \eqref{constraint2} before it can be purchased. Increasing the
WTS $\eta_j$ always makes it harder for the label to satisfy this criterion, and therefore directly lowers the probability of being selected. Decreasing the WTS does not increase the payment received. Hence, even before considering the
pricing rule, manipulating the WTS cannot improve expected revenue because it
interferes with the basic eligibility to be purchased. Even if one imagines removing the selection criterion entirely, increasing the
WTS still does not benefit the seller. In principle, a seller might try to raise
the WTS up to the buyer’s budget threshold \(B\), since this is the maximum amount
that the buyer could ever pay. However, the seller does not know the buyer’s budget:
it is internal information available only to the analyst. Therefore, in both BC and SC,
manipulating the WTS cannot improve the seller’s expected payoff. Another consideration is on data quality - if the data provided is not true, then the probability of the modified data being purchased is lower than the 
probability of purchasing the true data, therefore the expected utility is lower as well.

For these reasons, and within the scope of our simplified, non-strategic market
structure, truthful revelation is operationally advantageous for both buyers and
sellers.

\end{revisedblock}

(iv) \textit{Individual Rationality} \\
Individual rationality is satisfied if all sellers entering the market are not in a potential loss-making position. In view of the definition of our pricing approaches, constraints~\eqref{constraint2} and ~\eqref{constraint3} ensure that the payment will only be such that \(p_j \geq \eta_j\). On the analyst side, while labels are only purchased if they lead to a loss reduction, the constraint~\eqref{constraint4} ensures that they achieve a desirable level of loss reduction within the budget constraint.

(v) \textit{Zero-Element}. \\
The zero-element property requires that no transaction should occur when a seller does not provide a data point or provides a data point with no value (i.e., \ $l_j = 0$). \revised{In the relaxed optimisation problem~\eqref{analyst_problem_relax}, this property holds automatically: if $l_j = 0$, the point neither contributes to the model-improvement requirement nor lowers the analyst’s total cost. Hence the optimal solution sets $z_j = 0$, and the corresponding payment satisfies}
\begin{equation}
    z_j p_j = 0, 
    \qquad \forall j \colon l_j = 0.
\end{equation}

\revised{In contrast, the behaviour differs once pricing rules are applied in the algorithms. Under buyer-centric (BC) pricing, the payment rule $p_j=\phi l_j$ ensures $p_j = 0$ whenever $l_j = 0$, so the zero-element condition is operationally true. However, in seller-centric (SC) pricing, a label must be acquired before its improvement can be evaluated, and the price $p_j=\eta_j$ applies to any acquired label irrespective of $l_j$. Thus, while the zero-element property holds cleanly at the level of the relaxed optimisation problem, it cannot be enforced at the payment level under SC pricing. Later, in Section~\ref{Subs: strategy} and \ref{real-world applcations}, we show how active learning strategies reduce the \emph{frequency} with which non-improving points are queried in practice, thereby mitigating the impact of this limitation under SC pricing.}

\section{Methodology} 
\label{methodology}

Solving \eqref{analyst_problem_relax} is not tractable in practice, in view of the combinatorial nature of the optimisation problem. In contrast, here, we introduce an active learning framework that allows one to operationalize data purchasing under budget constraints. It should be seen as a greedy approach to solving the original problem. The existing literature on active learning often overlooks transactional frictions and assumes access to all labels without cost (or at the same cost). Our framework fills this methodological gap by explicitly modelling the acquisition of labels as a market interaction, where data is traded between sellers and a data analyst constrained by a budget and incentivized by model performance gains. This section proceeds by describing the modelling structure, evaluation metrics, and the specific learning strategies deployed.

\subsection{Modelling framework}
\begin{figure}[!htbp]
    \centering
    \includegraphics[width=\linewidth]{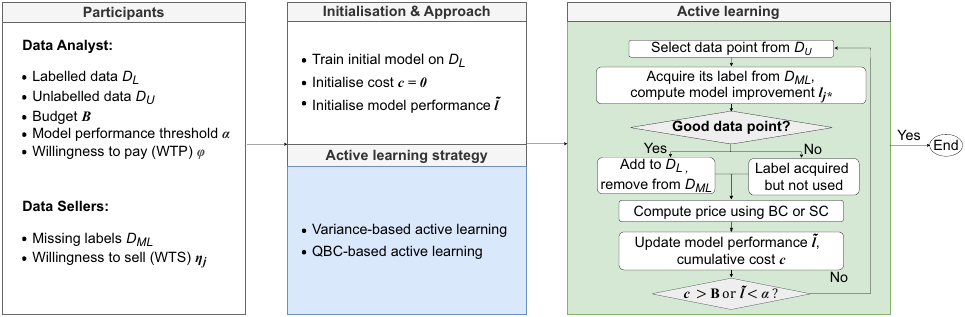}
    \caption{\revised{Overview of the active learning market. Random Sampling Corrected (RSC) serves as the baseline method, in which data points are selected randomly rather than through active learning and are purchased only if their labels yield positive model improvement. This baseline is omitted from the diagram for clarity.}}
    \label{fig:market_overview}
    
\end{figure}

To simulate this process, we model a dynamic market environment, visualized in Figure~\ref{fig:market_overview}, where a data analyst begins with a small labelled dataset \( D_\text{L} \) and a large pool of unlabelled data \( D_\text{U} \) while data providers possess missing labels \( D_\text{ML} \). At each iteration, the analyst computes model performance based on the current labelled data \( D_\text{L} \) and evaluates whether the acquisition of an additional data point is expected to reduce the loss \( l \) of the model. A purchase decision is made only if the marginal loss reduction \( l_j \), achieved by purchasing a data point \( y_j \in D_\text{ML} \), justifies its cost, subject to a budget constraint \( B \). Data points are integrated into the model only if their contribution is positive and cost-effective, and this process continues until the budget is exhausted or a predefined performance threshold \( \alpha \) is met.

\subsection{Quality metrics}

\subsubsection{Focus on estimation quality}

In many applications, accurate parameter estimation is essential for interpreting relationships between input features and target outcomes. In fields such as real estate or pharmaceuticals, regression coefficients serve as interpretable measures—indicating, for example, how location or size affects property prices, or how dosage influences drug efficacy. In such contexts, the focus is on \emph{in-sample} model fit and the stability of estimated parameters.

We adopt the linear regression model as the foundation.
\begin{revisedblock}
  \begin{equation} \label{eq:linear model}
        Y_k = \mathbf{x}_k^\top \beta + \epsilon_k, \quad \forall k,
    \end{equation}  
\end{revisedblock}
where \(Y_k\) is the response variable, \revised{\(\mathbf{x}_k\) is the vector of input features for sample \(k\), 
\(\beta\) is the coefficient vector,} and \(\epsilon_k\) denote the error term. Under standard assumptions, the variance of the estimated parameters is:
\begin{equation} \label{eq:parameter variance}
    \text{Var}[\beta] = (\mathbf{X}^\top \mathbf{X})^{-1} \, \sigma_Y^2
\end{equation}
which highlights the role of data geometry—via the information matrix \( \mathbf{X}^\top \mathbf{X} \)—in determining estimation precision. In the above, $\sigma_Y^2$ is the variance of the response variable $Y$. Lower parameter variance implies greater model stability and more reliable interpretation.

\subsubsection{Focus on predictive ability}
\label{MSE scenario}

In forecasting-oriented domains such as energy demand prediction or financial modelling, model performance is primarily assessed by its ability to generalize to unseen data. Rather than focusing on parameter interpretability, these applications prioritize minimizing out-of-sample prediction error. To quantify this, the mean squared error (MSE) is employed as the primary evaluation metric.

In this study, we use a validation-based evaluation approach. The model is trained on an initially labelled dataset, and candidate data points are acquired from an unlabelled pool using various active learning strategies. Model performance is evaluated on a separate, held-out validation set that simulates future or independent data. This setup reflects practical decision-making scenarios, in which the effectiveness of data acquisition must be measured against a real-world prediction task.

Formally, given a linear regression model as defined in~\eqref{eq:linear model}, the predictions \( \hat{Y}_k \) are evaluated using:
\begin{equation} \label{eq:MSE}
    L = \frac{1}{N} \sum_{k=1}^N (Y_k - \hat{Y}_k)^2,
\end{equation}
i.e., with a Mean Squared Error (MSE) criterion, where \( Y_k \) and \( \hat{Y}_k \) denote the true and predicted values in the validation set (with $N$ forecast-observation pairs). The goal of each acquisition strategy is to improve the model performance in this validation set while respecting budget constraints.





\subsection{Data purchase strategies}\label{Subs: strategy}

The data purchase problem addressed in this paper is inherently constrained by budget and governed by heterogeneous pricing. \revised{Active learning, a framework designed to select the most informative data point under label scarcity, provides a principled foundation for addressing such challenges \citep{zheng2006selectively, zheng2002active}.} In particular, active learning allows the analyst to determine which data points should be acquired to maximize model improvement while minimizing redundant or uninformative acquisitions. Active learning strategies are typically divided into two families: optimization-based and heuristic-based approaches \citep{cohn1996active, tong2001active, settles2011theories}. Optimization-based methods rely on objective functions that formalize notions of informativeness, such as uncertainty reduction or expected loss. In contrast, heuristic-based approaches draw on approximations or model ensembles to identify uncertain points. In our study, we focus on one strategy from each category: variance-based active learning (VBAL), rooted in optimal experimental design, and Query-by-Committee active learning (QBCAL), a heuristic method based on committee disagreement. \minorrev{As benchmarks, we include RSC and a \textit{Greedy Knapsack} (GK) baseline that greedily purchases the lowest-cost candidates by ranking them using $1/\eta_j$, since $l_j$ is not observable ex ante.}

\subsubsection{Variance-Based Active Learning (VBAL)} \label{sec:VBAL}

Variance-based active learning (VBAL) \revised{is based} in Optimal Experimental Design (OED), which provides criteria for selecting data points that most effectively reduce model uncertainty in regression tasks \citep{lu2024data, atkinson1996usefulness, atkinson2001one, atkinson2007optimum}. \revised{A central quantity in OED is the information matrix \( \mathbf{X}^\top \mathbf{X} \),
} derived from the feature \revised{\(\mathbf{X}\)} of the labelled dataset \citep{zheng2006selectively}. Several classical criteria operate on this matrix to guide selection: D-optimality seeks to minimize the volume of the confidence ellipsoid around \revised{the} regression coefficients by maximizing \revised{\( \det(\mathbf{X}^\top \mathbf{X}) \)}; The A-optimality minimizes the average variance of the coefficient estimates by minimizing \revised{\( \text{tr}(\mathbf{X}^\top \mathbf{X}) ^{-1}\)}; and the G-optimality focuses on predictive reliability, aiming to minimize the maximum prediction variance in the design space. While D and A-optimality optimize parameter estimation globally, G-optimality is more aligned with our goal: ensuring that predictions are uniformly reliable across all regions of the input space. In formal terms, G-optimality minimizes the worst-case prediction variance, defined as

\begin{revisedblock}
    \begin{equation} \label{g-optimality}
\min_{\mathbf{X}} \max_{x \in \mathcal{X}} x^\top (\mathbf{X}^\top \mathbf{X})^{-1} x
\end{equation}
\revised{
where \(\mathbf{X}\in\mathbb{R}^{n\times p}\) stacks the currently labelled covariates as rows; 
\(\mathcal X\subset\mathbb{R}^p\) is the candidate design space; 
and \(x\in\mathcal X\) is a column vector. 
We assume \(\mathbf{X}\) has full column rank so that \(\mathbf{X}^\top \mathbf{X}\) is invertible; when this matrix is ill-conditioned early in the process, we replace 
\((\mathbf{X}^\top \mathbf{X})^{-1}\) with \((\mathbf{X}^\top \mathbf{X}+\lambda I)^{-1}\) for a small \(\lambda>0\).
}

\end{revisedblock}

However, in our setting, \revised{instead of} optimizing \revised{the entire} design, we operate in a sequential and budget-constrained environment where data is acquired one point at a time. Therefore, we adopt a greedy approximation of G-optimality by selecting, at each step, the candidate data point with the highest unscaled prediction variance (UPV):
\begin{equation}\label{eq:upv}
x^* = \arg\max_{x \in D_\text{U}} x^\top \revised{(\mathbf{X}^\top \mathbf{X})}^{-1}x
\end{equation}

This UPV score reflects the epistemic uncertainty \revised{of the model} on each candidate and acts as a local pointwise surrogate for G-optimality. Though it does not yield a globally optimal design, it is well-suited for iterative acquisition under limited budget and incomplete information.

\revised{In both active-learning strategies and the RSC baseline, pricing follows an ex-post mechanism. The analyst first proposes to purchase a candidate point and, after acquiring its label, temporarily augments the model with this labelled point for evaluation purposes. This allows her to compute the realised improvement $l_j$. Only after observing $l_j$ is the price $p_j$ determined—e.g., via constraint~\eqref{constraint3} under the buyer-centric scheme. Thus, she proposes a purchase, inspects the actual contribution of the point, and then pays a price directly tied to its realised value. This ex-post pricing structure is standard in many practical settings (e.g., value-based pricing, performance-contingent payments) and naturally integrates into our label purchase problem here. Eventually, the detailed approach to implementing VBAL is described in Algorithm~\ref{algorithm: VBAL}. }

\begin{algorithm}[!ht]
\caption{ Variance-Based Active Learning (VBAL) Algorithm} \label{algorithm: VBAL}
\begin{algorithmic}[1]
\begin{revisedblock}
\small

    \State \textbf{Input:}
    \Statex \hspace{\algorithmicindent} Labelled data: \(D_\text{L} = \left\{ \left( x_{1i}, x_{2i}, \dots, x_{Pi}, y_{i} \right) \mid i = 1, 2, \dots, K \right\} \) 
    \Statex \hspace{\algorithmicindent}  Unlabelled data: \(D_\text{U} = \left\{ \left( x_{1j}, x_{2j}, \dots, x_{Pj} \right) \mid j = K+1, \dots, N \right\} \)
    \Statex \hspace{\algorithmicindent} Missing labels: \(D_\text{ML} = \left\{ y_{j} \mid j = K+1, \dots, N \right\} \)
    \Statex \hspace{\algorithmicindent} Budget: \( B \)
    \Statex \hspace{\algorithmicindent} Model performance threshold: \( \alpha \)
    \Statex \hspace{\algorithmicindent} Willingness to pay: \( \phi \)
    \Statex \hspace{\algorithmicindent} Willingness to sell: \( \eta_j \)

\State \textbf{Output:} Updated \( D_\text{L} \) and reduced \( D_\text{ML} \)
    \State Initialise total cost $c \gets 0$, purchase number $t \gets 0$ and current loss $\tilde{l} \gets \text{Loss}(\text{Model}(D_\text{L}))$
\While{$c < B$ \textbf{and} $\tilde{l} > \alpha$}

    \State Select $x_{j^*}$ according to the prediction–variance criterion (using~\eqref{eq:upv})

            \State Acquire $y_{j^*
            }$, $t \gets t + 1$         \Comment{The label must be purchased first to compute the loss reduction}

    \State Temporarily compute the loss reduction:
    \[
  l_{j^*} = \tilde{l} - \mathrm{Loss}\!\bigl(\text{Model}(D_{\mathrm{L}} \cup \{(x_{j^*},y_{j^*})\})\bigr)
    \]

        \If{\(l_{j^*} > 0\) \textbf{and} \(\eta_{j^*}/l_{j^*} \le \phi\)} \Comment{good data point}
  \State $D_\text{L} \gets D_\text{L} \cup \{(x_{j^*}, y_{j^*})\}$,
         $D_\text{ML} \gets D_\text{ML} \setminus \{y_{j^*}\}$
  \State Compute price:
  \[
    p_{j^*} =
    \begin{cases}
      \phi l_{j^*} & \text{(Buyer-centric)}\\
      \eta_{j^*} & \text{(Seller-centric)}
    \end{cases}
  \]
  \State $\tilde{l} \gets \tilde{l} - l_{j^*}$
\Else \Comment{bad data point}
  \State $D_\text{ML} \gets D_\text{ML} \setminus \{y_{j^*}\}$ \Comment{label acquired but never used}
  \State Compute price:
  \[
    p_{j^*} =
    \begin{cases}
      0 & \text{(Buyer-centric)}\\
      \eta_{j^*} & \text{(Seller-centric)}
    \end{cases}
  \]
\EndIf

\State $c \gets c + p_{j^*}$ \Comment{Update total cost}

\EndWhile
\end{revisedblock}
\end{algorithmic}
\end{algorithm}

\subsubsection{Query-by-Committee-Based Active Learning (QBCAL)} \label{sec:QBCAL}

In addition to VBAL, we employ Query-by-Committee-based active learning (QBCAL) as a robust alternative for data acquisition under uncertainty. QBCAL is particularly well-suited to settings in which a single model may provide unreliable or overconfident uncertainty estimates, especially in the presence of sparse labelled data, model misspecification, or complex, multimodal feature spaces. By leveraging model disagreement as a proxy for epistemic uncertainty, QBCAL enables the identification of data points where the current labelled set provides insufficient or inconsistent information.

Classical QBCAL approaches in classification typically fall into two categories: \emph{vote entropy} \citep{freund1997selective, argamon1999committee}, which quantifies disagreement among classifiers by evaluating the entropy of predicted labels, and \emph{margin confidence} \citep{abe1998query, settles2011theories}, which prioritizes points with the smallest difference between the most probable class labels. Although effective in classification tasks, these strategies are not directly applicable in regression settings where the output is continuous rather than categorical.

To adapt QBCAL to our regression-based context, we adopt a variance-based formulation that directly measures prediction dispersion. At each iteration, a model committee \( M = \{M_1, M_2, \dots, M_m\} \) is trained using bootstrap replicates of the current labelled dataset \( D_\text{L} \). For each candidate point \( x_j \in D_\text{U} \), the predictive uncertainty is quantified by the variance of the committee output:
\begin{equation} \label{eq:VBAL}
\text{Var}(\hat{y}(x_j)) = \frac{1}{M} \sum_{m=1}^{M} \left( \hat{y}_m(x_j) - \bar{\hat{y}}(x_j) \right)^2
\end{equation}

where \( \hat{y}_m(x_j) \) is the prediction of the model \( M_m \) and \( \bar{\hat{y}}(x_j) \) is the mean prediction of all members of the committee. The candidate with the highest predictive variance is selected as the most informative. As in VBAL, the analyst estimates the expected loss reduction \( l_j \) and evaluates whether the point satisfies the cost-efficiency constraint given in~\eqref{constraint2}. If the expected benefit is positive and the acquisition cost is within budgetary limits, the point is added to the labelled dataset, the model is updated, and the budget is adjusted. This process continues until the performance threshold \( \alpha \) is achieved or the available budget \( B \) is fully consumed. The detailed version of QBCAL is in Algorithm~\ref{algorithm:QBCAL}. 


\begin{algorithm}[!ht]
\caption{Query-by-Committee Active Learning (QBCAL)} \label{algorithm:QBCAL}
\begin{algorithmic}[1]
\begin{revisedblock}
\small
\State \textbf{Input:} Labelled data \( D_\text{L} \), Unlabelled data \( D_\text{U} \), Missing labels \( D_\text{ML} \), Budget \( B \), Model performance threshold \( \alpha \), WTP \( \phi \), WTS \( \eta_j \)
\State \textbf{Output:} Updated \( D_\text{L} \), reduced \( D_\text{ML} \)
    \State Initialise total cost $c \gets 0$, purchase number $t \gets 0$ and current loss $\tilde{l} \gets \text{Loss}(\text{Model}(D_\text{L}))$
\While{\( c < B \) and \( \tilde{l} > \alpha \)}
    \State Train a committee of models on \( D_\text{L} \) using bootstrap sampling
    \For{each \( x_j \in D_\text{U} \)}
        \State Compute predictive variance \( \text{Var}(\hat{y}(x_j)) \) using~\eqref{eq:VBAL} \Comment{Selection criterion}
    \EndFor

    \State Select the most ambiguous point:
    \[
        x_{j^*} = \arg\max_{x_j \in D_{\mathrm{U}}}\mathrm{Var}(\hat{y}(x_j))
    \]

\State Acquire the label $y_{j^*}$, $t \gets t + 1$ and temporarily compute the loss reduction
\[
  l_{j^*} = \tilde{l} - \mathrm{Loss}\!\bigl(\text{Model}(D_{\mathrm{L}} \cup \{(x_{j^*},y_{j^*})\})\bigr)
\]

\State Compute the price $p_{j^*}$ \emph{as in Algorithm~\ref{algorithm: VBAL}}

\If{$l_{j^*} > 0$ \textbf{and} $\eta_{j^*}/l_{j^*} \le \phi$}
  \State $D_{\mathrm{L}} \gets D_{\mathrm{L}} \cup \{(x_{j^*}, y_{j^*})\}$,
         $\tilde{l} \gets \mathrm{Loss}(\text{Model}(D_{\mathrm{L}}))$
\EndIf
\State $D_\text{ML} \gets D_\text{ML} \setminus \{y_{j^*}\}$ \Comment{label is acquired in all cases}
\State $c \gets c + p_{j^*}$

\EndWhile
\end{revisedblock}

\end{algorithmic}
\end{algorithm}

\subsubsection{Random Sampling Corrected (RSC) strategy}
The Random Sampling Corrected (RSC) strategy, formally detailed in Algorithm~\ref{algorithm:RSC}, serves as a baseline in our analysis. It allows us to evaluate the benefits of more sophisticated active learning strategies against a simple, non-informed approach that still incorporates basic cost awareness. The method is based on traditional random sampling, but is corrected to avoid unproductive data acquisitions. At each iteration, the data analyst randomly selects a data point from the pool of unlabelled data with missing labels, denoted \( D_\text{ML} \). This selected data point is evaluated by estimating the potential reduction in model loss \( l_j \) that would result from incorporating it into the labelled dataset \( D_\text{L} \). If this estimated loss reduction is positive—i.e., if the data point is expected to improve the model, and the cost-effectiveness of this data point satisfies the constraint given in~\eqref{constraint2}, then the data point is purchased. Upon purchase, the data point is removed from \( D_\text{ML} \), added to \( D_\text{L} \), and the model is retrained with the augmented dataset. The total cost \( c \) is updated by adding the price of the newly acquired data point. This process repeats until the budget is exhausted or the model reaches the desired performance level. While RSC does not leverage structure in the feature space or model uncertainty, it avoids purchasing uninformative or harmful data points, making it a reasonable benchmark for evaluating active learning methods.

\begin{algorithm}[!ht]
\small
\caption{Random Sampling Corrected (RSC) Algorithm} \label{algorithm:RSC}
\begin{algorithmic}[1]
\small
\begin{revisedblock}
    \State \textbf{Input:} Labelled data $D_\text{L}$, Unlabelled data $D_\text{U}$, Missing labels $D_\text{ML}$, Budget $B$, Model performance threshold \( \alpha \), WTP $\phi$, WTS $\eta_j$
    \State \textbf{Output:} Updated $D_\text{L}$ and reduced $D_\text{ML}$

    \State Initialise total cost $c \gets 0$, purchase number $t \gets 0$ and current loss $\tilde{l} \gets \text{Loss}(\text{Model}(D_\text{L}))$

    \While{$c < B$ \textbf{and} $\tilde{l} > \alpha$}

        \State Randomly select an index $j$ with $x_j \in D_\text{U}$ and $y_j \in D_\text{ML}$
        \State Acquire $y_j$, $t \gets t + 1$ \Comment{number of purchased data points}
 \State Temporarily compute the loss reduction
        \[
  l_j = \tilde{l} - \text{Loss}\bigl(\text{Model}(D_\text{L} \cup \{(x_j, y_j)\})\bigr)
        \]
        \State Set the price
        \[
          p_j =
          \begin{cases}
            \phi\, l_j & ( l_j > 0, \text{Buyer-centric}) \\[4pt]
            0 &  (l_j \leq 0, \text{Buyer-centric})\\[4pt]
            \eta_j & (\forall l_j, \text{Seller-centric})
          \end{cases}
        \]

        \If{$l_j > 0$}
        \State $D_\text{L} \gets D_\text{L} \cup \{(x_j, y_j)\}$, $\tilde{l} \gets \text{Loss}(\text{Model}(D_\text{L}))$
        \EndIf

        \State $D_\text{ML} \gets D_\text{ML} \setminus \{y_j\}$ \Comment{label is acquired in all cases}
        \State $c \gets c + p_j$

    \EndWhile
    \end{revisedblock}

\end{algorithmic}
\end{algorithm}

\section{Real-world applications} \label{real-world applcations}

Our proposed active learning markets can be applied to broad scenarios, either to \revised{improve estimation quality, or to improve predictive ability.} Therefore, here we focus on the real estate data set and the energy building data set, respectively. Under each dataset, we use active learning as our data purchase strategy to show how our strategy can (i) improve cost-efficiency in decision making, (ii) benefit both the analyst and the sellers, and (iii) ensure robustness in real-world applications.


\subsection{Focus on estimation quality}\label{est_quality}

\subsubsection{Dataset and case-study}
We used a public real estate data set from Sindian District, New Taipei City \citep{yeh2021realestate}, containing key transaction features (date, age, MRT distance, convenience stores) and unit-area price. A linear regression model is used to explain the variation in prices.

\subsubsection{What are the effects of applying active learning?}\label{sec:var_scen_cost_efficiency}
\begin{figure}[!ht]
    \subfloat[\minorrev{Model improvement}]{ \label{fig:V_BC_general-a}
    \includegraphics[width=0.33\linewidth]{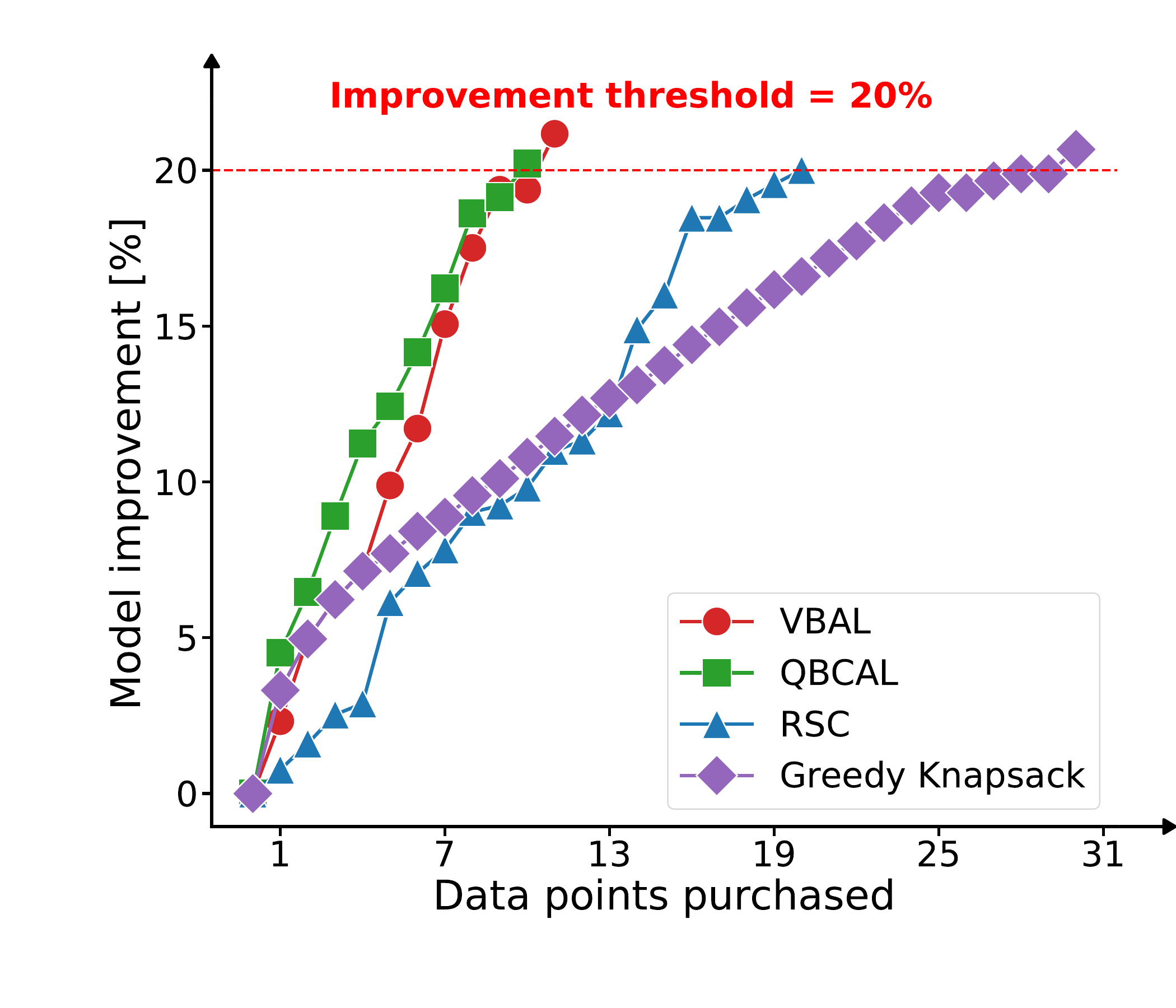}
    }
    \subfloat[\minorrev{Cumulative budget-BC}]{
        \label{fig:V_BC_general-b}
    \includegraphics[width=0.33\linewidth]{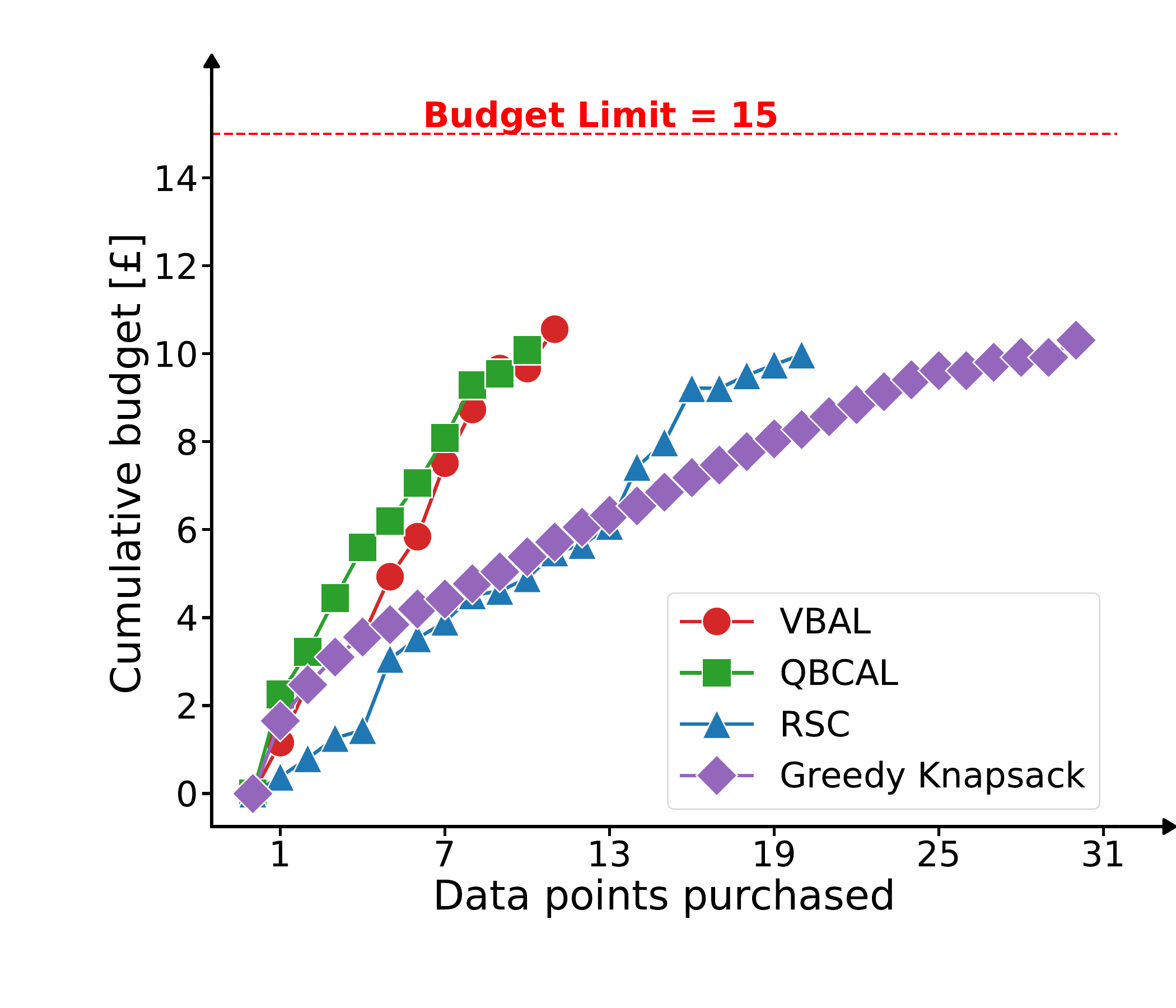} 
    }
        \subfloat[\minorrev{Cumulative budget-SC}]{\includegraphics[width=.33\linewidth]{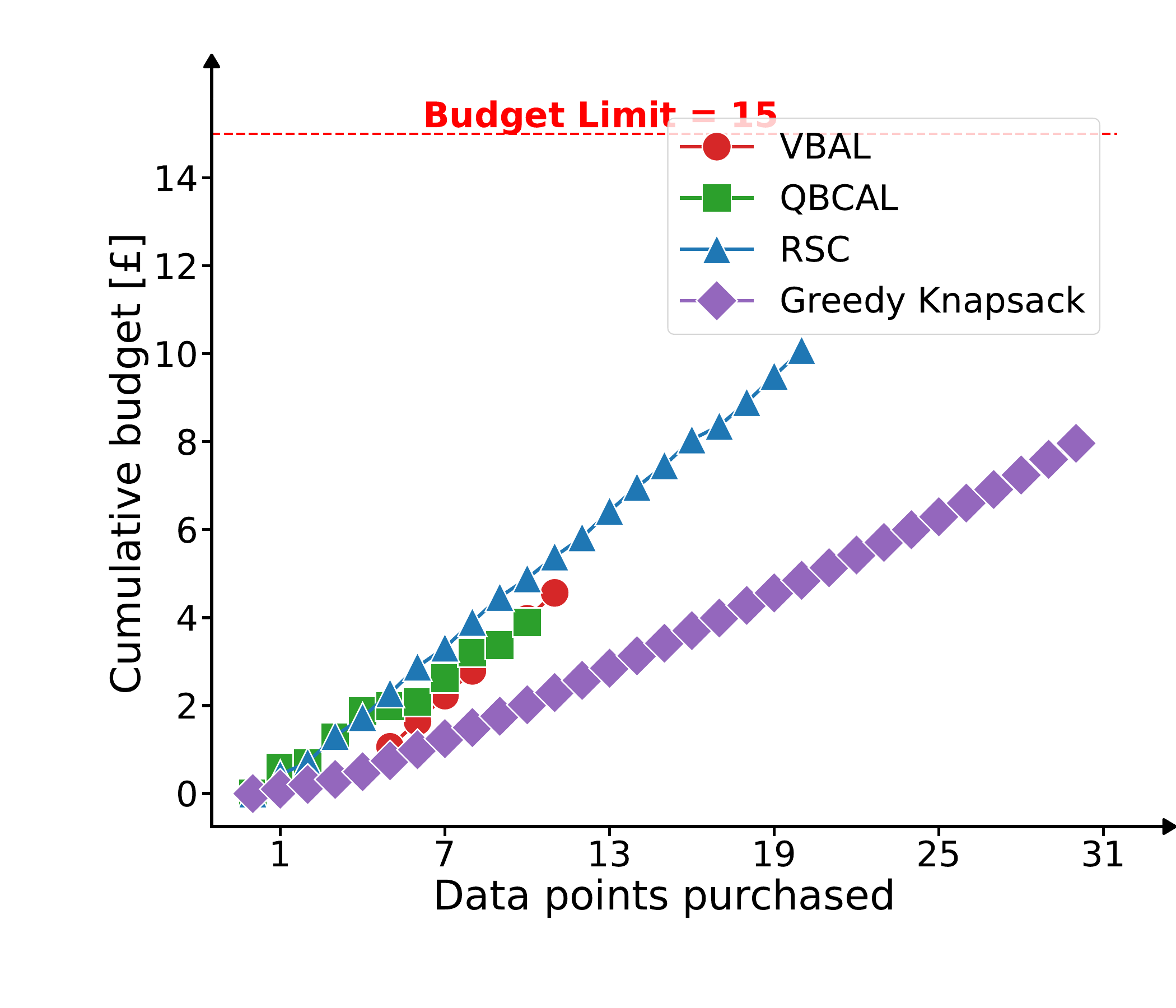}
    \label{fig:V_SC_general-b}}

    \vspace{3mm}
\caption{Different pricing approach (the starting point is not purchasing any data point)}
 \label{fig:V_BC_general}
\end{figure}

We set the WTP at $\phi=1200$ (£/$\Delta$), budget at \(B=\text{£}15\), and a target improvement of 20\%.  
\revised{To introduce controlled heterogeneity on the seller side, the WTS $\eta_j$ is generated synthetically from the ``distance to MRT'' feature:}
\[
\eta_j = d_0 + d_1\left(1 - \frac{X_{3j} - \underline{X_3}}{\overline{X_3} - \underline{X_3}}\right),
\]
with $d_0=0.1$, $d_1=0.5$, and $[\underline{X_3}, \overline{X_3}]$ the observed range.  
\revised{This serves only to induce heterogeneous seller costs (e.g., privacy, effort, reluctance) in a controlled manner; it does \emph{not} imply that property characteristics determine economic WTS.} Prices follow~\eqref{constraint3}, with all other parameters identical in BC and SC.  
\minorrev{Under BC (Figure~\ref{fig:V_BC_general-b}), VBAL reaches higher improvement (21.17\%) than RSC (20.01\%) and GK (20.67\%) at similar cost ($\approx£10$). Under SC (Figure~\ref{fig:V_SC_general-b}), both VBAL and QBCAL improve faster while costing far less (£4.56–£3.89 vs £10.08-£7.96). A summary is provided in Table~\ref{tab:comparison1}.}

\paragraph{Statistical validation.}
To assess whether the differences observed in Table~\ref{tab:comparison1} are statistically significant, we applied the non-parametric \textbf{Wilcoxon signed-rank test} on \emph{improvement per purchased point} (\%/point), with significance level $p=0.05$.
The test compares paired outcomes across 50 random splits for each method under both buyer-centric (BC) and seller-centric (SC) pricing.
Table~\ref{tab:pairwise_avgcost} reports Wilcoxon test results for improvement per purchased point. Under BC pricing, VBAL significantly outperforms the RSC baseline, while QBCAL vs RSC is not statistically significant. Against GK, both VBAL and QBCAL achieve significantly higher per-point improvement under both BC and SC pricing, indicating that GK’s low-price selection does not translate into stronger improvement per purchased point.

\begin{table}[h!]
\centering
\small

\caption{Comparison of Data Purchasing Strategies: Estimation-quality-focused scenario}
\label{tab:comparison1}
\begin{tabularx}{\textwidth}{lccccccc}
\toprule
\textbf{Strategy} & \textbf{Approach} & \textbf{\(B\) exhausted?} & \textbf{\(\alpha\) met?} &
\textbf{Purchased} & \textbf{Spent (£)} & \textbf{Improvement (\%)} & \textbf{Avg. Impro.
(\%/point)} \\
\midrule
\multirow{2}{*}{VBAL}
  & BC & \multirow{2}{*}{No} & \multirow{2}{*}{Yes} & \multirow{2}{*}{11}
  & 10.56 & \multirow{2}{*}{21.17} & \multirow{2}{*}{1.92} \\
  & SC &  &  &
  & 4.56  &  &  \\
\midrule
\multirow{2}{*}{QBCAL}
  & BC & \multirow{2}{*}{No} & \multirow{2}{*}{Yes} & \multirow{2}{*}{10}
  & 10.08 & \multirow{2}{*}{20.22} & \multirow{2}{*}{2.02} \\
  & SC &  &  &
  & 3.89  &  &  \\
\midrule
\multirow{2}{*}{RSC}
  & BC & \multirow{2}{*}{No} & \multirow{2}{*}{Yes} & \multirow{2}{*}{20}
  & 9.98 & \multirow{2}{*}{20.01} & \multirow{2}{*}{1.00} \\
  & SC &  &  &
  & 10.08 &  &  \\
\midrule
\multirow{2}{*}{GK}
  & BC & \multirow{2}{*}{No} & \multirow{2}{*}{Yes} & \multirow{2}{*}{30}
  & 10.31 & \multirow{2}{*}{20.67} & \multirow{2}{*}{0.69} \\
  & SC &  &  &
  & 7.96  &  &  \\
\bottomrule
\end{tabularx}

\end{table}

\begin{table}[ht]
\centering
\caption{Wilcoxon signed-rank test comparing improvement per purchased point (estimation-quality-focused scenario)}
\label{tab:pairwise_avgcost}
\begin{tabular}{llcc}
\toprule
\textbf{Approach} & \textbf{Comparison} & \textbf{$p$-value} & \textbf{Median $\Delta$ (\%/point)} \\
\midrule
BC & VBAL vs RSC            & $3.349 \times 10^{-8}$  & $0.7748$ \\
BC & QBCAL vs RSC           & $1.28 \times 10^{-1}$   & $0.1988$ \\
BC & VBAL vs GK             & $9.521 \times 10^{-13}$ & $1.0428$ \\
BC & QBCAL vs GK            & $1.045 \times 10^{-5}$  & $0.4370$ \\
SC & VBAL vs RSC            & $3.706 \times 10^{-9}$  & $0.7748$ \\
SC & QBCAL vs RSC           & $5.547 \times 10^{-2}$  & $0.1988$ \\
SC & VBAL vs GK             & $5.862 \times 10^{-14}$ & $1.0428$ \\
SC & QBCAL vs GK            & $9.364 \times 10^{-7}$  & $0.4370$ \\
\bottomrule
\end{tabular}
\end{table}

\subsubsection{Does active learning improve cost-efficiency for analysts?} \label{var:costeff}
\begin{figure}[!ht]
    \centering
    \includegraphics[width=0.42\linewidth]{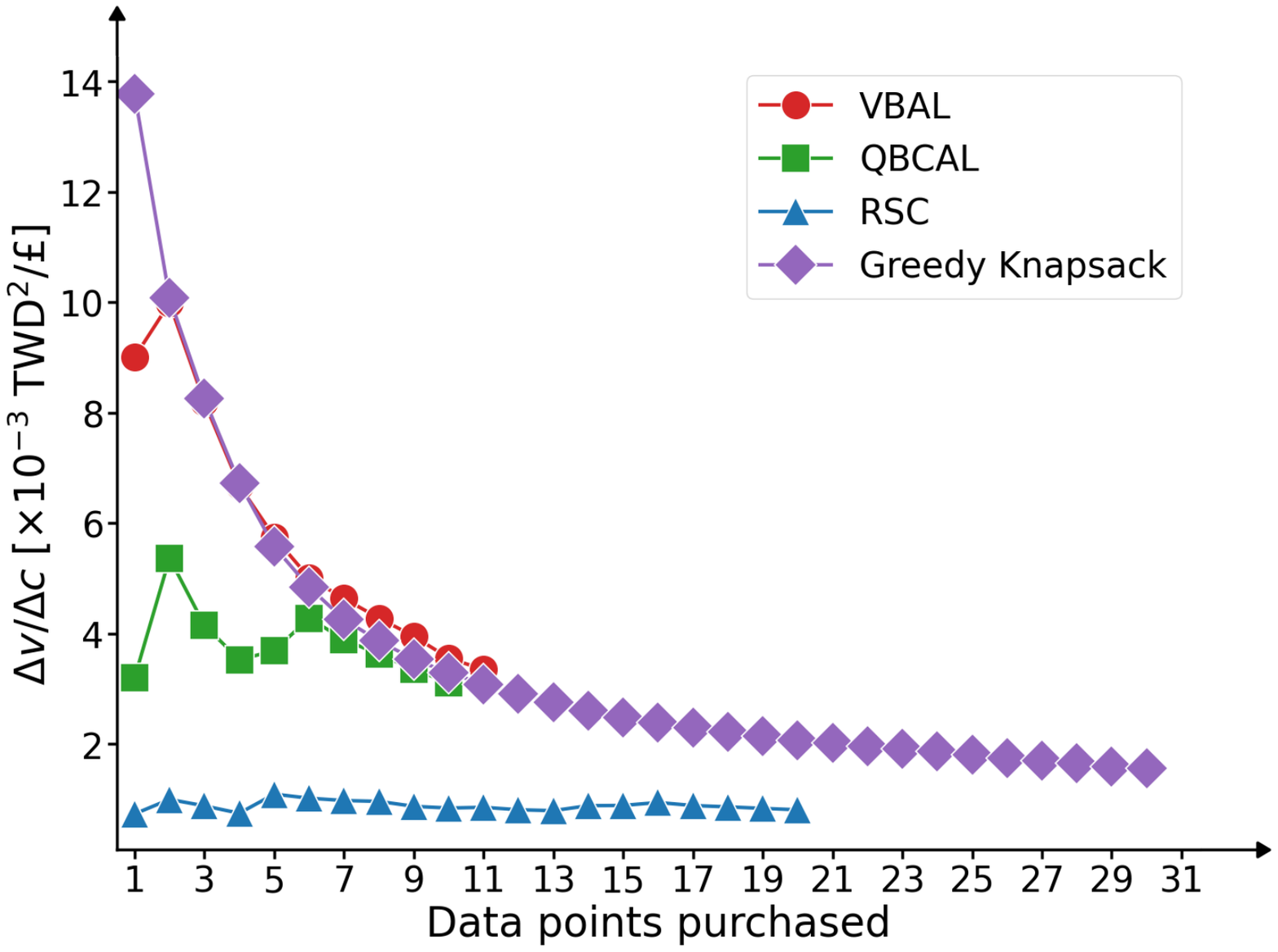}
    \caption{\minorrev{Analyst's side analysis on SC approach: variance reduction per unit cost. Results are represented as cumulative averages (i.e., as the average up to that data point purchased) }}
    \label{fig:V_SC_seller}
\end{figure}

    Beyond the average improvements reported in Table \ref{tab:comparison1}, a more revealing picture arises when we track how cost efficiency evolves from the analyst’s perspective. \minorrev{Figure~\ref{fig:V_SC_seller} illustrates the cumulative average of the cost-efficiency ratio, $\Delta v / \Delta c$, achieved up to the data points purchased, where $\Delta v$ represents the reduction in posterior variance due to each data point, and $\Delta c$ denotes its associated price. A higher value of $\Delta v / \Delta c$ indicates a greater reduction in variance per unit cost.} As shown in this figure, VBAL and QBCAL consistently outperform the baseline RSC method.  \minorrev{Greedy Knapsack (GK) exhibits a high initial cumulative $\Delta v/\Delta c$ mainly because it purchases the lowest-cost points first ($1/\eta_j$ in a descending order), which inflates the early running average. However, this advantage quickly diminishes as additional low-cost points yield smaller marginal variance reductions, causing GK’s cumulative $\Delta v/\Delta c$ to decline. Consistently, Table~\ref{tab:comparison1} shows that GK delivers lower improvement per purchased point overall than VBAL and QBCAL.} Note that we report dynamics only for SC because, under constraints \eqref{constraint2}–\eqref{constraint3} in the optimization problem \eqref{analyst_problem_relax}, the cost-efficiency ratio in BC is fixed at \(1/\phi\) and thus omitted


\subsubsection{Do pricing approaches affect sellers?}
\label{seller in v}
\begin{figure}[!ht]
    \centering
    \subfloat[VBAL] {
        \includegraphics[width=.35\linewidth]{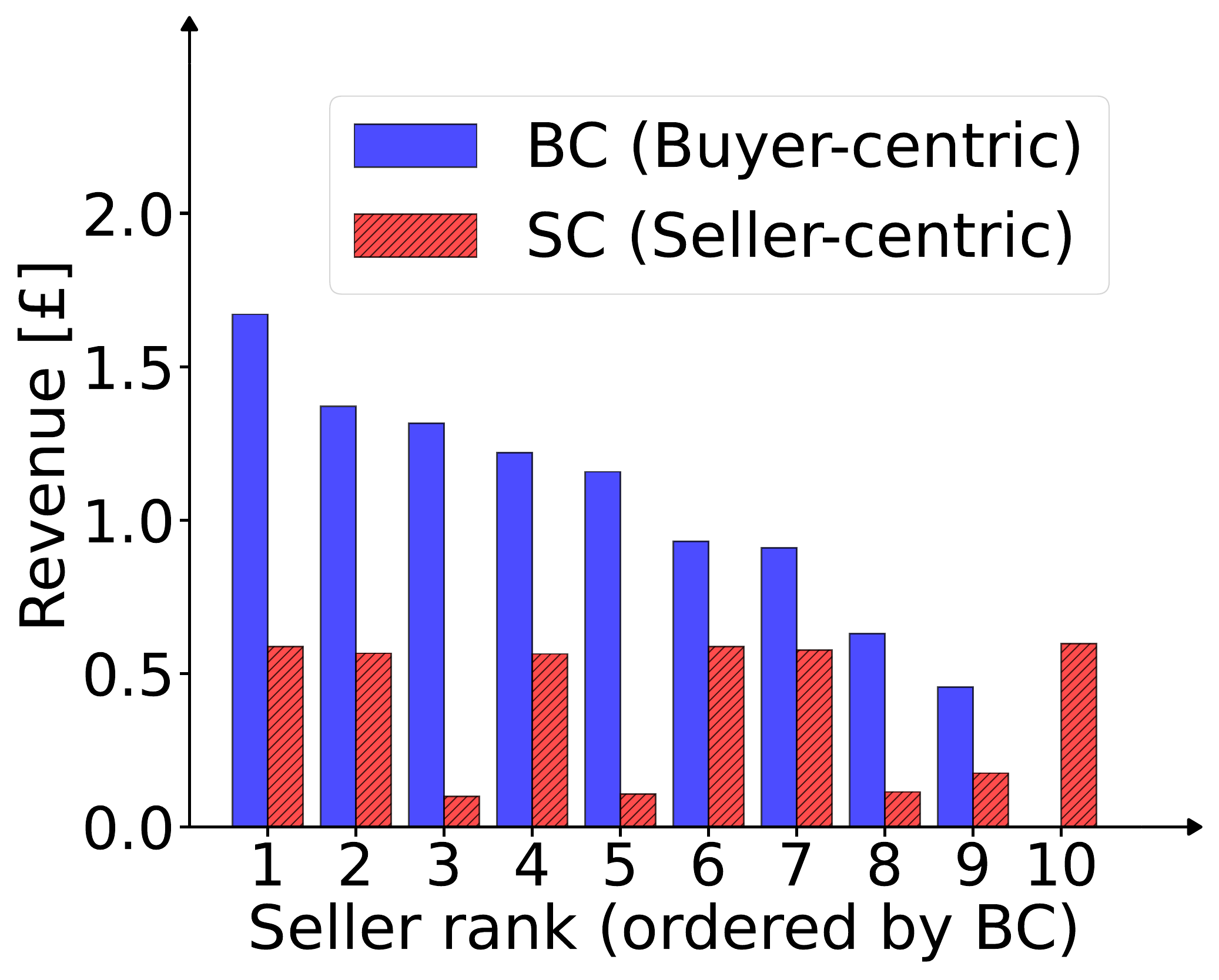}
        \label{fig:V_Revenue_Difference-a}}
\hfill
    \subfloat[QBCAL]{
        \includegraphics[width=.35\linewidth]{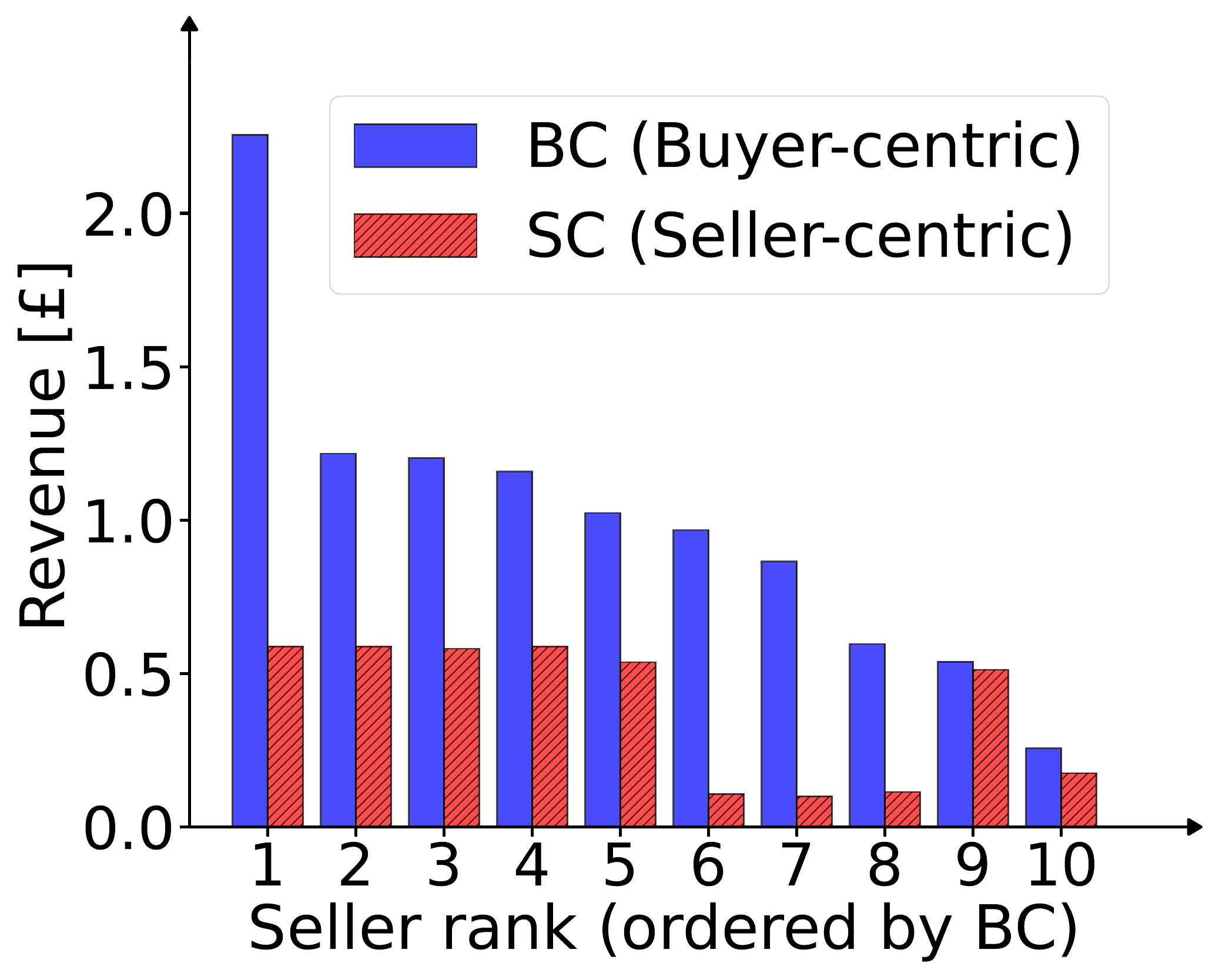}
        \label{fig:V_Revenue_Difference-b}}
    \hfill
    \subfloat[RSC]{
        \includegraphics[width=.35\linewidth]{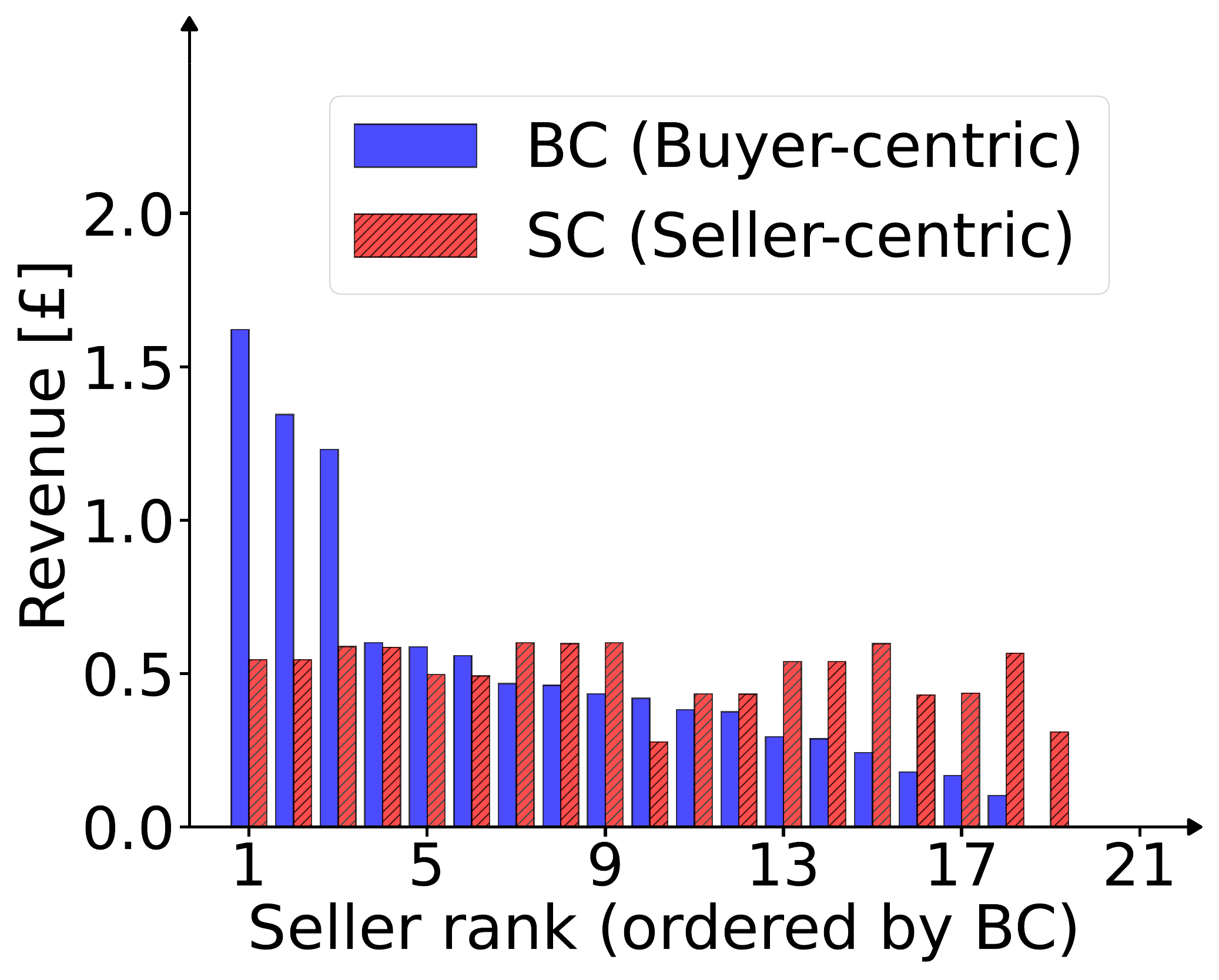}
        \label{fig:V_Revenue_Difference-c}}
    \hfill
    \subfloat[\minorrev{Greedy Knapsack}]{

        \includegraphics[width=.35\linewidth]{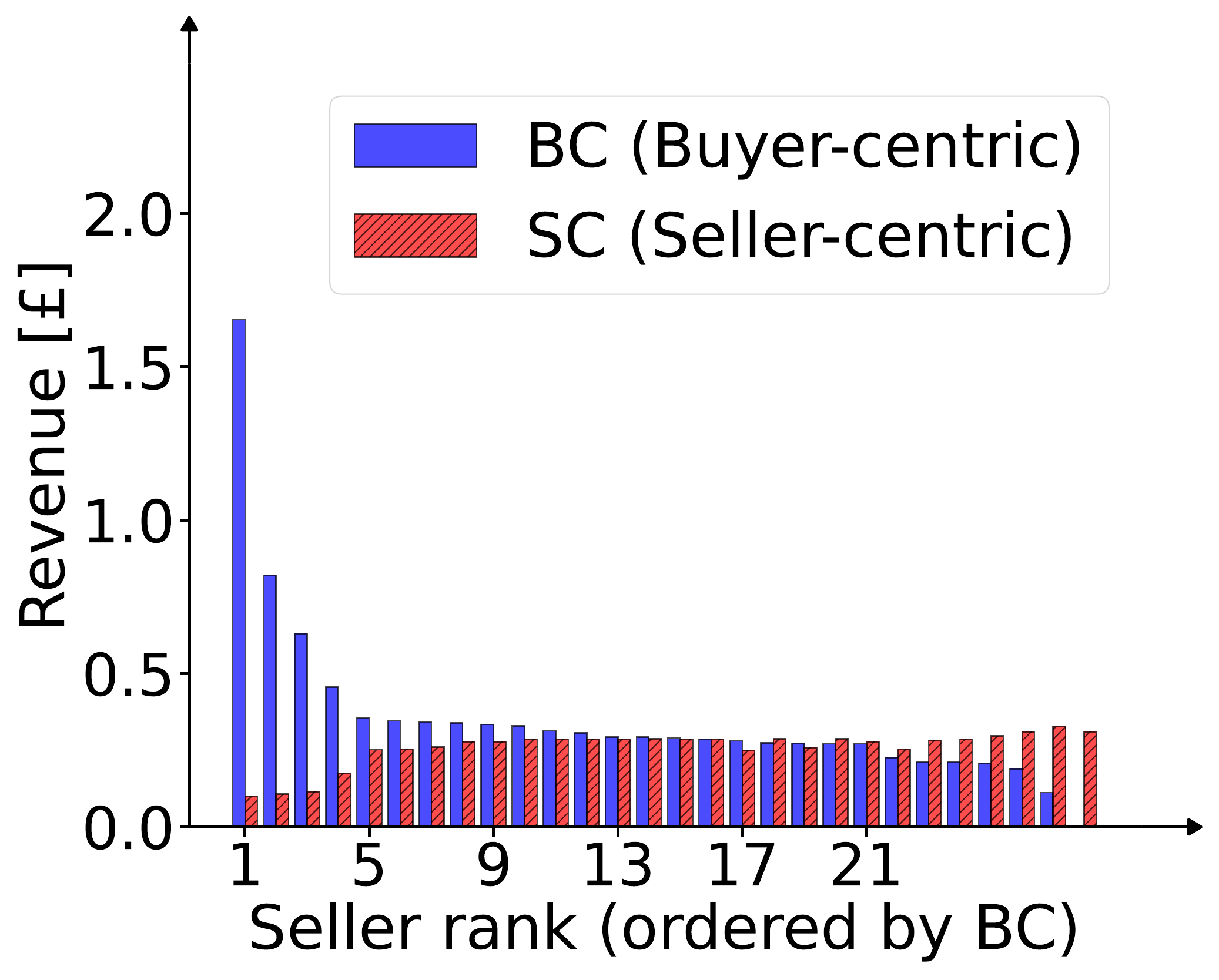}
        \label{fig:V_Revenue_Difference-d}}

     \caption{Revenue comparison for sorted data sellers under BC and SC pricing approaches\label{fig:V_Revenue_Difference}}
    
\end{figure}

To assess how pricing schemes influence sellers’ revenue, we examine the set of sellers selected under the BC and SC approaches, ranking them in descending order by their BC revenue. Figure~\ref{fig:V_Revenue_Difference} reports the revenue differences between the two schemes. The x-axis represents the number of sellers rather than specific IDs; the two adjacent bars correspond to the same seller under BC and SC. The results show that data prices in BC are consistently higher than in SC. In the buyer-centric scheme, the analyst allocates her budget to maximize improvement and is willing to pay more for highly informative labels, resulting in substantially higher revenue for the most valuable sellers. In contrast, the seller-centric scheme spreads the budget more evenly, producing smaller revenue differences between sellers. In general, BC rewards high-value sellers more strongly by highlighting informativeness, while SC produces a more balanced distribution of revenue across the market.

\label{sec:var_scene_monte_carlo}
 \subsubsection{Is Our Active Learning Strategy Robust?}\label{sec:sens_1}
\label{sec:robustness_analysis}
\begin{revisedblock}

For all robustness experiments in this section, we evaluate the algorithms using the 
\emph{number of effectively purchased data points} — the subset of acquired points 
that were actually incorporated into the model and contributed to improvement.  
This distinguishes true learning progress from mere expenditure. To assess robustness, we performed two Monte Carlo–based analyses:  
the first examined how random data perturbations affect performance, and the second explored the sensitivity to key model and economic parameters.

\paragraph{Data Variability.}
A Monte Carlo simulation with 1,000 replications was conducted by repeatedly resampling
subsets of the original data, with a model-improvement threshold fixed at 20\%.
Figures \ref{fig:MC_BC} and \ref{fig:MC_SC} show the resulting distributions of effectively purchased data points (i.e., data points used to improve the model) for buyer-centric (BC) and seller-centric (SC) pricing schemes.
VBAL and QBCAL exhibit concentrated frequency distributions
((Figures~\ref{fig:MC_BC}\subref{fig:MC_BC-a}--\subref{fig:MC_BC-b} and
\ref{fig:MC_SC}\subref{fig:MC_SC-a}--\subref{fig:MC_SC-b}),
indicating stable behaviour across replications,
while the random (RSC) and the Greedy Knapsack (GK) baselines produce flatter distributions
(Figures~\ref{fig:MC_BC}\subref{fig:MC_BC-c}--\subref{fig:MC_BC-d} and
\ref{fig:MC_SC}\subref{fig:MC_SC-c}--\subref{fig:MC_SC-d}),
reflecting higher variability and lower robustness.
\begin{figure}[H]
  \centering

  \begin{subfigure}{0.33\linewidth}
    \centering
    \includegraphics[width=\linewidth]{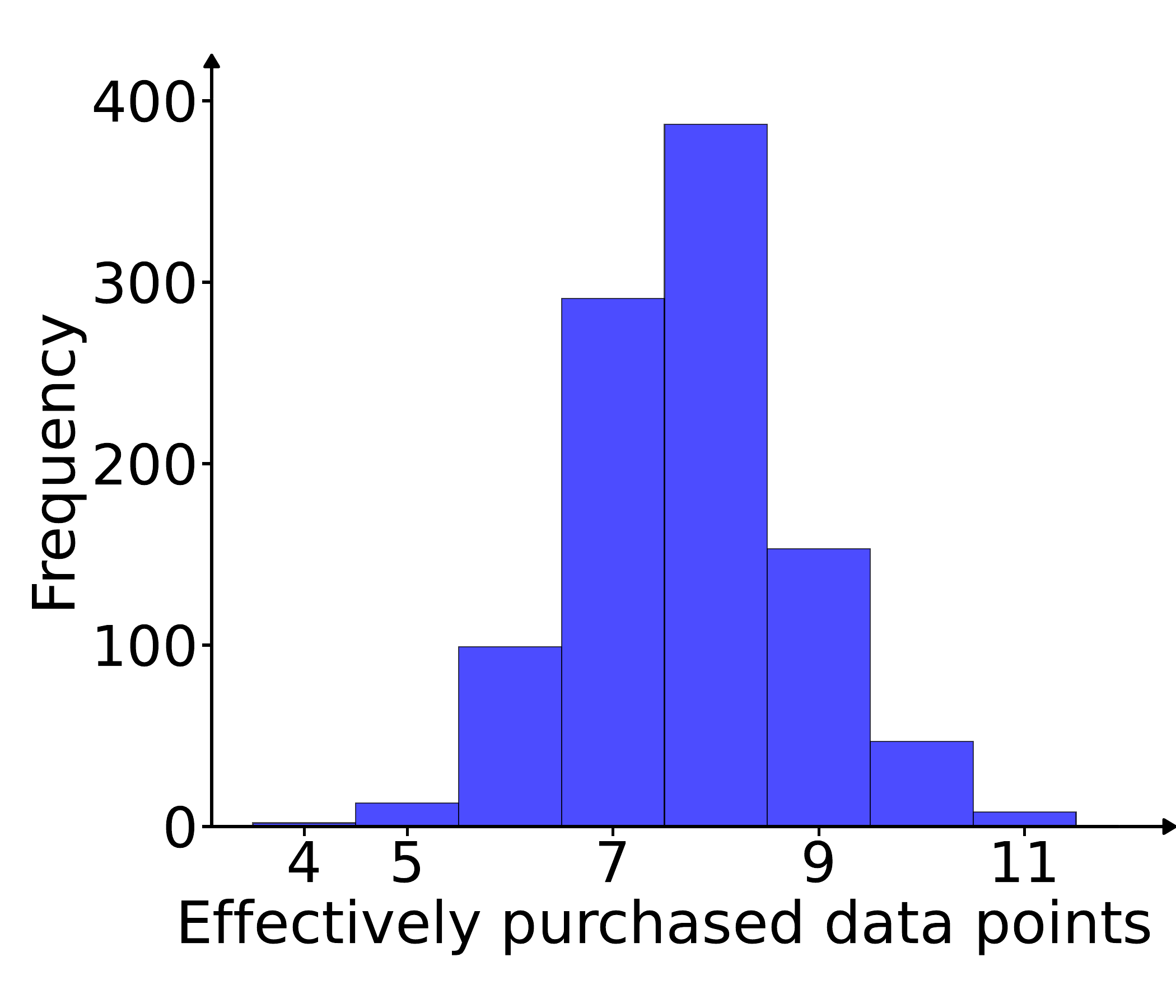}
    \caption{VBAL}
    \label{fig:MC_BC-a}
  \end{subfigure}\hfill
  \begin{subfigure}{0.33\linewidth}
    \centering
    \includegraphics[width=\linewidth]{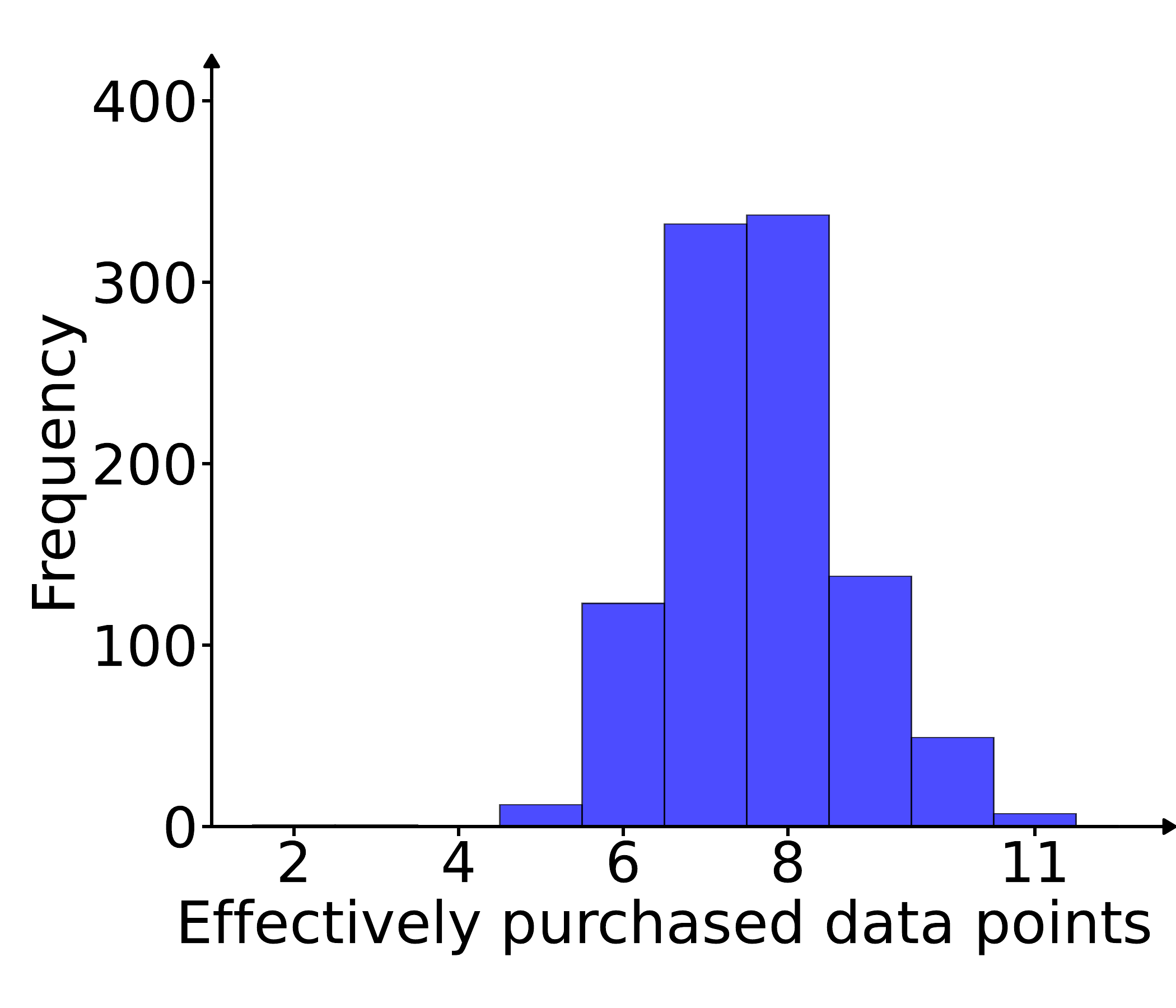}
    \caption{QBCAL}
    \label{fig:MC_BC-b}
  \end{subfigure}

  \vspace{2mm}

  \begin{subfigure}{0.33\linewidth}
    \centering
    \includegraphics[width=\linewidth]{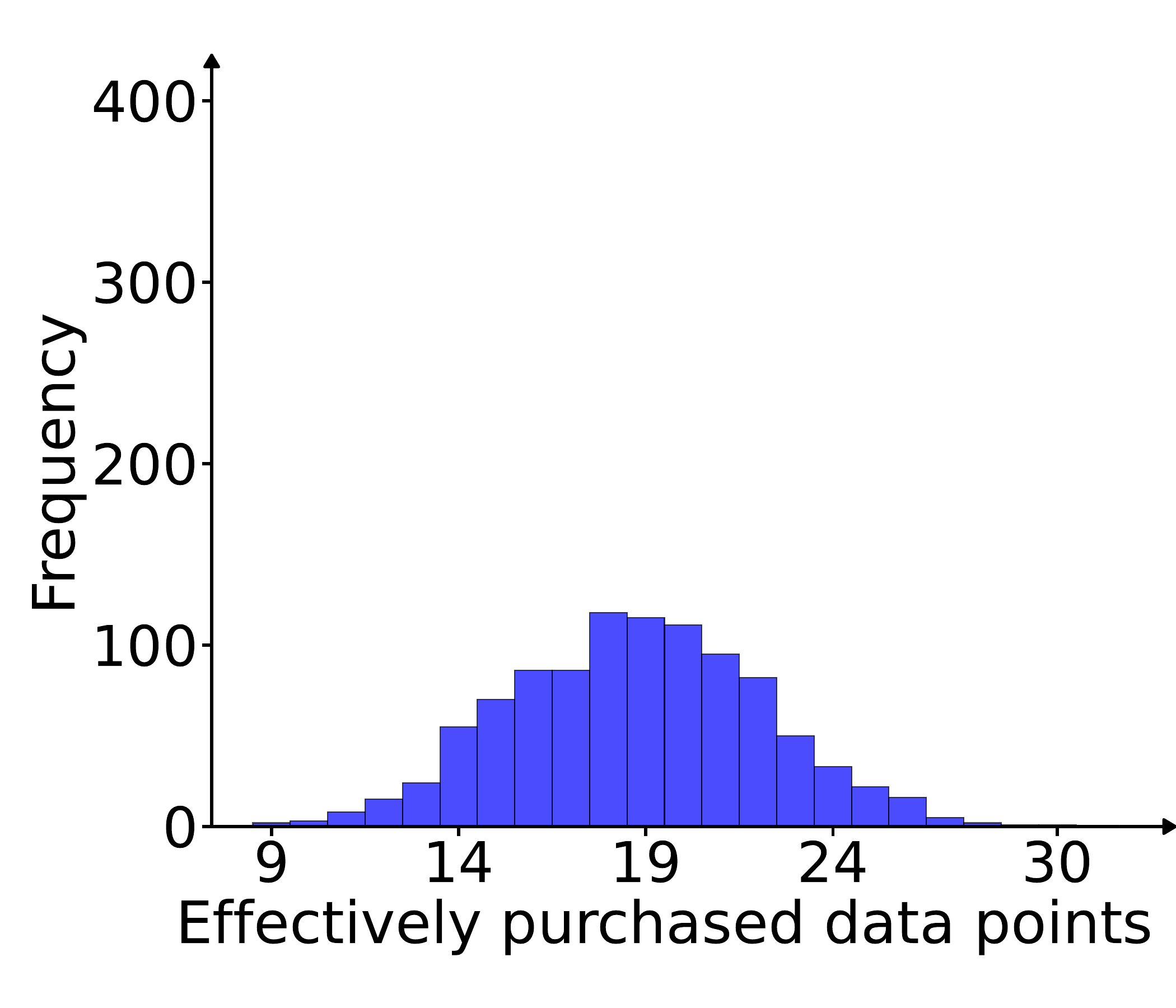}
    \caption{RSC}
    \label{fig:MC_BC-c}
  \end{subfigure}\hfill
  \begin{subfigure}{0.33\linewidth}
    \centering
    \includegraphics[width=\linewidth]{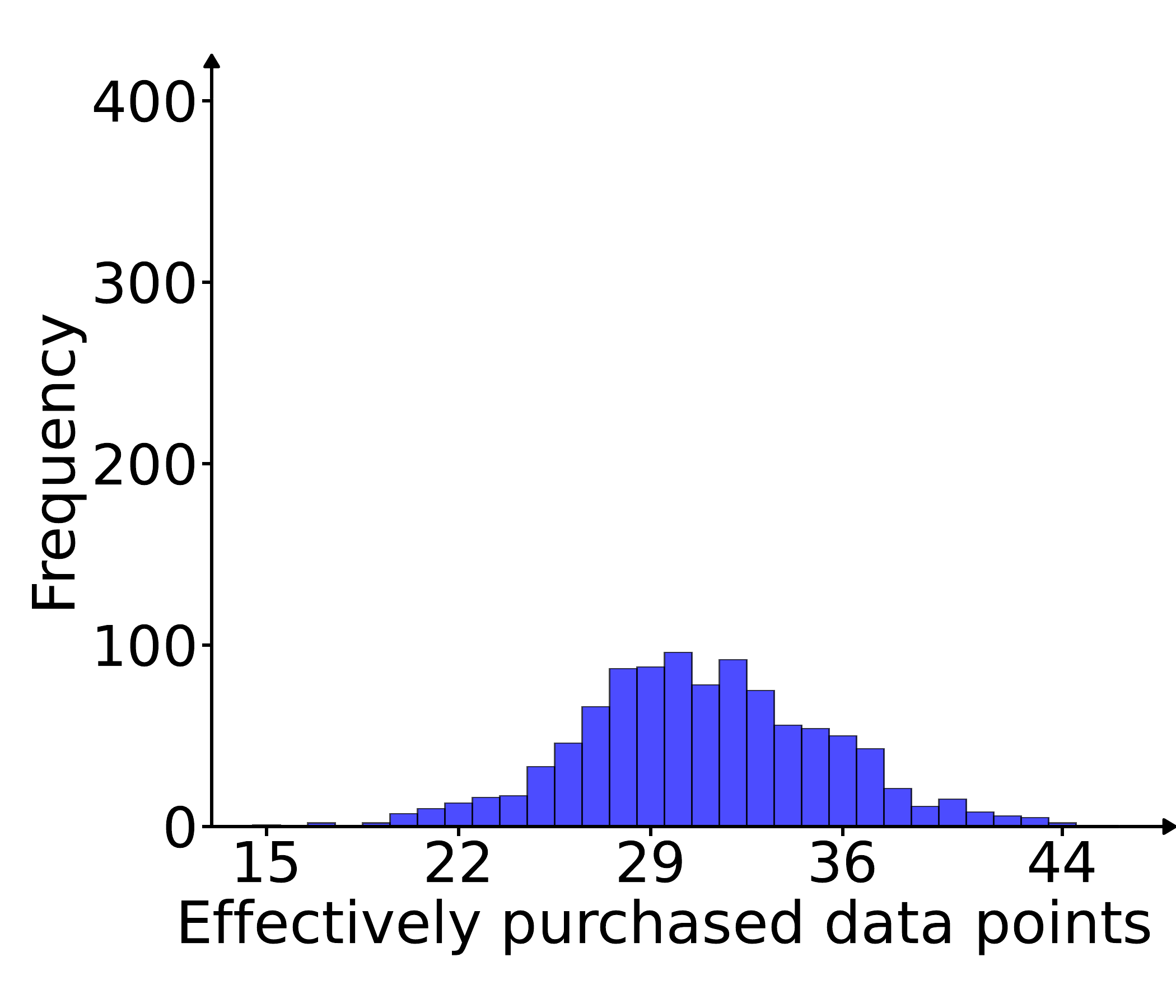}
    \caption{\minorrev{Greedy Knapsack}}
    \label{fig:MC_BC-d}
  \end{subfigure}

  \caption{Monte-Carlo simulation for the buyer-centric pricing approach}
  \label{fig:MC_BC}
\end{figure}

\begin{figure}[!ht]
  \centering

  \begin{subfigure}{0.33\linewidth}
    \centering
    \includegraphics[width=\linewidth]{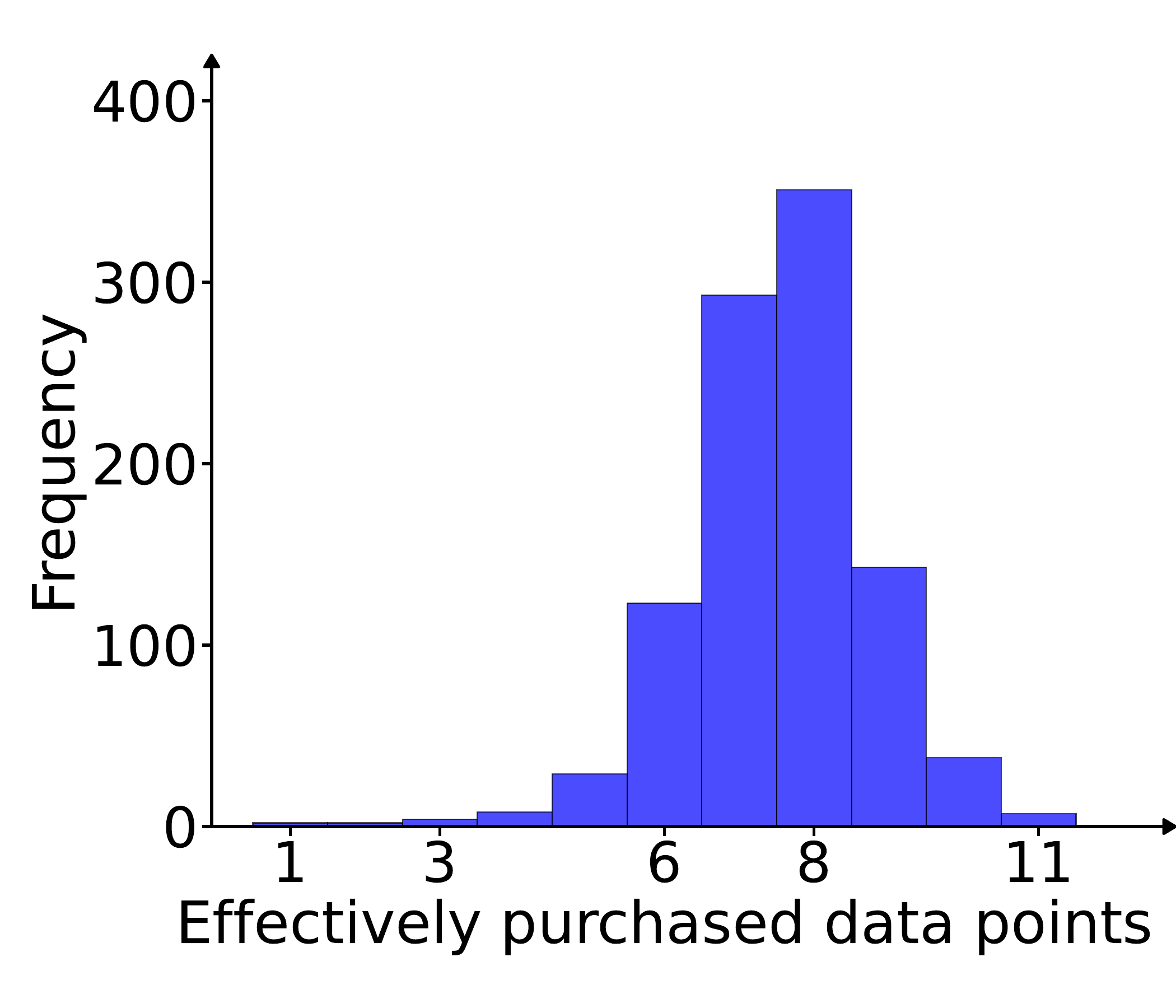}
    \caption{VBAL}
    \label{fig:MC_SC-a}
  \end{subfigure}\hfill
  \begin{subfigure}{0.33\linewidth}
    \centering
    \includegraphics[width=\linewidth]{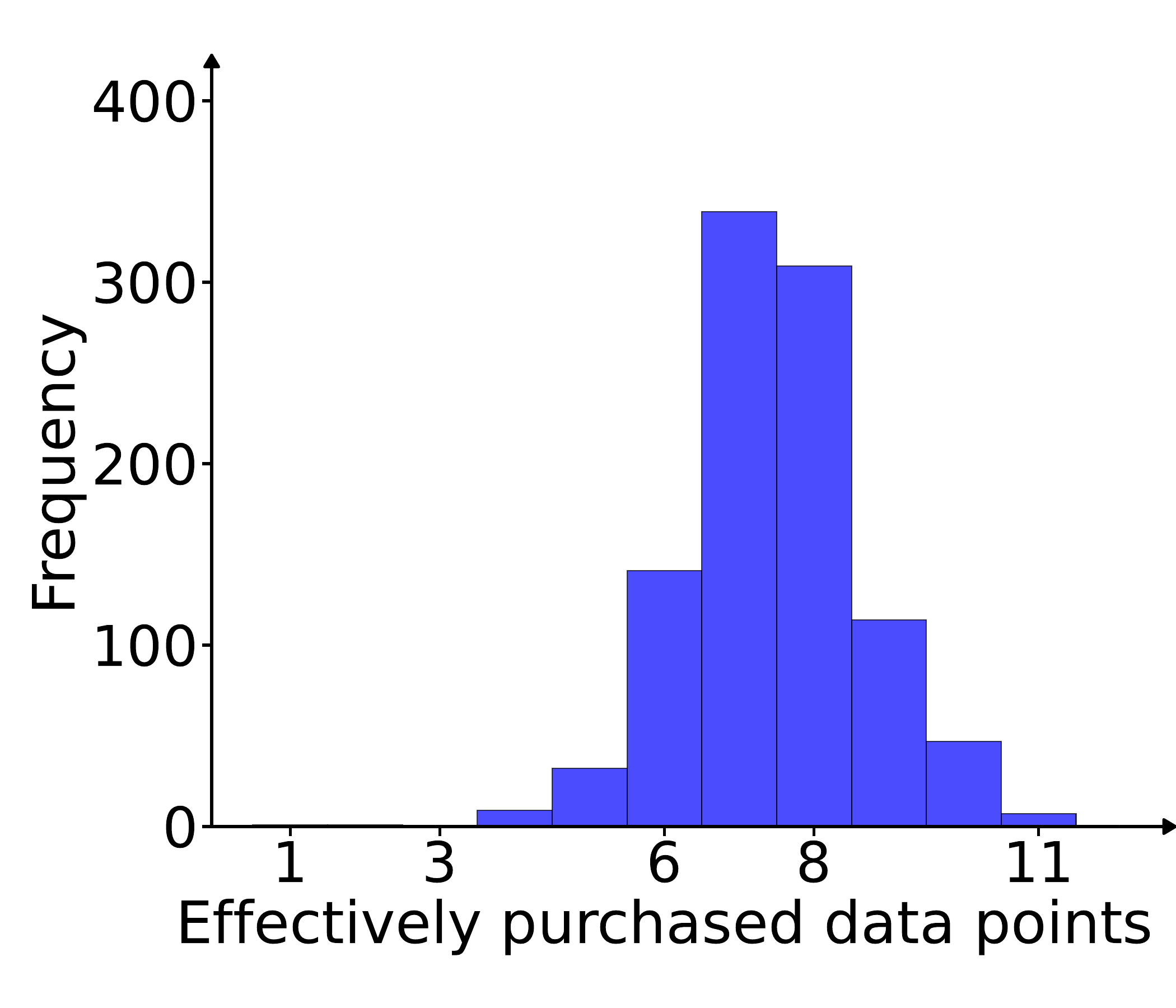}
    \caption{QBCAL}
    \label{fig:MC_SC-b}
  \end{subfigure}

  \vspace{2mm}

  \begin{subfigure}{0.33\linewidth}
    \centering
    \includegraphics[width=\linewidth]{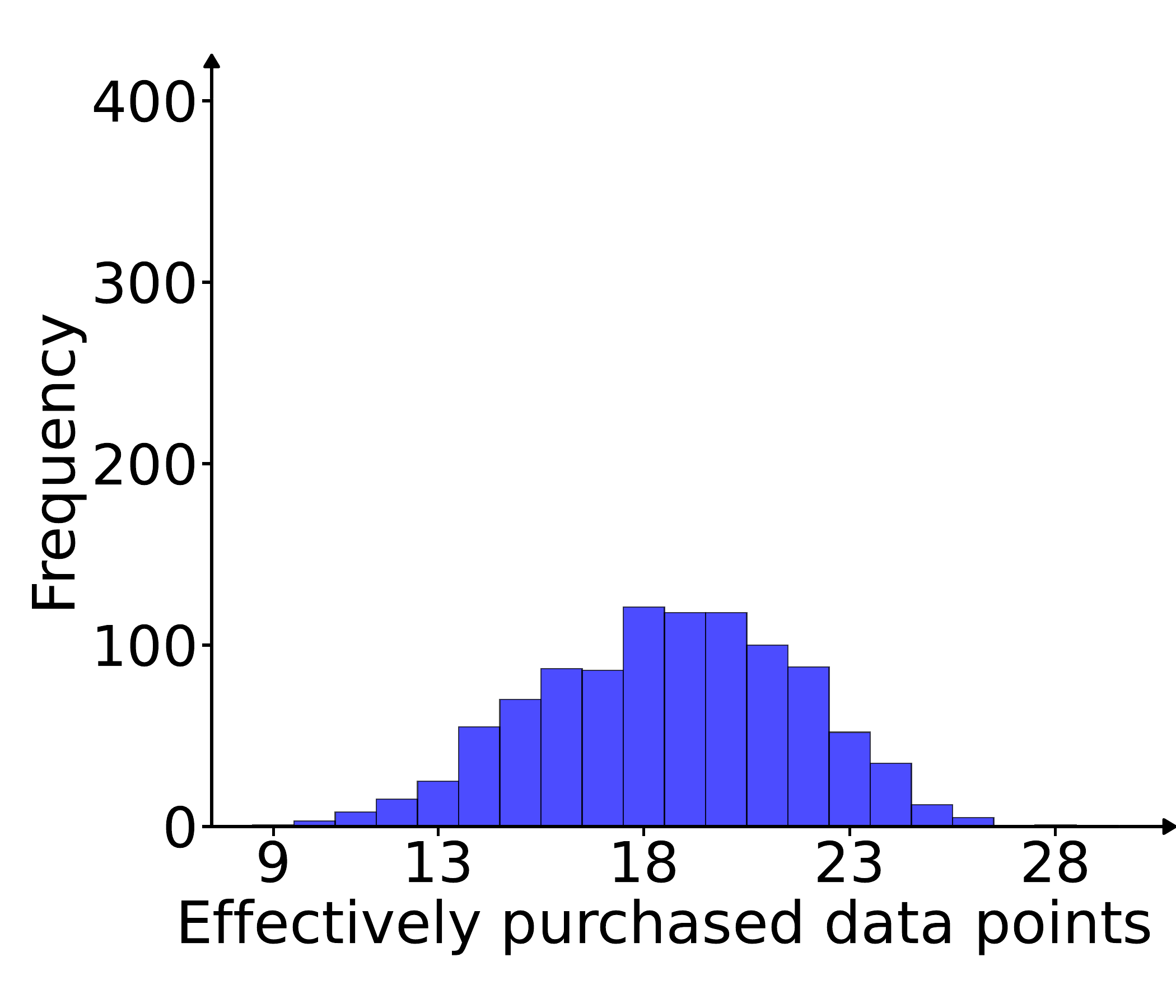}
    \caption{RSC}
    \label{fig:MC_SC-c}
  \end{subfigure}\hfill
  \begin{subfigure}{0.33\linewidth}
    \centering
    \includegraphics[width=\linewidth]{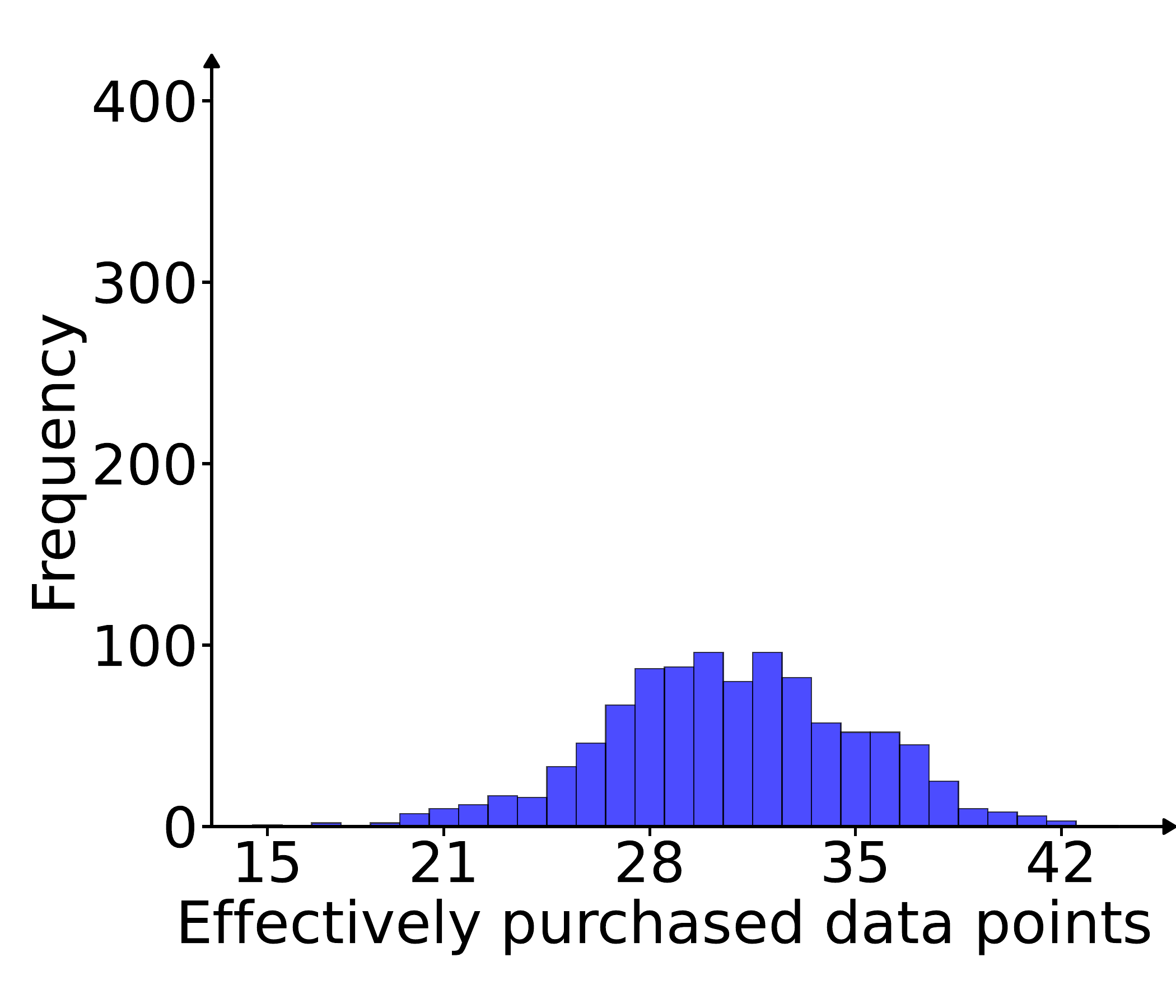}
    \caption{\minorrev{Greedy Knapsack}}
    \label{fig:MC_SC-d}
  \end{subfigure}

  \caption{Monte-Carlo simulation for the seller-centric pricing approach}
  \label{fig:MC_SC}
\end{figure}

\paragraph{Parameter Sensitivity.}
We further examine parameter sensitivity in the estimation-quality-focused setting by varying (i) the buyer’s WTP $\phi$, (ii) the seller’s WTS scaling coefficient $d_1$, and (iii) the budget limit $B$, while keeping all other parameters at their baseline values.
\minorrev{Figures~\ref{fig:sensitivity_WTP_BC}--\ref{fig:sensitivity_B_SC} and Table~\ref{tab:montecarlo_bc_sc_var} (Appendix) show that the effective number of purchased points is largely insensitive to $\phi$ in the tested range under both BC and SC pricing: VBAL and QBCAL consistently purchase only a small set of informative points (around 10), whereas RSC and Greedy Knapsack purchase substantially more points (around 19--21 and $28$, respectively). Increasing $B$ from \pounds10 to \pounds30 has a limited impact in these experiments, as most methods reach the variance threshold before exhausting the larger budgets.}

\begin{figure}[H]
\centering
\begin{subfigure}{0.45\textwidth}
    \centering
    \includegraphics[width=\linewidth]{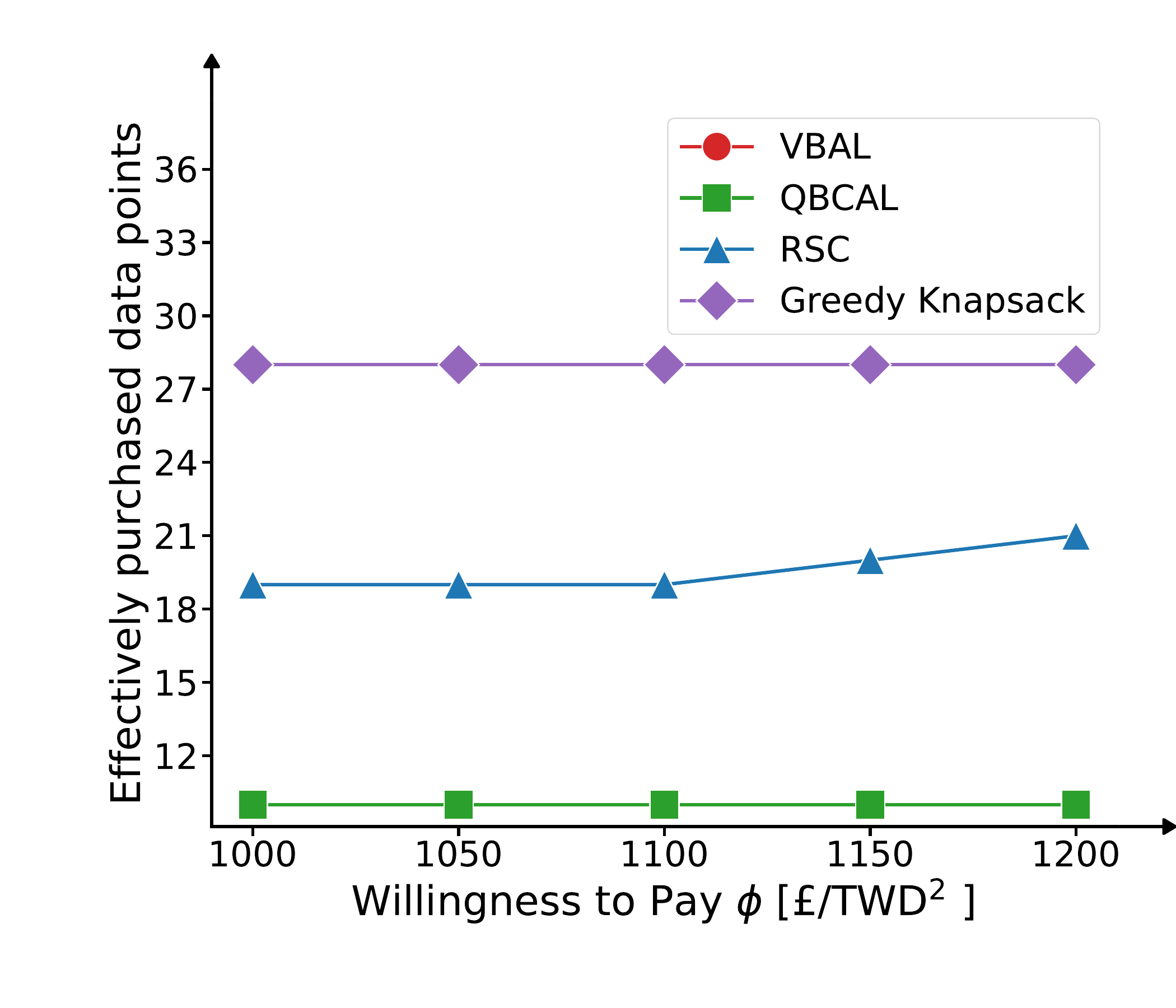}
    \caption{WTP variation under BC pricing}
    \label{fig:sensitivity_WTP_BC}
\end{subfigure}\hfill
\begin{subfigure}{0.45\textwidth}
    \centering
    \includegraphics[width=\linewidth]{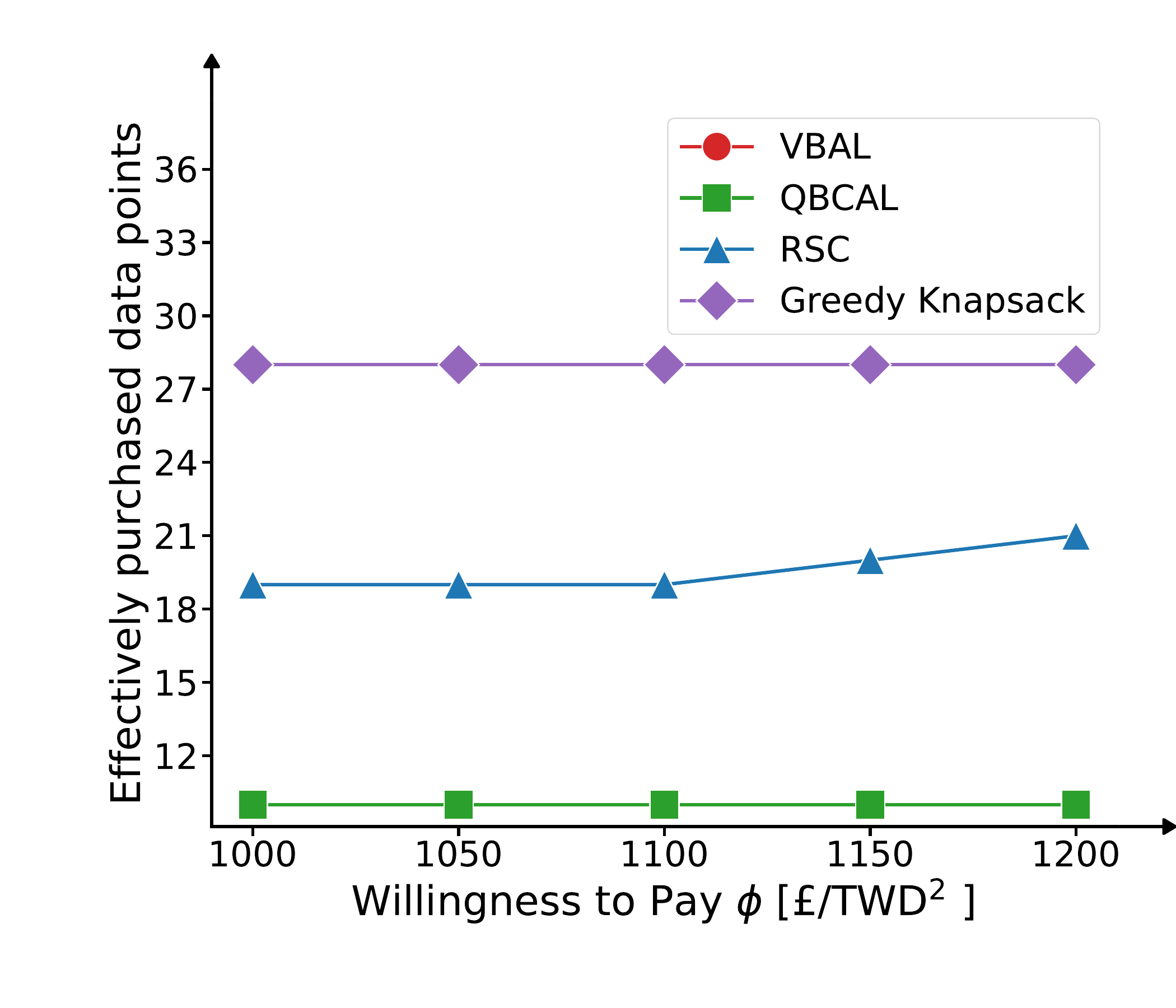}
    \caption{WTP variation under SC pricing}
    \label{fig:sensitivity_WTP_SC}
\end{subfigure}

\vspace{0.6em} 

\begin{subfigure}{0.45\textwidth}
    \centering
    \includegraphics[width=\linewidth]{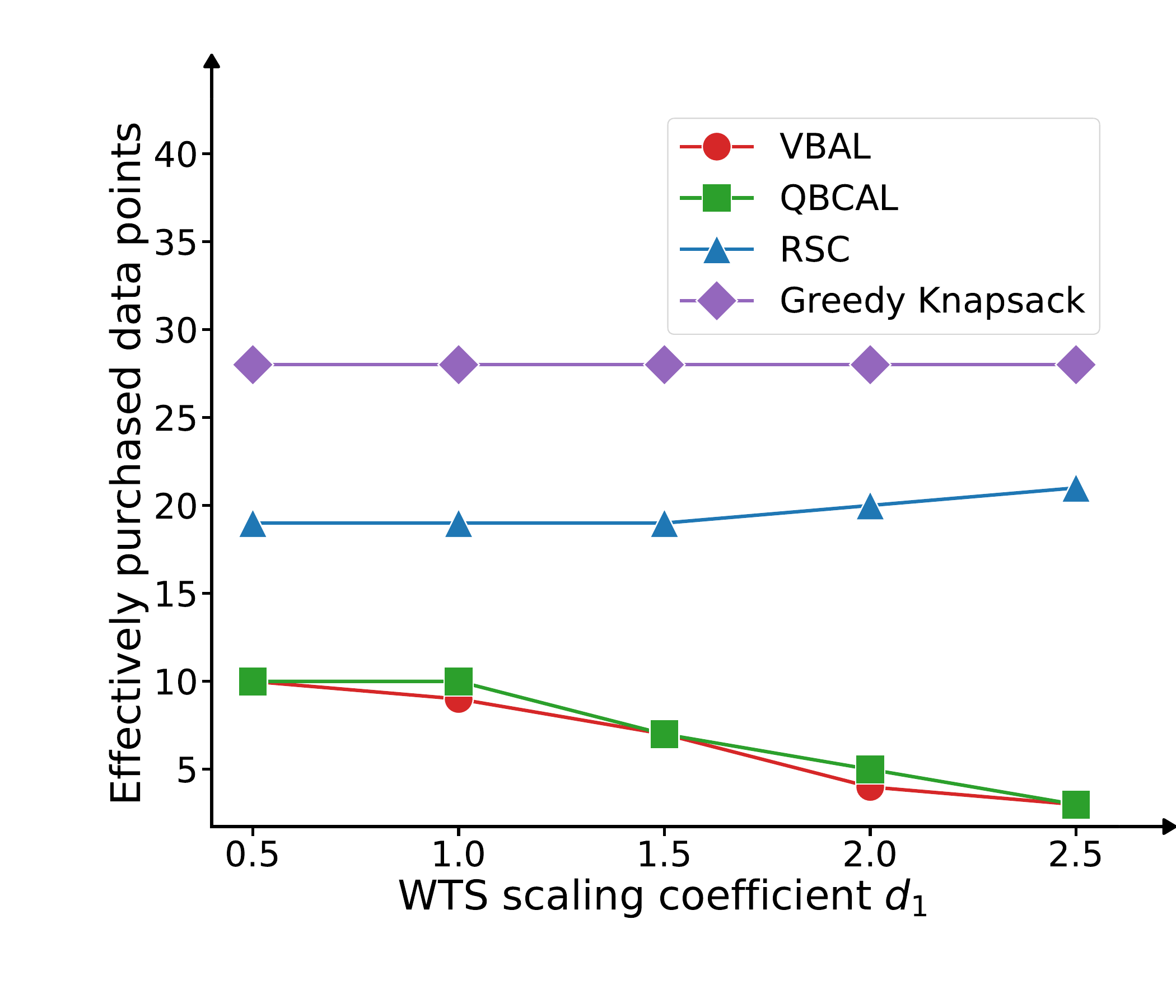}
    \caption{WTS variation under BC pricing}
    \label{fig:sensitivity_WTS_BC}
\end{subfigure}\hfill
\begin{subfigure}{0.45\textwidth}
    \centering
    \includegraphics[width=\linewidth]{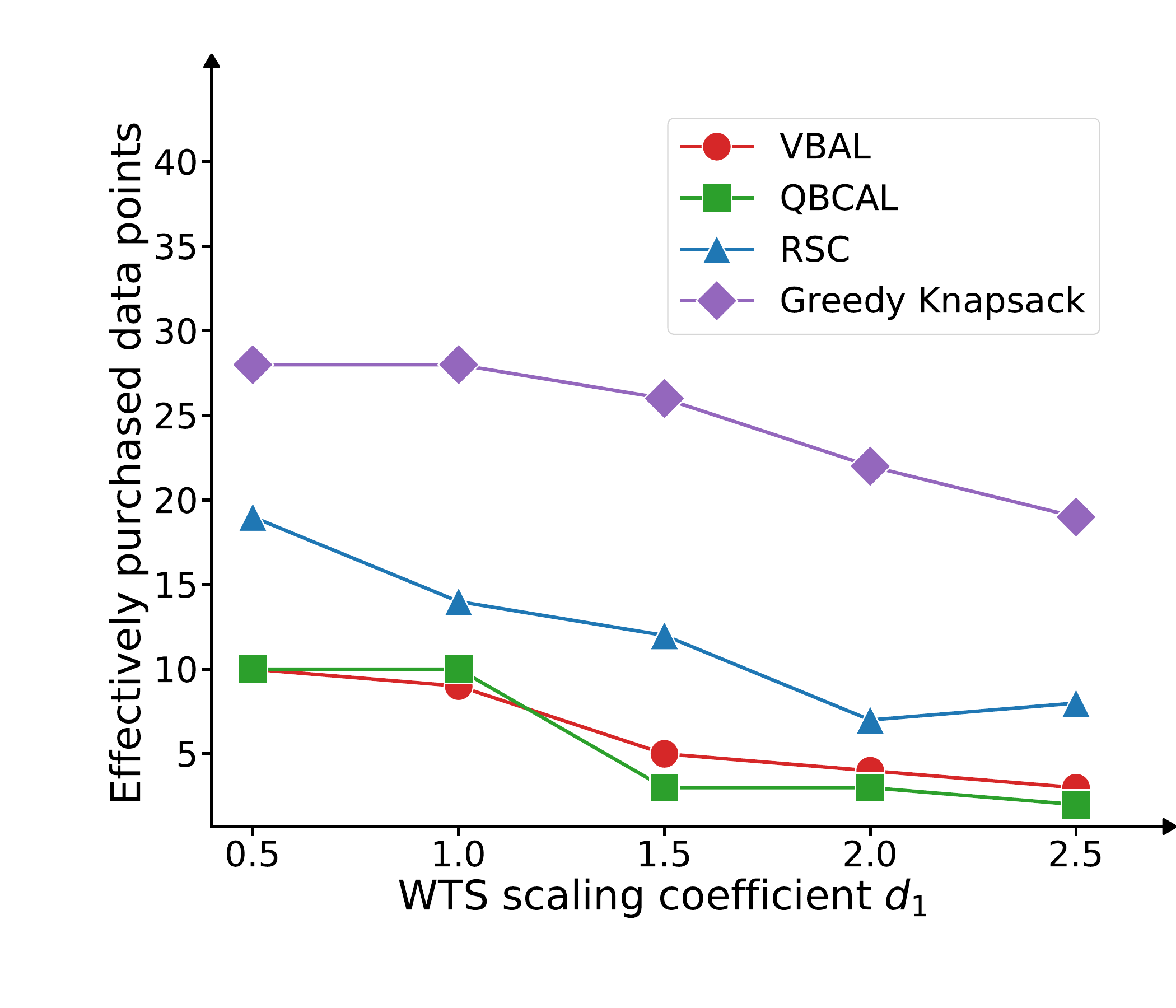}
    \caption{WTS variation under SC pricing}
    \label{fig:sensitivity_WTS_SC}
\end{subfigure}

\vspace{0.6em}

\begin{subfigure}{0.45\textwidth}
    \centering
    \includegraphics[width=\linewidth]{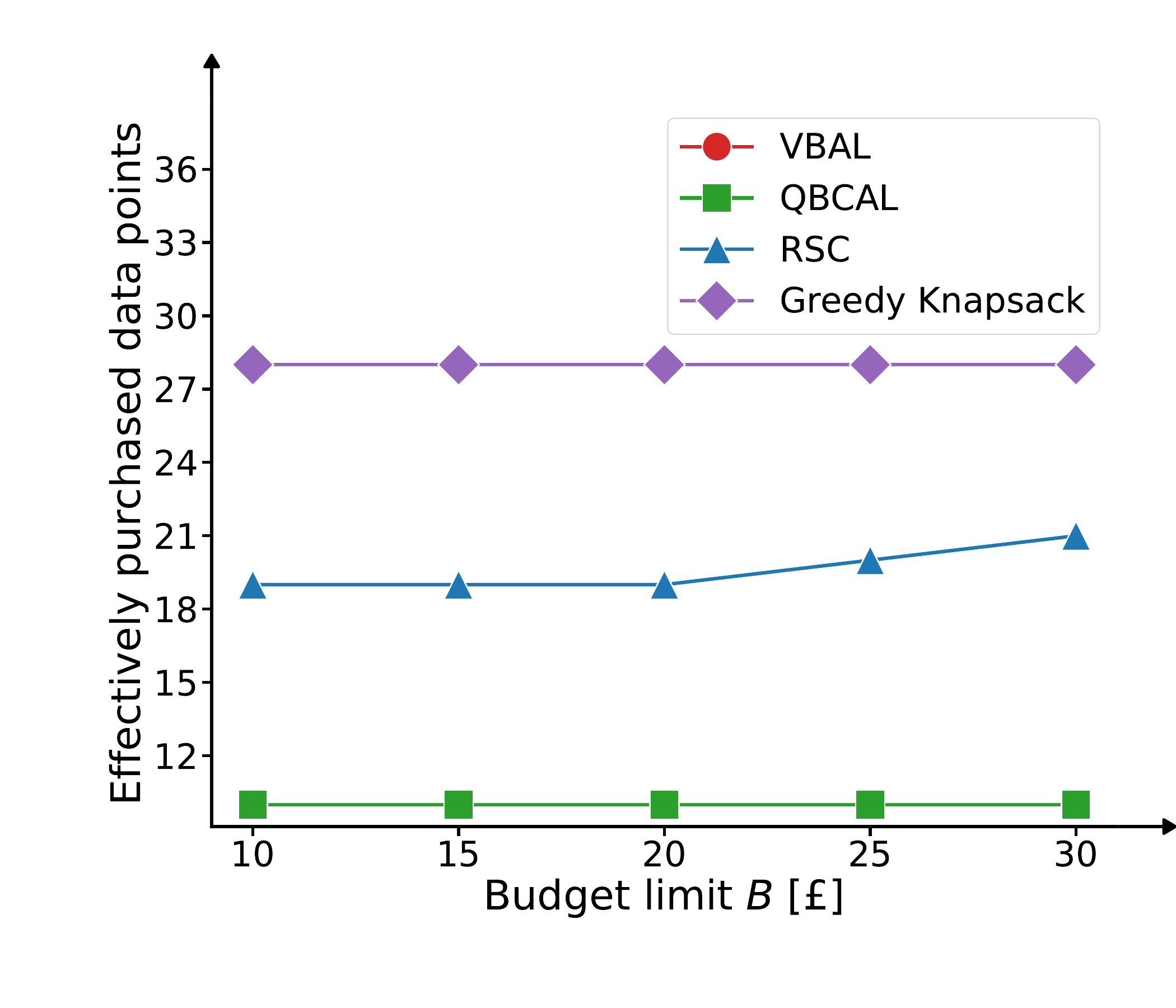}
    \caption{Budget variation under BC pricing}
    \label{fig:sensitivity_B_BC}
\end{subfigure}\hfill
\begin{subfigure}{0.45\textwidth}
    \centering
    \includegraphics[width=\linewidth]{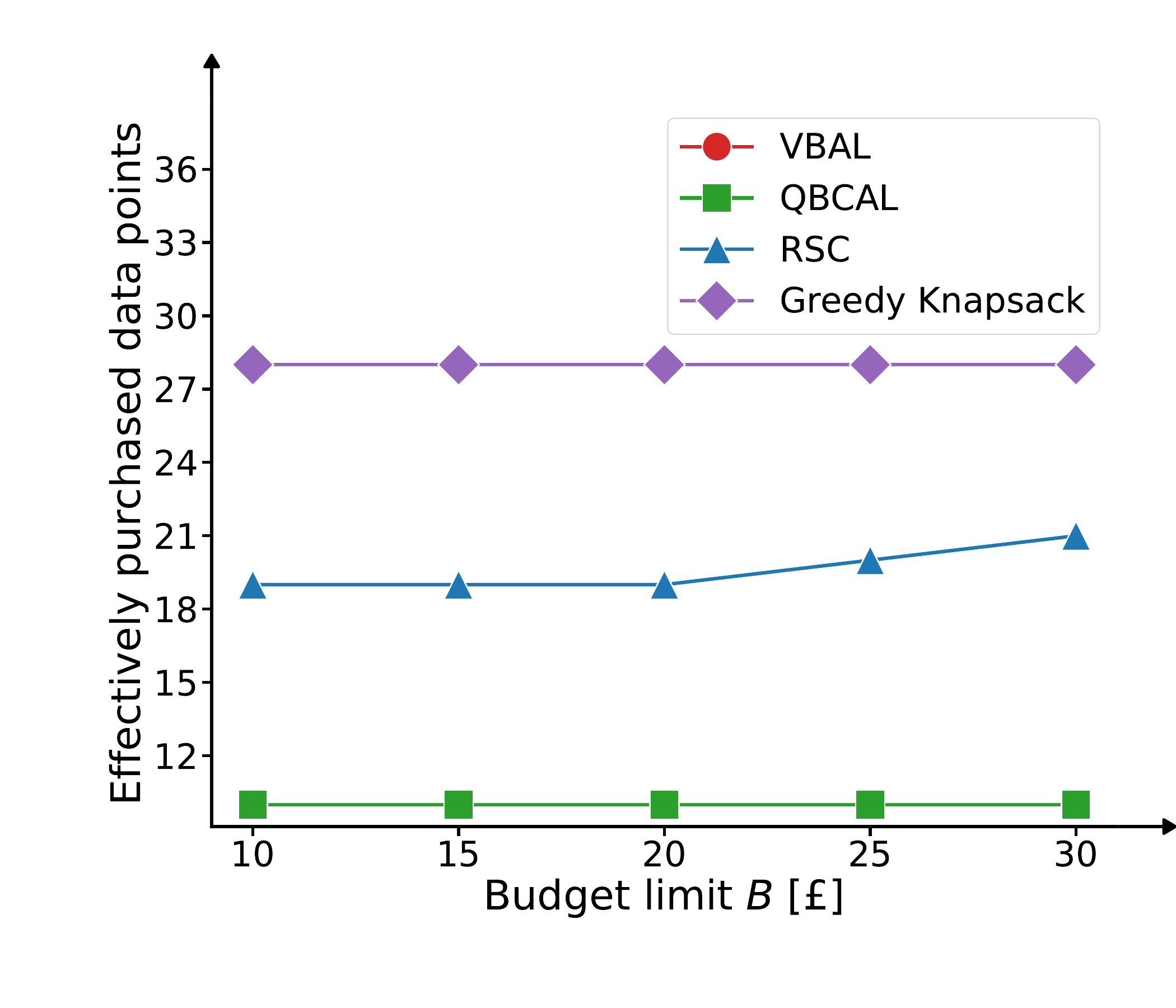}
    \caption{Budget variation under SC pricing}
    \label{fig:sensitivity_B_SC}
\end{subfigure}

\vspace{0.6em}

\caption{\minorrev{Sensitivity analyses of the buyer’s willingness-to-pay ($\phi$), the seller’s willingness-to-sell (WTS) scaling coefficient ($d_1$), and the budget limit ($B$). The left column shows the results for buyer-centric (BC) pricing, while the right column corresponds to seller-centric (SC) pricing.}}
\label{fig:sensitivity_combined}
\end{figure}
\end{revisedblock}

\subsection{Focus on predictive ability}

Although active learning in the \revised{estimation-quality-focused} scenario shows promising results, it is equally important to evaluate performance in a \revised{predictive-ability-focused} setting, where the emphasis is on generalization.

\subsubsection{Case study and dataset}

We use the BDG2 dataset \citep{wang2022adaptive}, which contains the hourly energy consumption of 1,636 buildings from 2016 to 2017. For illustration, we consider an educational building (“Rachael”) as the data analyst and identify its most similar building (“Madge”) via Euclidean distance in the feature space, treating it as the data seller. Each observation includes the month, weekday, day, hour, holiday indicator, air temperature, and energy consumption. We adopt a sliding-window approach using only the ``value'' characteristic (hourly consumption) to construct lagged inputs: the consumption of the past seven days’ at 24-hour intervals (e.g.\ $t{-}168$, $t{-}144$, \dots, $t{-}24$), with the current hour’s consumption at time $t$ as the prediction target. Then a linear forecasting model is trained on these lagged features.

\paragraph{Label availability.}
Although the BDG2 data set contains complete hourly labels, we simulate a market environment in which labels \(Y\) must be purchased at the time of use. This mimics realistic settings where timely, cleaned, and query-ready data carry an access cost—even if such data may later become freely available (as in IoT systems). In our batch-learning setup, labels become available only when purchased, consistent with the market-based learning framework.

In this experiment, \revised{model-improvement threshold is set to 20\%} and the budget is \(B=\text{£}1200\). The analyst’s WTP is \(\phi = 50\) (£/\(\Delta\)MSE). As in Section~\ref{MSE scenario}, “Rachael’’ represents the analyst and “Madge’’ is the seller. The WTS of the seller is set uniformly at \(\eta = 30\) (£/label). \revised{The mechanism sensitivity constant is set to $30$ and scaled by the feature-normalisation factor $\gamma$, giving an effective sensitivity of $30\gamma$.}

\subsubsection{What are the effects of applying active learning?}

\revised{Table~\ref{tab:comparison2} compares the performance of the proposed active learning strategies (VBAL and QBCAL) with the random sampling corrected strategy (RSC). In the BC setting, VBAL and QBCAL achieve similar levels of model improvement ($\sim17\%$) at lower average cost per data point (£12.81 and £7.68, respectively) relative to RSC (£13.53). In the SC setting, the average cost per data point is identical between methods by design, so performance differences arise solely from the selectivity of the strategies rather than from the pricing. The trends in model improvement and accumulated cost closely mirror those in Section~\ref{sec:var_scen_cost_efficiency} and are omitted for brevity. Under the seller-centric mechanism the reported average cost is determined by the fixed WTS schedule, so a Wilcoxon signed-rank test on cost differences is uninformative. For the same reason, we do not include a Greedy Knapsack baseline for this dataset as it cannot meaningfully differentiate points (any selection is equivalent).}


\begin{table}[h!]
\centering
\small
\caption{Comparison of Data Purchasing Strategies: Predictive-ability-focused  scenario}
\label{tab:comparison2}
\begin{tabular}{lccccccc}
\hline
\textbf{Strategy} & \textbf{Approach} & \textbf{\(B\) ?} & \textbf{\(\alpha\)?} & \textbf{Data Bought} & \textbf{Spent (£)} & \textbf{Improvement (\%)} & \textbf{Avg. Cost (£/data)} \\
\midrule
\multirow{2}{*}{VBAL} 
  & BC & Yes & No & 97 & 1242.77 & 17.12 & 12.81 \\
  & SC & Yes & No & 40 & 1200.00 & 11.12 & 30.00 \\
\hline
\multirow{2}{*}{QBCAL} 
  & BC & Yes  & No & 159 & 1221.84 & 16.83 & 7.68 \\
  & SC & Yes & No & 40  & 1200.00 & 11.39 & 30.00 \\
\hline
\multirow{2}{*}{RSC}   
  & BC & Yes & No & 90 & 1217.52 & 16.77 & 13.53 \\
  & SC & Yes & No & 40 & 1200.00 & 12.70 & 30.00 \\
\hline
\end{tabular}
\end{table}




\subsubsection{Does active learning improve cost-efficiency for analysts?}

\begin{revisedblock}
Similarly in Section~\ref{var:costeff}, Figure~\ref{fig:M_SC_analyst} illustrates the accumulative cost-efficiency metric, denoted by $\Delta MSE/ \Delta c$, as more data points are purchased. The VBAL and QBCAL strategies consistently achieve higher $\Delta MSE / \Delta c$ values than RSC, indicating greater model improvement per unit cost. This demonstrates that the active learning-based approaches provide higher returns on labeling investment, particularly under limited budgets.

\begin{figure}[H]
    \centering
    \includegraphics[width=0.42\linewidth]{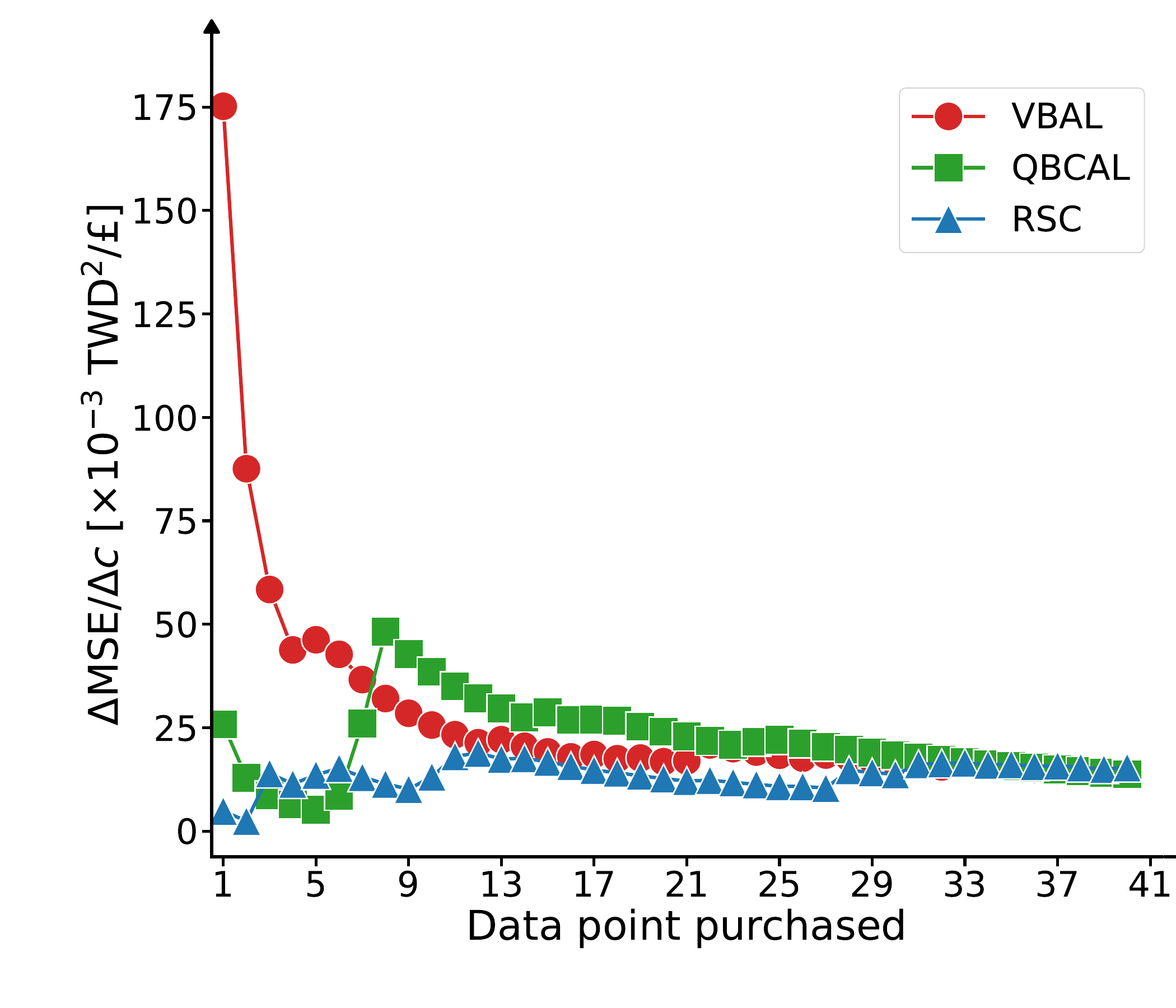}
    \caption{Analyst's side analysis for the SC approach: MSE reduction per unit cost. Results are represented as cumulative averages (i.e., as the average up to that data point purchased)}
    \label{fig:M_SC_analyst}
\end{figure}
\end{revisedblock}

\subsubsection{Do pricing approaches affect sellers?}

\begin{figure}[H]
    \centering
    \subfloat[VBAL]{
        \includegraphics[width=.32\linewidth]{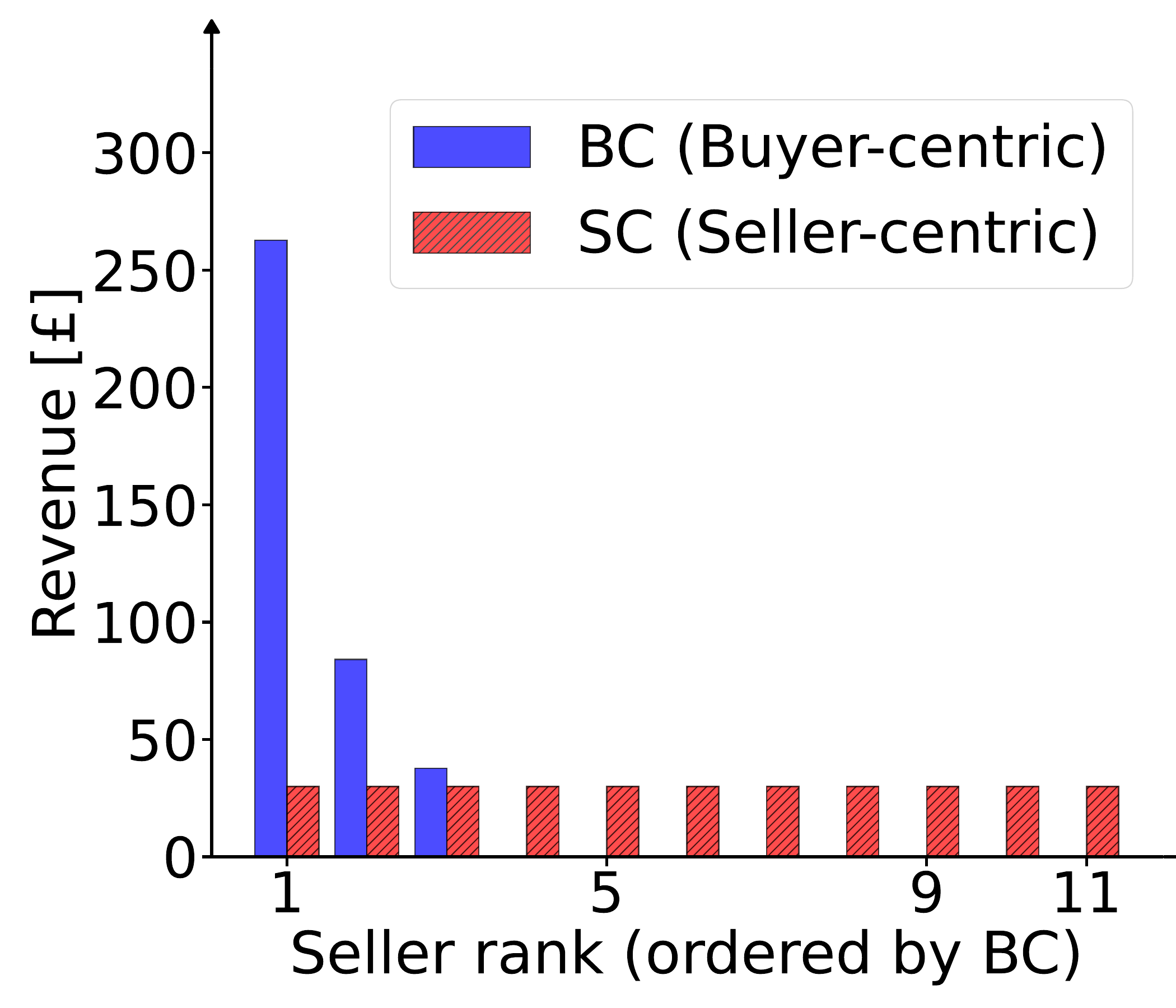}
        \label{fig:Revenue Difference-a}}
    \hfill
    \subfloat[QBCAL]{
        \includegraphics[width=.32\linewidth]{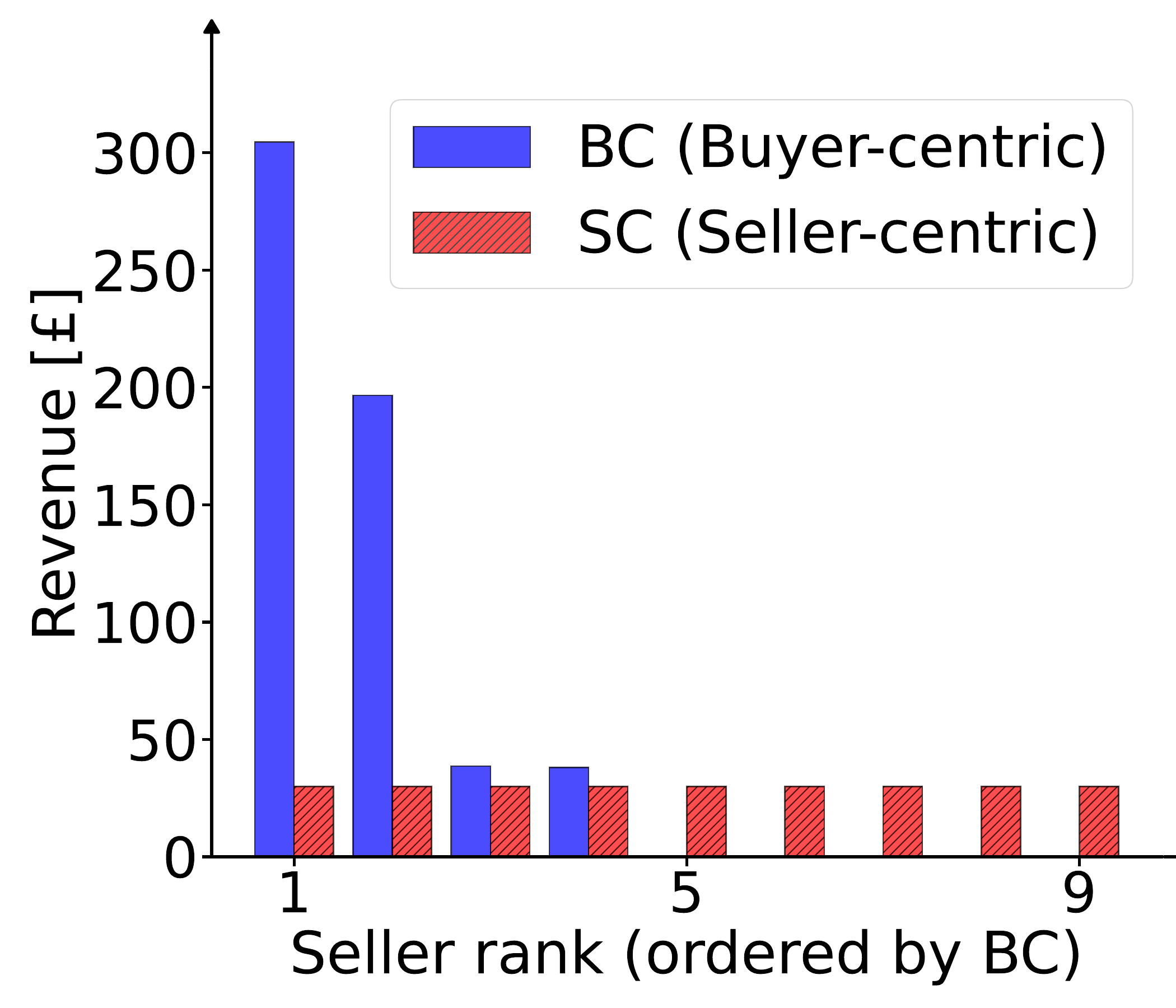}
        \label{fig:Revenue Difference-b}}
    \hfill
    \subfloat[RSC]{
        \includegraphics[width=.32\linewidth]{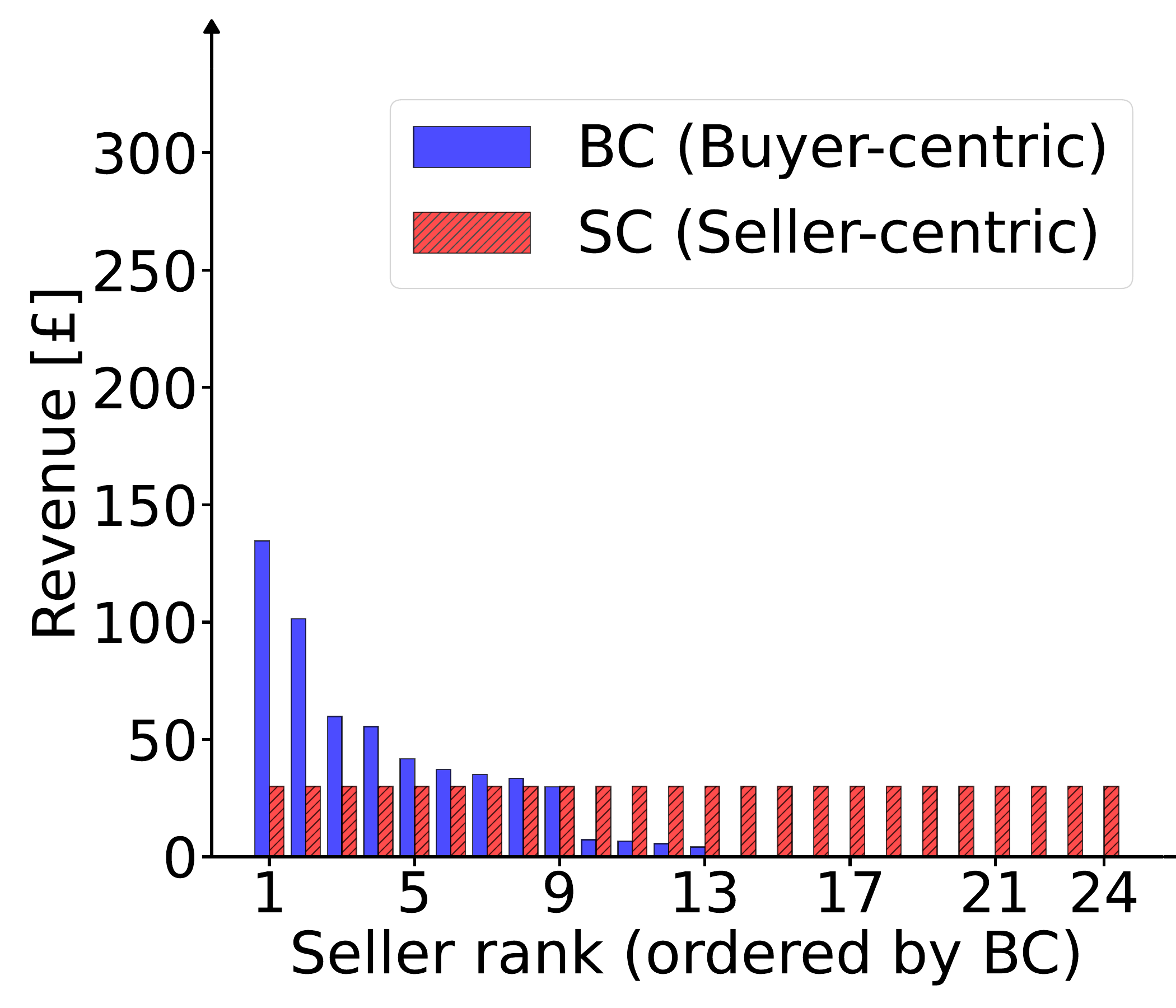}
        \label{fig:Revenue Difference-c}}

    \vspace{3mm}
    \caption{Revenue difference for each data seller under different acquisition and pricing approaches}
    \label{fig:Revenue Difference}
\end{figure}

Figure~\ref{fig:Revenue Difference} highlights the differences in the revenue distributions for sellers under the BC and SC pricing approaches. The BC approach results in a more concentrated revenue distribution compared to the SC approach. This suggests that BC pricing prioritizes sellers whose data points deliver the greatest utility, ensuring that compensation is more closely aligned with the quality and contribution of their data. In contrast, the broader revenue distribution in the SC approach indicates a less targeted allocation of compensation, potentially reducing incentives for sellers to offer high-utility data. Such findings highlight the effectiveness of the BC approach in incentivizing high-quality sellers and fostering efficient data markets by concentrating resources where they are most impactful. 

\subsubsection{Is our active learning strategy robust?}\label{sec:sens_2}

\begin{figure}[H]
    \centering
    \subfloat[VBAL]{
        \includegraphics[width=.32\linewidth]{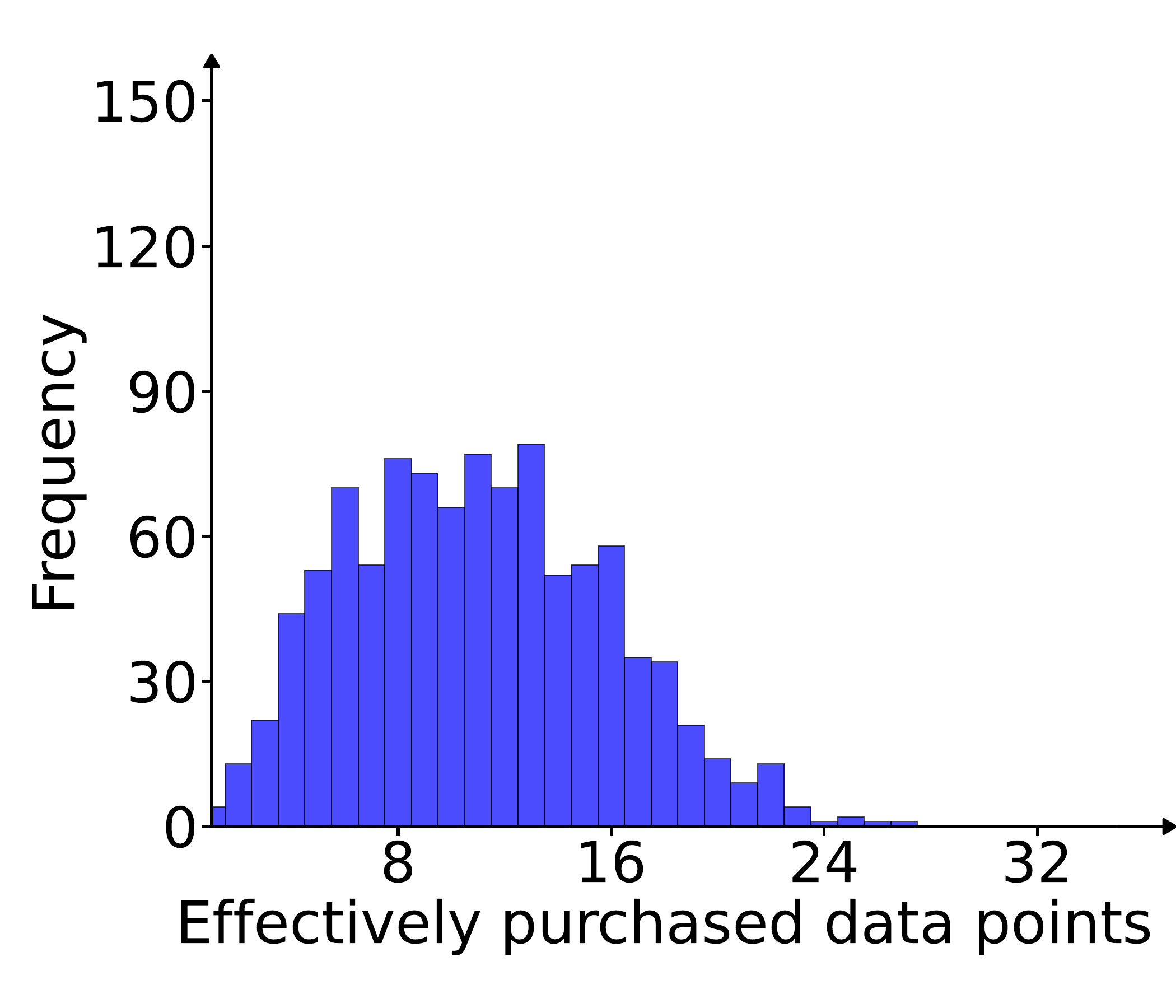}
        \label{fig:Monte_BC-a}}
    \hfill
    \subfloat[QBCAL]{
        \includegraphics[width=.32\linewidth]{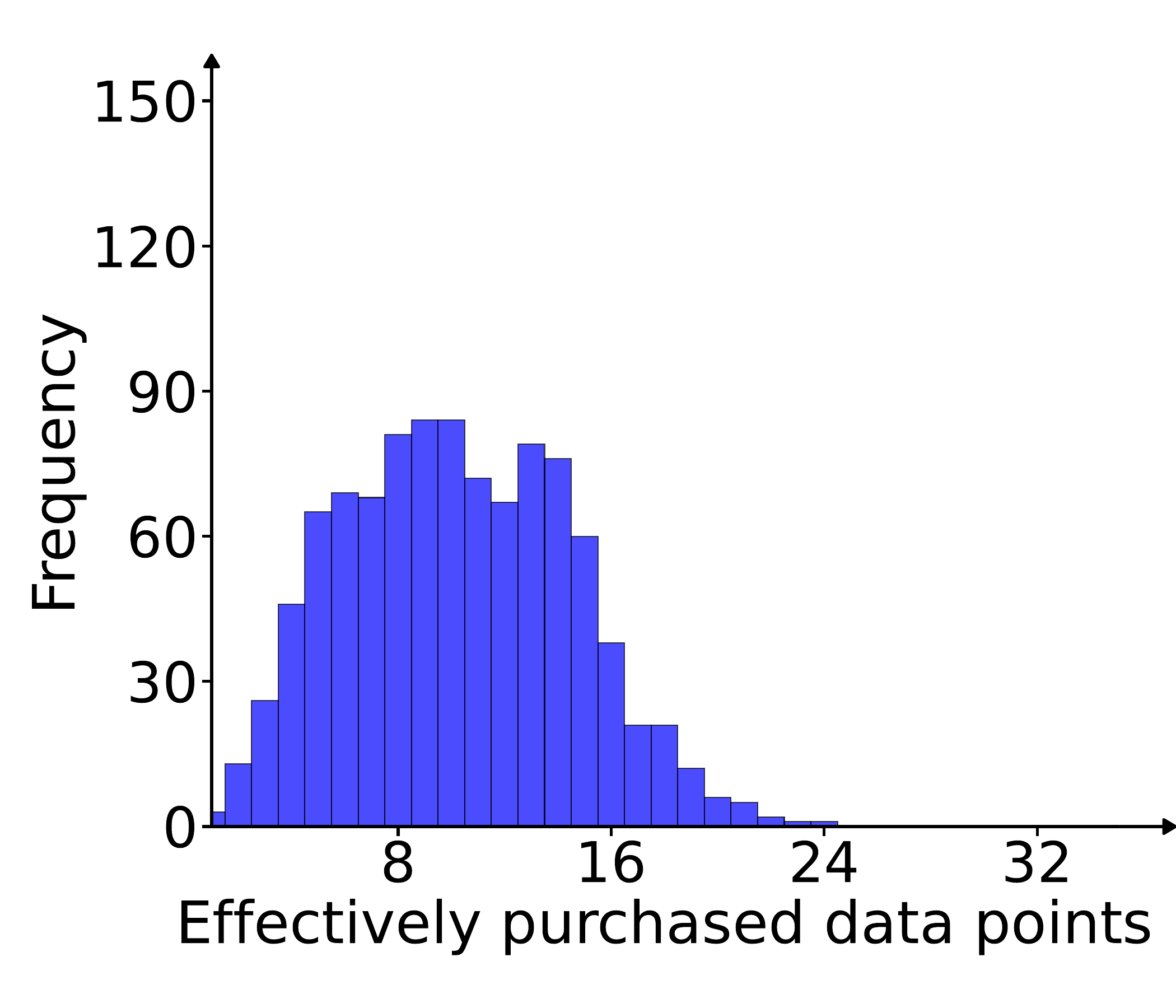}
        \label{fig:Monte_BC-b}}
    \hfill
    \subfloat[RSC]{
        \includegraphics[width=.32\linewidth]{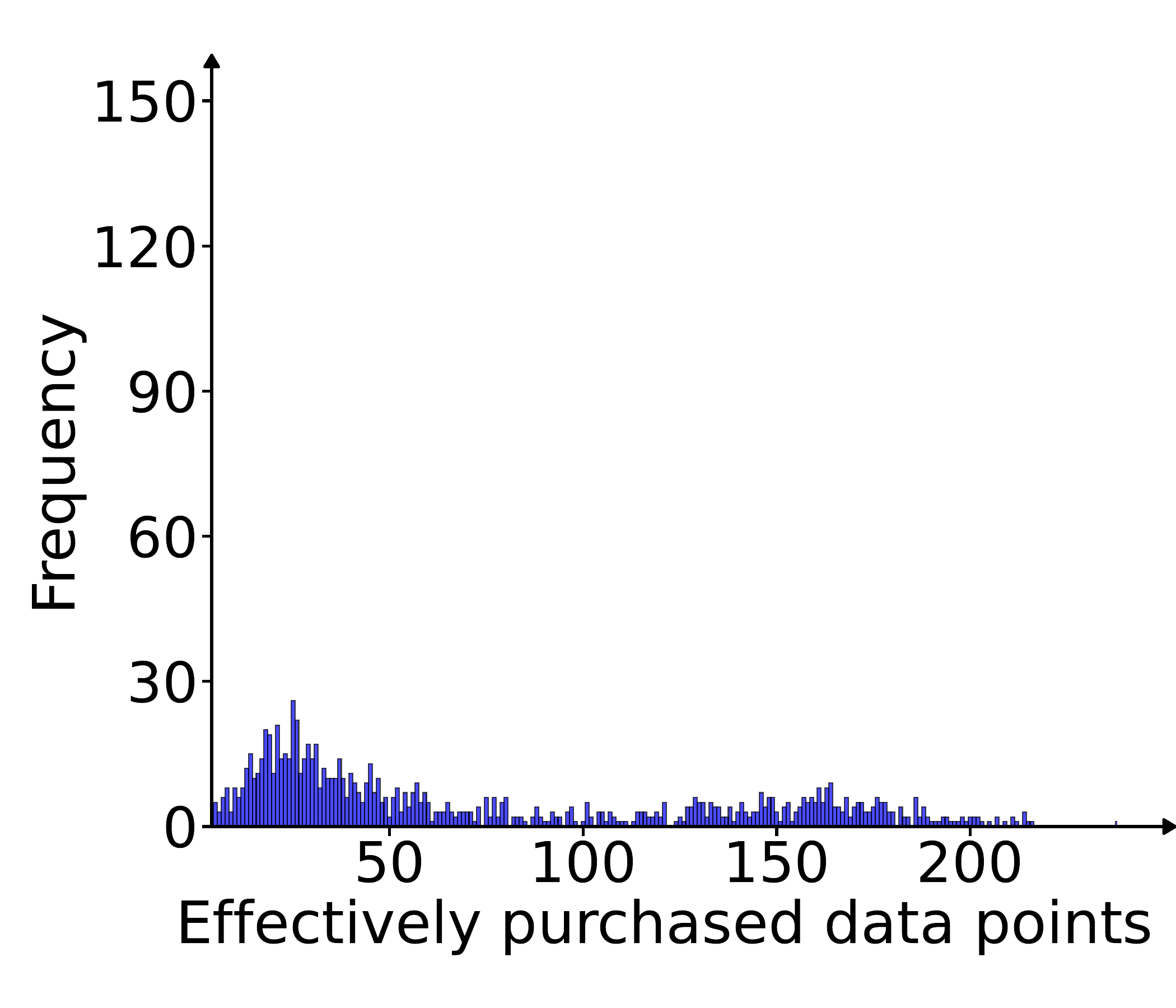}
        \label{fig:Monte_BC-c}}

    \vspace{3mm}
    \caption{Monte-carlo simulation for the buyer-centric approach    \label{fig:Monte_BC}}

\end{figure}

\begin{figure}[H]
    \centering
    \subfloat[VBAL]{
        \includegraphics[width=.32\linewidth]{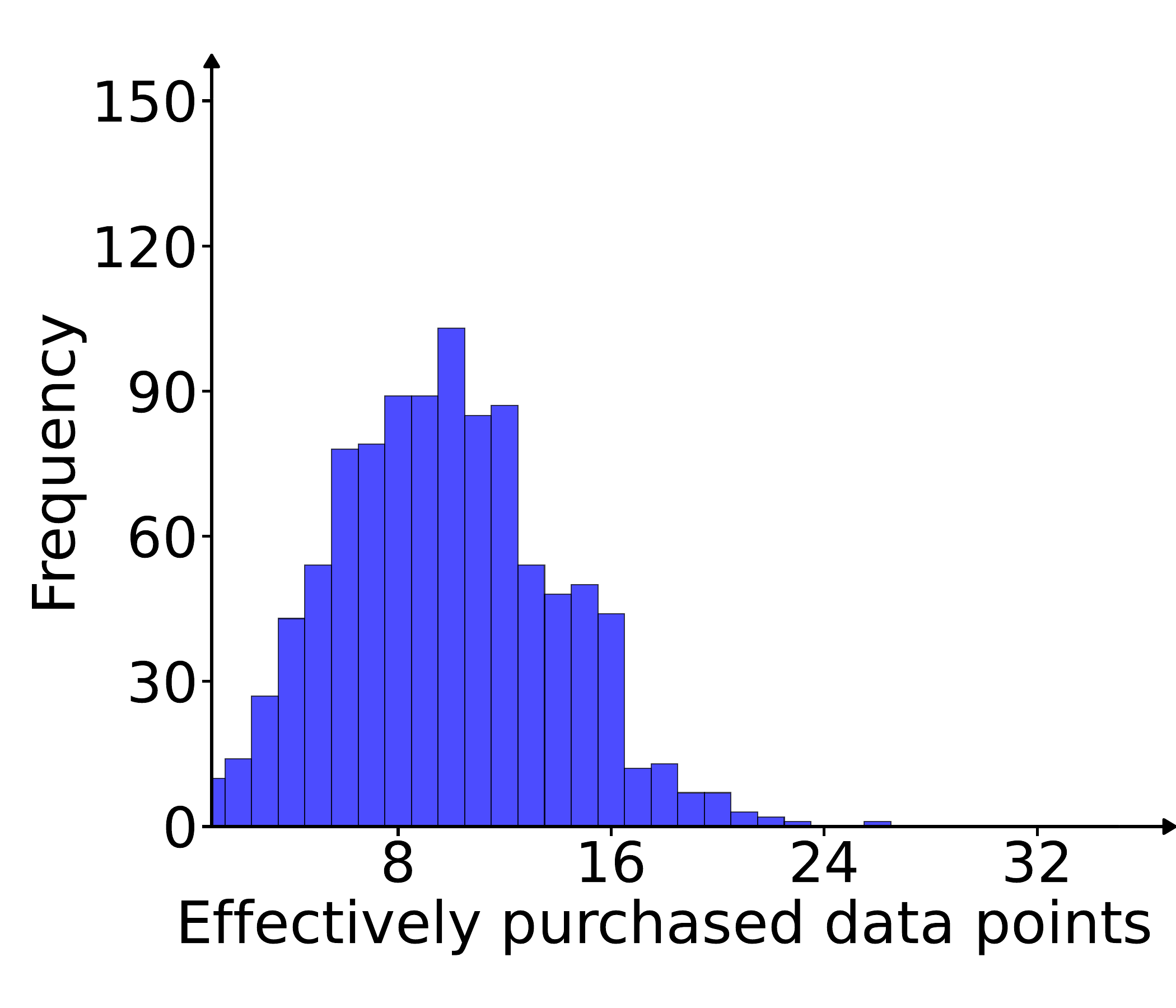}
        \label{fig:Monte_SC-a}}
    \hfill
    \subfloat[QBCAL]{
        \includegraphics[width=.32\linewidth]{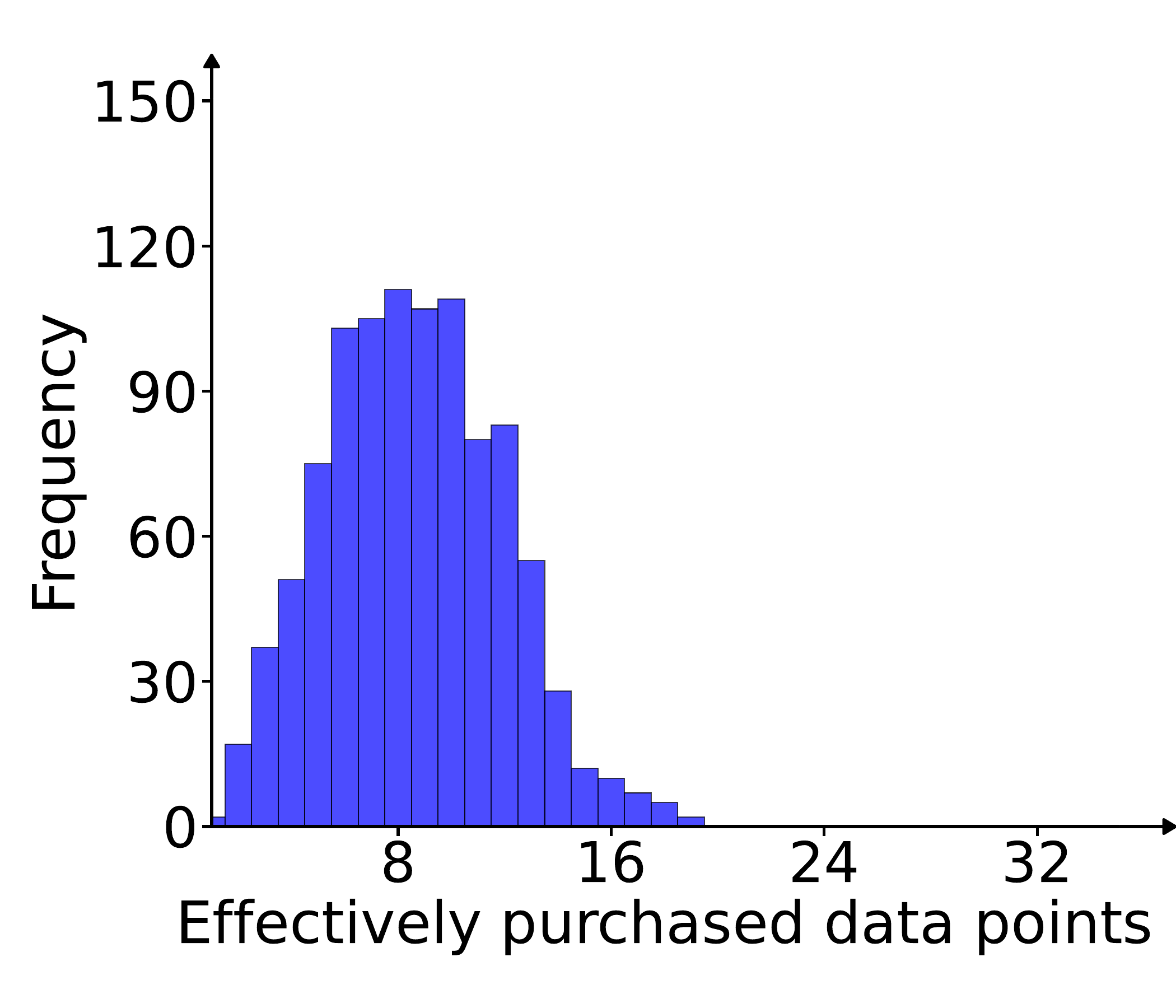}
        \label{fig:Monte_SC-b}}
    \hfill
    \subfloat[RSC]{
        \includegraphics[width=.32\linewidth]{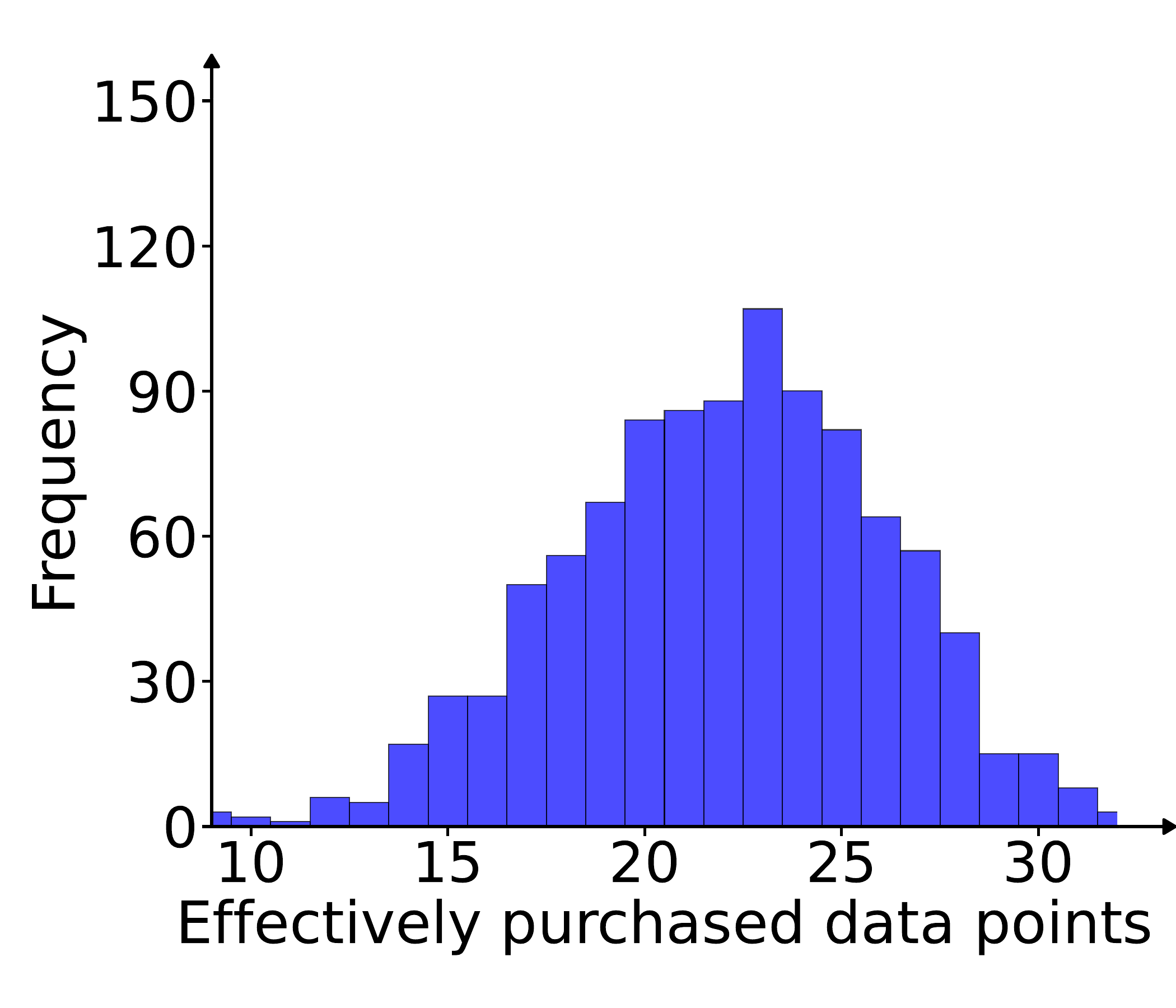}
        \label{fig:Monte_SC-c}}

    \vspace{3mm}
    \caption{Monte Carlo simulation for the seller-centric approach\label{fig:Monte_SC}}
    \end{figure}

\paragraph{Data Variability}
We ran 1{,}000 Monte Carlo iterations with a fixed 20\% improvement target.  
Figures~\ref{fig:Monte_BC} and~\ref{fig:Monte_SC} report the distributions of the data points needed under BC and SC, respectively.  
VBAL and QBCAL show narrow, stable distributions, indicating strong 
robustness to random data variation, while RSC exhibits substantially wider spread.

\begin{figure}[t]
\centering

\begin{subfigure}[t]{0.44\textwidth}
  \centering
  \includegraphics[width=\linewidth,clip,trim=0.9cm 0.6cm 0.9cm 0.6cm]{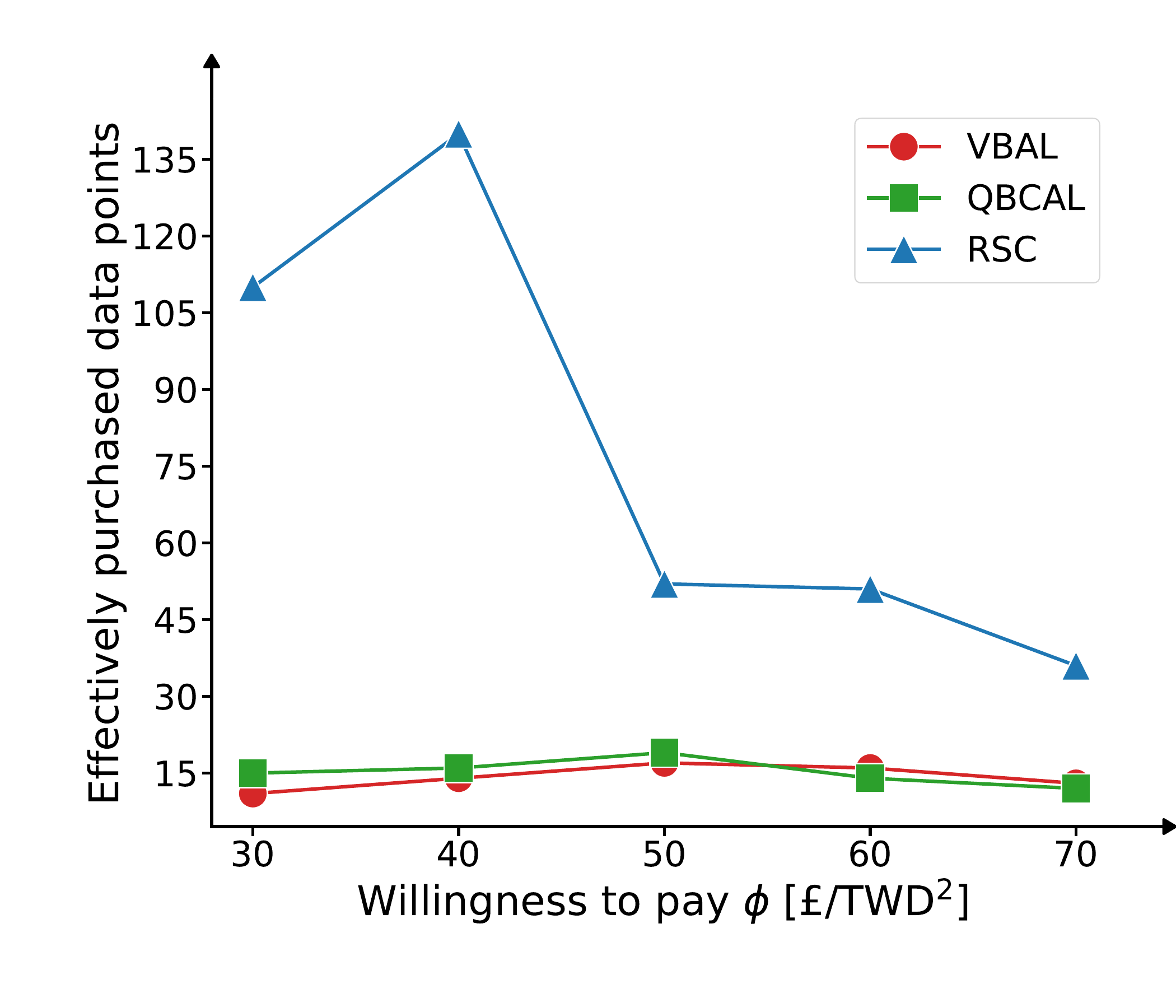}
  \caption{WTP variation under BC pricing}
      \label{fig:2_sensitivity_WTP_BC}
\end{subfigure}\hfill
\begin{subfigure}[t]{0.44\textwidth}
  \centering
  \includegraphics[width=\linewidth,clip,trim=0.9cm 0.6cm 0.9cm 0.6cm]{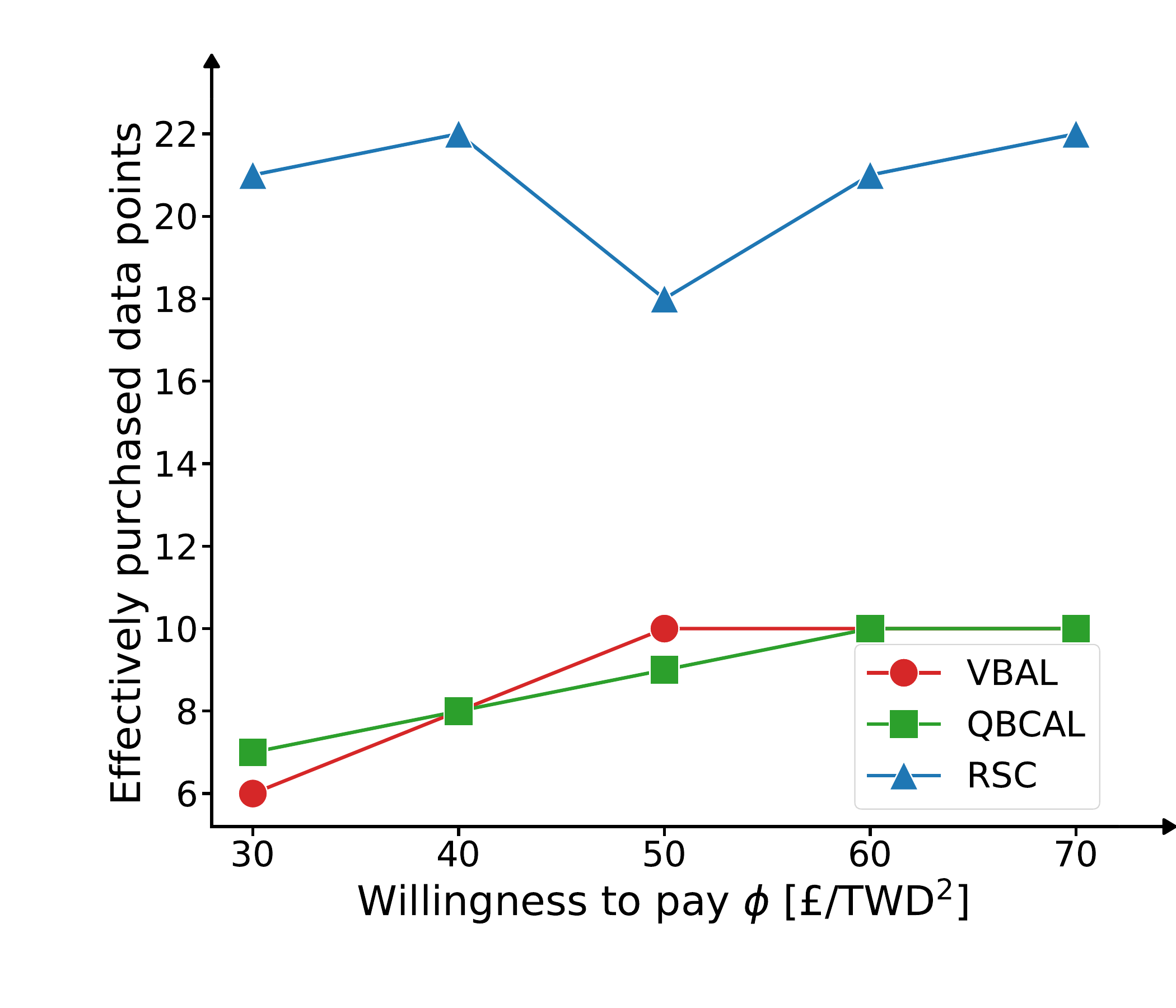}
  \caption{WTP variation under SC pricing}
      \label{fig:2_sensitivity_WTP_SC}
\end{subfigure}

\vspace{0.6em} 

\begin{subfigure}{0.44\textwidth}
    \centering
    \includegraphics[width=\linewidth,clip,trim=0.9cm 0.6cm 0.9cm 0.6cm]{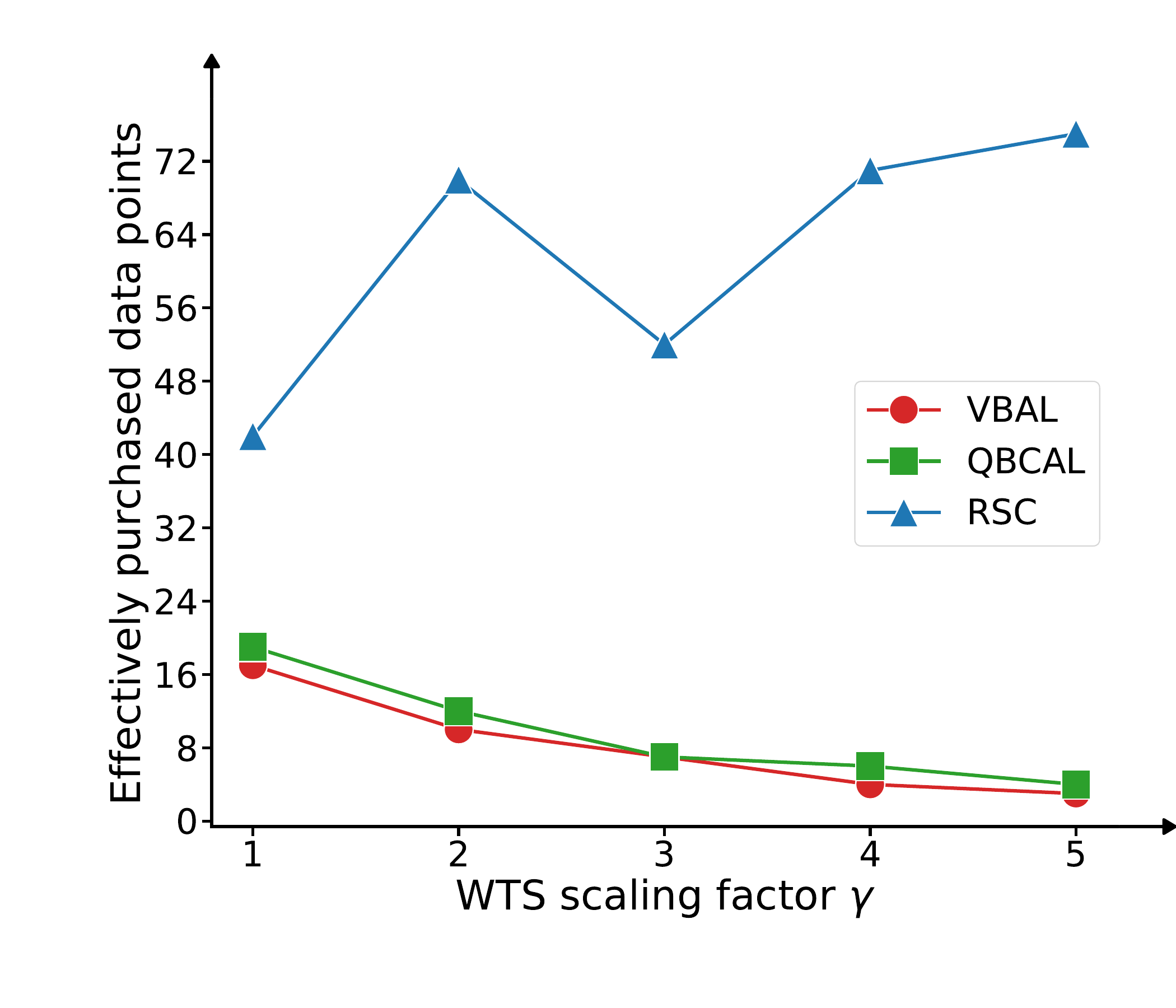}
    \caption{WTS variation under BC pricing}
    \label{fig:2_sensitivity_WTS_BC}
\end{subfigure}\hfill
\begin{subfigure}{0.44\textwidth}
    \centering
    \includegraphics[width=\linewidth,clip,trim=0.9cm 0.6cm 0.9cm 0.6cm]{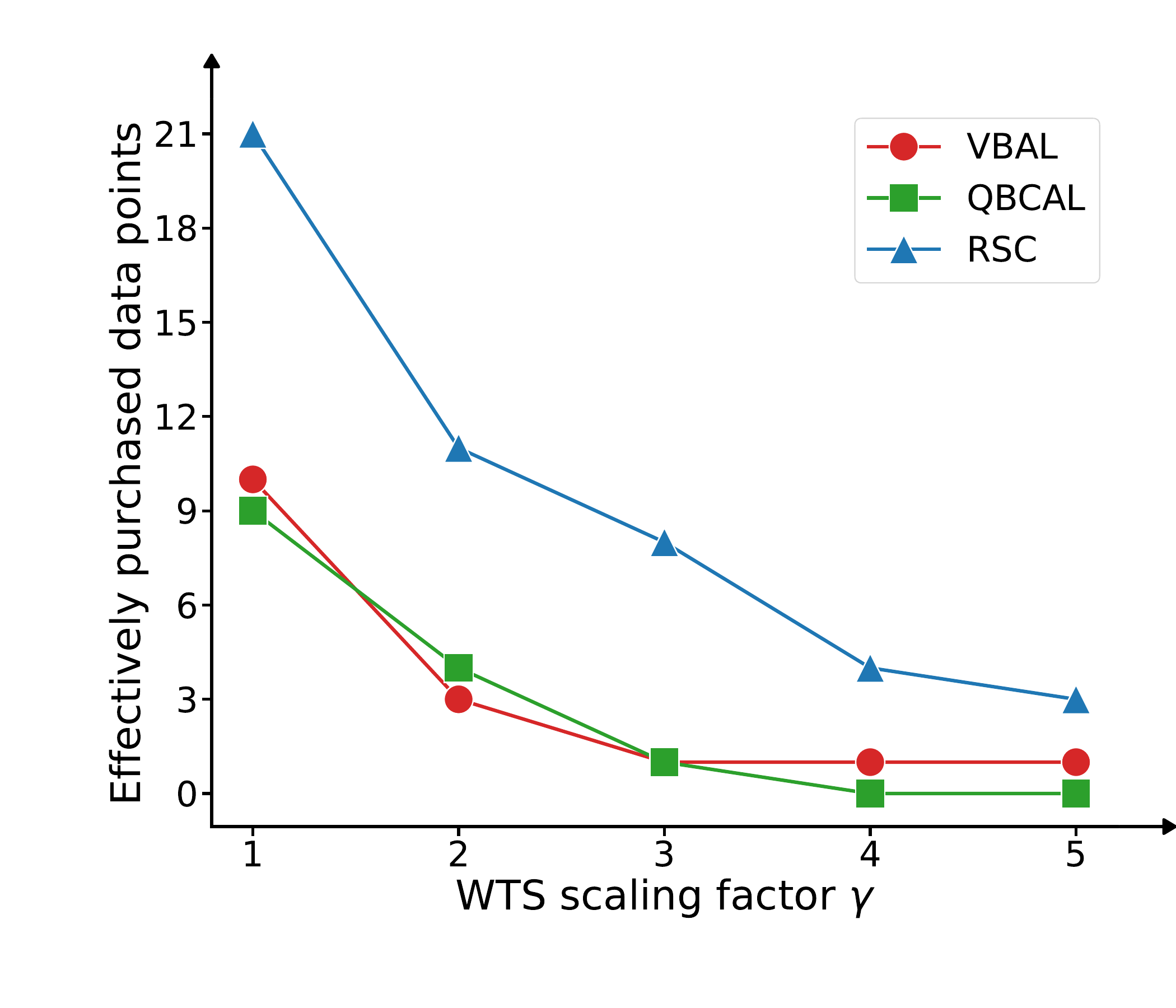}
    \caption{WTS variation under SC pricing}
    \label{fig:2_sensitivity_WTS_SC}
\end{subfigure}

\vspace{0.6em}

\begin{subfigure}{0.44\textwidth}
    \centering
    \includegraphics[width=\linewidth]{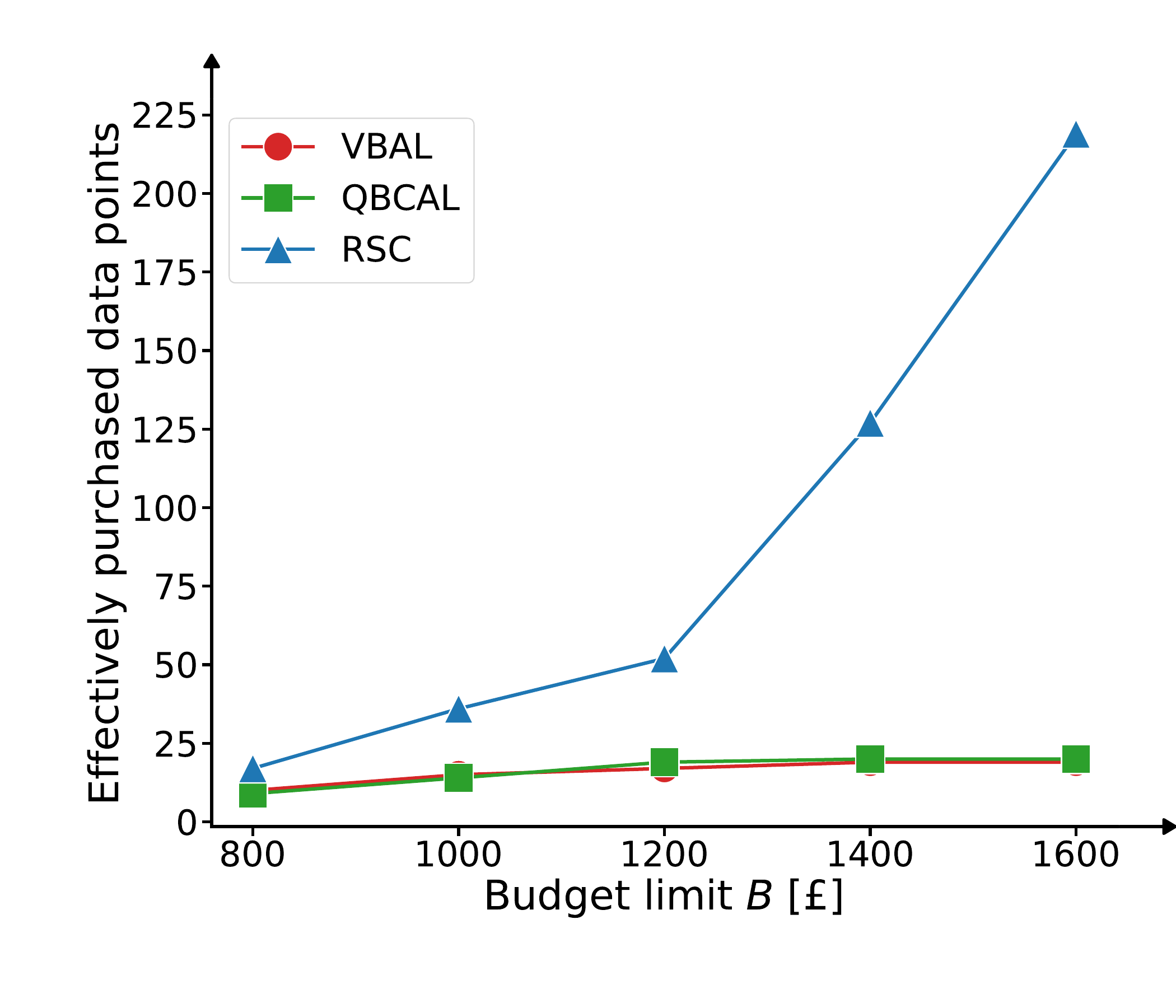}
    \caption{Budget variation under BC pricing}
    \label{fig:2_sensitivity_B_BC}
\end{subfigure}\hfill
\begin{subfigure}{0.44\textwidth}
    \centering
    \includegraphics[width=\linewidth]{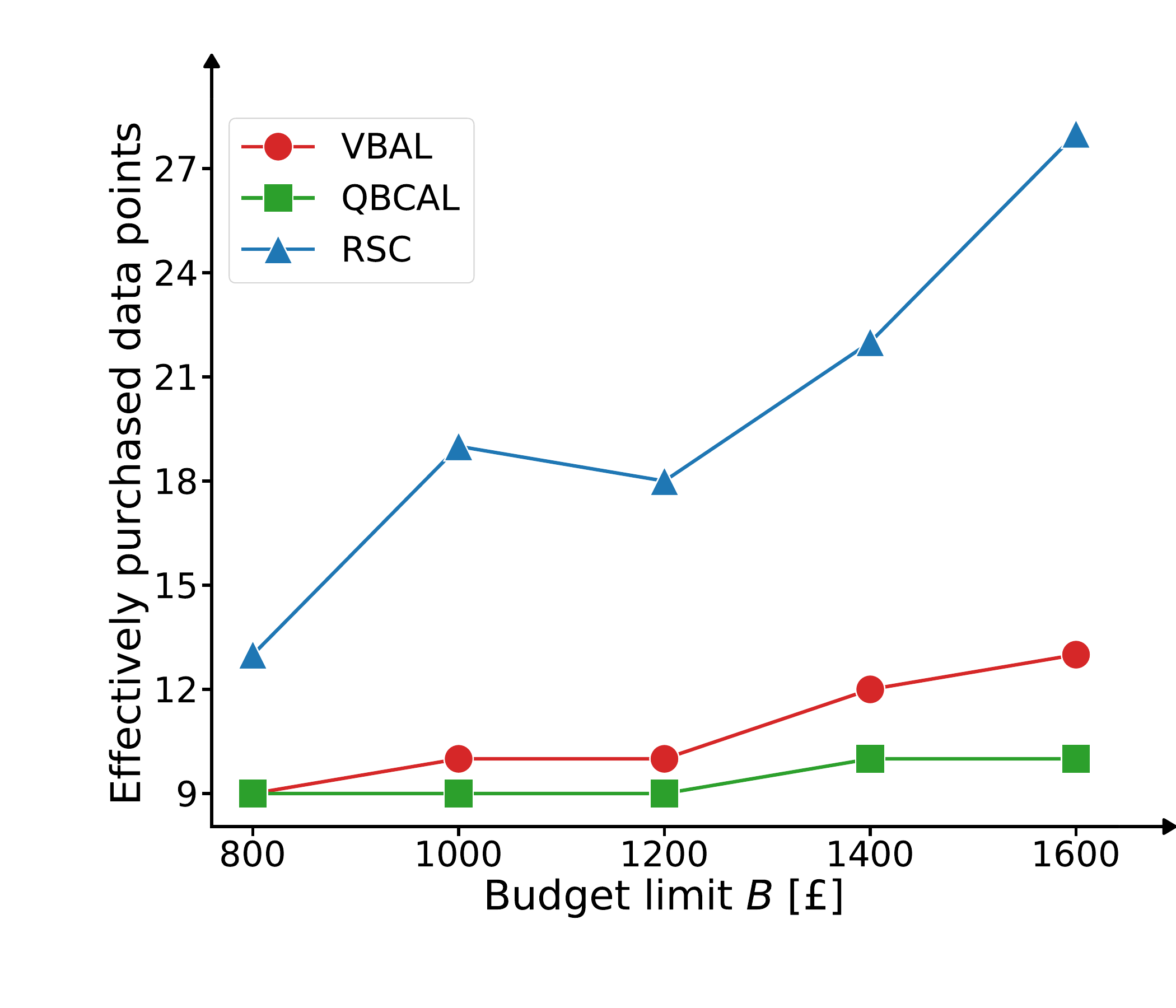}
    \caption{Budget variation under SC pricing}
    \label{fig:2_sensitivity_B_SC}
\end{subfigure}

\vspace{0.6em}

\caption{Sensitivity analyses of the buyer’s willingness-to-pay ($\phi$), the seller’s willingness-to-sell (WTS) scaling coefficient ($\gamma$), and the budget limit ($B$). The left column shows the results for buyer-centric (BC) pricing, while the right column corresponds to seller-centric (SC) pricing.}

\label{fig:sensitivity_combined}
\end{figure}

\paragraph{Parameter Sensitivity}
We examine robustness through two analyses: (i) one-at-a-time parameter variations
(Figure~\ref{fig:sensitivity_combined}) and (ii) a Monte Carlo study over 50 random splits (summary in Table~\ref{tab:montecarlo_bc_sc_mse}, Appendix).  
Figures~\ref{fig:2_sensitivity_WTP_BC}--\ref{fig:2_sensitivity_WTP_SC} show the effect of WTP: in BC, VBAL and QBCAL purchase more data as $\phi$ increases up to~50, then decline once the budget constraint~\eqref{constraint4} binds; in SC, effective purchases decrease monotonically with~WTP, while RSC fluctuates due to its non-adaptive design.  
Figures~\ref{fig:2_sensitivity_WTS_BC}--\ref{fig:2_sensitivity_WTS_SC} vary the seller’s WTS scale, and all methods buy fewer labels as the WTS increases.  
Figures~\ref{fig:2_sensitivity_B_BC}--\ref{fig:2_sensitivity_B_SC} vary budget limits (\textit{with the improvement threshold raised to 90\% to sharpen trends}): more budget leads to more acquisitions, with VBAL and QBCAL consistently requiring fewer points than RSC. In general, both VBAL and QBCAL remain selective and cost-effective across all settings.

\captionsetup[subfigure]{font=small,skip=2pt}

\clearpage
\section{Conclusions and perspectives for future work} \label{conclusions}

\subsection{Conclusions}
Active learning markets are introduced as a way to purchase labels using active learning approaches, as motivated by real-world applications. The proposed framework addresses challenges prevalent across various industrial sectors, including real estate and energy forecasting, which serve as the primary scenarios in our analysis. Specifically, we employ two active learning approaches for cost-effective data acquisition: variance-based active learning (VBAL) and Query-by-Committee active learning (QBCAL). Additionally, we examine two distinct pricing mechanisms -- buyer-centric (BC) and seller-centric (SC)—representing extreme real-world conditions. Our analysis highlights the payment structure for data analysts and compares revenue distributions among data sellers under the proposed active learning approaches and established benchmarks.

This work yields three key findings. First, both VBAL and QBCAL demonstrate significant cost-efficiency in purchasing data when benchmarked against a random sampling approach. Second, under the two extreme pricing conditions, active learning leads to a more concentrated revenue distribution. Third, while both VBAL and QBCAL outperform random sampling in BC and SC pricing scenarios, practical market designs may benefit from hybrid pricing mechanisms to better define data prices in active learning markets.

Importantly, our study is the first to conceptualize the active learning market while building on prior research into observation and feature markets. The methodologies employed—spanning machine learning, active learning, and optimal experimental design (OED)—offer a rich foundation for future exploration. 

\subsection{Future works}\label{sec:future works}
Efficient data purchase is becoming crucial in our data-driven economies. Our proposed innovative active learning market provides a strong foundation to address this significant problem by offering a trade-off between cost and efficiency. Building on this success, further advancements can enhance its applicability. First, while current data is purchased in a batch setting, real-world data may arrive continuously as a stream, creating opportunities to develop a real-time label market. Second, a valuable extension is to investigate the multiple-buyer-multiple-seller settings which involve competition between sellers and buyers in a resource-constrained environment. In such competitive environments, game-theoretic tools and combinatorial optimization methods (e.g., knapsack heuristics) may play a larger role, complementing active learning in analyzing pricing, market dynamics, and efficiency. Additionally, a one-shot formulation of VBAL is also possible, even in the single-buyer–multiple-seller case, where the budget is allocated in a single step. This connects naturally to knapsack heuristics, but requires the strong assumption that the utilities (or priorities) of all labels are known in advance. For this reason, we focus on iterative allocation, which better reflects adaptive practice, while leaving one-shot variants as an interesting theoretical direction for future work.
Third, the design of the market for data and information requires careful reconsideration. Unlike traditional commodities, the uniqueness of data lies in defining both willingness to pay and willingness to sell, as its value is highly context-dependent. For data buyers, their WTP may depend on relevance of tasks, timeliness of sharing, quality of data, among others. For data sellers, their WTS can be influenced by the cost of acquiring data (and storing or sharing it), heterogeneous competition, or privacy constraints. Studying these interplay factors will provide a more comprehensive understanding of how the data market should be structured. Through modelling those dynamics, one can develop more rigorous pricing mechanisms that incorporate strategic behaviour, and thus ensure a fair, efficient, and incentive-compatible data market in the real-world scenarios.
\minorrev{Fourth, while our empirical results are strong, we note that deriving formal performance or regret guarantees is particularly challenging in our setting. Classical theoretical results in active learning typically assume that all labels have identical unit cost, while our market-based setting allows heterogeneous prices that depend on sellers’ costs or preferences. This means that performance depends jointly on label informativeness and cost, and utilities evolve adaptively as labels are acquired, making standard analyses inapplicable. Developing theoretical guarantees in such heterogeneous-cost settings is therefore a challenging but important direction for future research. More concrete partial insights may still be possible under stylised assumptions, e.g., bounded or i.i.d. cost heterogeneity, and informativeness proxies that are bounded and (approximately) unbiased for the true marginal improvement. In such regimes, one could aim for comparison-based guarantees against random baselines, or one-step (myopic) optimality/Pareto-efficiency properties of cost-weighted greedy rules.}
 Finally, while our framework currently focuses on linear regression and models linear in parameters, extending it to non-convex models such as deep neural networks is an important direction. Estimating label value in such settings is substantially more challenging due to non-convex optimization landscapes, stochastic training procedures, and state dependence. Possible avenues include using influence functions, gradient-based embeddings, or Bayesian uncertainty estimates to approximate label utilities.


%

\clearpage
\appendix
\section{Monte Carlo results}

This appendix reports additional Monte Carlo simulation results that complement the main estimation results presented in Sections~\ref{sec:sens_1} (Table~\ref{tab:montecarlo_bc_sc_var}) and~\ref{sec:sens_2} (Table~\ref{tab:montecarlo_bc_sc_mse}). 
For each configuration, we vary key model parameters—namely willingness-to-pay (WTP)~$\phi$, willingness-to-sell (WTS) scaling coefficient~$\eta$, and budget limit~$B$—and report the mean and interquartile range of purchased data points across 50 independent runs. 

\begin{table}[!htbp]
\centering
\begingroup

\caption{Monte Carlo summary (50 runs) for estimation quality under the buyer-centric (BC)
and seller-centric (SC) pricing schemes. Values denote mean [p25--p75] of purchased data points.}
\label{tab:montecarlo_bc_sc_var}

\footnotesize
\setlength{\extrarowheight}{0pt}
\captionsetup{skip=2pt} 
\renewcommand{\arraystretch}{0.9}

\begin{tabular}{lcccc}
\toprule
\multicolumn{5}{c}{\textbf{BC}} \\
\midrule
\textbf{Parameter / Value} & VBAL & QBCAL & RSC & GreedyKnapsack \\
\midrule

\multicolumn{5}{l}{\textit{WTP $\phi$ [£/TWD$^2$]}} \\
\midrule
1000 & 8.06 [7.25--9.00] & 8.12 [7.00--9.00] & 18.34 [16.00--20.75] & 29.74 [27.00--33.00] \\
1050 & 8.16 [8.00--9.00] & 8.38 [7.00--9.00] & 18.34 [16.00--20.75] & 29.74 [27.00--33.00] \\
1100 & 8.20 [8.00--9.00] & 8.46 [7.00--9.00] & 18.34 [16.00--20.75] & 29.74 [27.00--33.00] \\
1150 & 8.34 [8.00--9.00] & 8.58 [8.00--9.00] & 18.34 [16.00--20.75] & 29.74 [27.00--33.00] \\
1200 & 8.36 [8.00--9.00] & 8.66 [8.00--9.00] & 18.34 [16.00--20.75] & 29.74 [27.00--33.00] \\

\midrule
\multicolumn{5}{l}{\textit{WTS scaling coefficient $\eta$}} \\
\midrule
0.5 & 8.36 [8.00--9.00] & 8.66 [8.00--9.00] & 18.34 [16.00--20.75] & 29.74 [27.00--33.00] \\
1   & 6.56 [4.00--8.00] & 6.58 [5.00--8.00] & 18.34 [16.00--20.75] & 29.74 [27.00--33.00] \\
1.5 & 4.56 [2.00--7.00] & 4.68 [3.00--7.00] & 18.34 [16.00--20.75] & 29.74 [27.00--33.00] \\
2   & 3.22 [2.00--4.00] & 3.46 [2.00--4.75] & 18.34 [16.00--20.75] & 29.74 [27.00--33.00] \\
2.5 & 2.48 [2.00--3.00] & 2.62 [2.00--3.00] & 18.34 [16.00--20.75] & 29.74 [27.00--33.00] \\

\midrule
\multicolumn{5}{l}{\textit{Budget limit $B$ [£]}} \\
\midrule
10 & 8.14 [8.00--9.00] & 8.42 [8.00--9.00] & 17.88 [16.00--20.00] & 28.84 [25.25--33.00] \\
15 & 8.36 [8.00--9.00] & 8.66 [8.00--9.00] & 18.34 [16.00--20.75] & 29.74 [27.00--33.00] \\
20 & 8.36 [8.00--9.00] & 8.66 [8.00--9.00] & 18.34 [16.00--20.75] & 29.74 [27.00--33.00] \\
25 & 8.36 [8.00--9.00] & 8.66 [8.00--9.00] & 18.34 [16.00--20.75] & 29.74 [27.00--33.00] \\
30 & 8.36 [8.00--9.00] & 8.66 [8.00--9.00] & 18.34 [16.00--20.75] & 29.74 [27.00--33.00] \\
\bottomrule
\end{tabular}

\vspace{0.9em}

\begin{tabular}{lcccc}
\toprule
\multicolumn{5}{c}{\textbf{SC}} \\
\midrule
\textbf{Parameter / Value} & VBAL & QBCAL & RSC & GreedyKnapsack \\
\midrule

\multicolumn{5}{l}{\textit{WTP $\phi$ [£/TWD$^2$]}} \\
\midrule
1000 & 8.06 [7.25--9.00] & 7.44 [6.25--8.75] & 18.30 [16.00--20.75] & 29.74 [27.00--33.00] \\
1050 & 8.14 [8.00--9.00] & 7.74 [7.00--9.00] & 18.30 [16.00--20.75] & 29.74 [27.00--33.00] \\
1100 & 8.16 [8.00--9.00] & 7.94 [7.00--9.00] & 18.30 [16.00--20.75] & 29.74 [27.00--33.00] \\
1150 & 8.26 [8.00--9.00] & 8.08 [7.00--9.00] & 18.30 [16.00--20.75] & 29.74 [27.00--33.00] \\
1200 & 8.34 [8.00--9.00] & 8.16 [7.00--9.00] & 18.30 [16.00--20.75] & 29.74 [27.00--33.00] \\

\midrule
\multicolumn{5}{l}{\textit{WTS scaling coefficient $\eta$}} \\
\midrule
0.5 & 8.34 [8.00--9.00] & 8.16 [7.00--9.00] & 18.30 [16.00--20.75] & 29.74 [27.00--33.00] \\
1   & 6.40 [4.00--8.00] & 4.68 [3.00--7.00] & 14.36 [13.00--16.00] & 27.48 [26.00--29.00] \\
1.5 & 4.30 [2.00--6.75] & 2.80 [1.00--4.75] & 10.00 [9.00--11.00] & 24.06 [23.00--26.00] \\
2   & 3.10 [2.00--4.00] & 2.10 [1.00--3.00] & 7.74 [7.00--9.00] & 21.70 [21.00--23.00] \\
2.5 & 2.48 [2.00--3.00] & 1.60 [1.00--3.00] & 6.40 [6.00--7.00] & 18.88 [18.00--20.00] \\

\midrule
\multicolumn{5}{l}{\textit{Budget limit $B$ [£]}} \\
\midrule
10 & 8.28 [8.00--9.00] & 7.72 [7.00--9.00] & 16.44 [14.25--18.00] & 28.22 [27.00--31.00] \\
15 & 8.34 [8.00--9.00] & 8.16 [7.00--9.00] & 18.30 [16.00--20.75] & 29.74 [27.00--33.00] \\
20 & 8.36 [8.00--9.00] & 8.32 [7.00--9.00] & 18.34 [16.00--20.75] & 29.74 [27.00--33.00] \\
25 & 8.36 [8.00--9.00] & 8.38 [7.00--9.00] & 18.34 [16.00--20.75] & 29.74 [27.00--33.00] \\
30 & 8.36 [8.00--9.00] & 8.56 [7.25--9.00] & 18.34 [16.00--20.75] & 29.74 [27.00--33.00] \\
\bottomrule
\end{tabular}

\endgroup
\end{table}

\begin{table}[!htbp]
\centering
\caption{Monte Carlo summary (50 runs) for \textbf{predictive ability}. Values denote the mean and interquartile range [p25--p75] of effectively purchased data points.}
\label{tab:montecarlo_bc_sc_mse}

\footnotesize
\setlength{\extrarowheight}{0pt}
\captionsetup{skip=2pt} 
\renewcommand{\arraystretch}{0.9}
\begin{tabular}{lccc}
\toprule
 & \multicolumn{3}{c}{\textbf{BC}} \\
\cmidrule(lr){2-4}
\textbf{Parameter / Value} & \textbf{VBAL} & \textbf{QBCAL} & \textbf{RSC} \\
\midrule

\multicolumn{4}{l}{\textit{WTP $\phi$ [£/TWD$^2$]}} \\
\midrule
30 & 8.84 [7.25--10.00] & 8.96 [7.00--11.00] & 120.36 [71.50--159.75] \\
40 & 10.36 [8.00--13.00] & 10.60 [7.25--13.00] & 113.76 [53.75--159.75] \\
50 & 10.46 [7.00--13.75] & 10.98 [7.00--14.00] & 88.88 [35.00--141.50] \\
60 & 10.24 [7.00--14.00] & 10.42 [6.00--13.00] & 70.90 [25.25--125.50] \\
70 & 9.56 [6.00--13.00] & 10.10 [7.00--13.00] & 59.62 [21.25--103.00] \\

\midrule
\multicolumn{4}{l}{\textit{WTS scaling coefficient $\eta$}} \\
\midrule
1 & 10.46 [7.00--13.75] & 10.98 [7.00--14.00] & 88.88 [35.00--141.50] \\
2 & 6.70 [5.00--8.00] & 6.76 [5.25--8.00] & 88.88 [35.00--141.50] \\
3 & 4.64 [3.25--6.00] & 5.00 [3.25--7.00] & 88.88 [35.00--141.50] \\
4 & 3.56 [2.00--5.00] & 3.80 [3.00--5.00] & 88.88 [35.00--141.50] \\
5 & 2.76 [1.00--4.00] & 2.94 [1.00--4.00] & 88.88 [35.00--141.50] \\

\midrule
\multicolumn{4}{l}{\textit{Budget limit $B$ [£]}} \\
\midrule
800  & 7.42 [5.00--9.75] & 8.02 [6.00--9.75] & 53.94 [20.25--68.00] \\
1000 & 9.20 [6.00--12.75] & 9.56 [6.00--12.00] & 70.90 [25.25--125.50] \\
1200 & 10.46 [7.00--13.75] & 10.98 [7.00--14.00] & 88.88 [35.00--141.50] \\
1400 & 11.44 [8.25--14.00] & 11.90 [8.00--14.75] & 106.36 [46.50--154.00] \\
1600 & 12.12 [9.25--15.00] & 12.56 [9.00--16.00] & 116.78 [57.00--159.75] \\
\bottomrule
\end{tabular}

\vspace{0.3em}
\begin{tabular}{lccc}
\toprule
 & \multicolumn{3}{c}{\textbf{SC}} \\
\cmidrule(lr){2-4}
\textbf{Parameter / Value} & \textbf{VBAL} & \textbf{QBCAL} & \textbf{RSC} \\
\midrule

\multicolumn{4}{l}{\textit{WTP $\phi$ [£/TWD$^2$]}} \\
\midrule
30 & 6.62 [5.00--8.00] & 5.80 [4.00--8.00] & 22.18 [20.00--24.00] \\
40 & 7.96 [6.00--10.00] & 7.02 [5.00--9.00] & 22.18 [20.00--24.00] \\
50 & 9.00 [7.00--11.00] & 8.20 [5.25--11.00] & 22.18 [20.00--24.00] \\
60 & 9.86 [8.00--12.00] & 8.84 [6.00--12.00] & 22.18 [20.00--24.00] \\
70 & 10.44 [8.00--12.75] & 9.40 [7.00--12.00] & 22.18 [20.00--24.00] \\

\midrule
\multicolumn{4}{l}{\textit{WTS scaling coefficient $\eta$}} \\
\midrule
1 & 9.00 [7.00--11.00] & 8.20 [5.25--11.00] & 22.18 [20.00--24.00] \\
2 & 4.56 [3.00--6.00] & 3.94 [3.00--5.00] & 11.02 [10.00--12.00] \\
3 & 2.78 [1.25--4.00] & 2.44 [1.00--3.00] & 7.52 [6.00--9.00] \\
4 & 1.76 [1.00--3.00] & 1.60 [1.00--2.00] & 5.22 [4.00--6.00] \\
5 & 1.20 [0.00--2.00] & 1.14 [0.00--2.00] & 4.18 [3.00--5.00] \\

\midrule
\multicolumn{4}{l}{\textit{Budget limit $B$ [£]}} \\
\midrule
800  & 7.76 [6.00--9.00] & 6.86 [5.00--9.00] & 14.90 [13.00--16.75] \\
1000 & 8.46 [7.00--10.00] & 7.62 [5.25--10.00] & 18.72 [16.00--20.00] \\
1200 & 9.00 [7.00--11.00] & 8.20 [5.25--11.00] & 22.18 [20.00--24.00] \\
1400 & 9.50 [7.00--11.75] & 8.68 [6.00--11.00] & 25.36 [23.00--28.00] \\
1600 & 9.82 [7.25--12.00] & 9.08 [7.00--11.75] & 28.80 [26.00--31.75] \\
\bottomrule
\end{tabular}

\end{table}

\clearpage
\bibliographystyle{plainnat}
\bibliography{my} 






  



\end{document}